# Algorithms for Generating Ordered Solutions for Explicit AND/OR Structures

**Priyankar Ghosh**                                          PRIYANKAR@CSE.IITKGP.ERNET.IN
**Amit Sharma**                                              AMIT.ONTOP@GMAIL.COM
**P. P. Chakrabarti**                                        PPCHAK@CSE.IITKGP.ERNET.IN
**Pallab Dasgupta**                                          PALLAB@CSE.IITKGP.ERNET.IN
*Department of Computer Science and Engineering*
*Indian Institute of Technology Kharagpur*
*Kharagpur-721302, India*

## Abstract

We present algorithms for generating alternative solutions for explicit acyclic AND/OR structures in non-decreasing order of cost. The proposed algorithms use a best first search technique and report the solutions using an implicit representation ordered by cost. In this paper, we present two versions of the search algorithm – (a) an initial version of the best first search algorithm, ASG, which may present one solution more than once while generating the ordered solutions, and (b) another version, LASG, which avoids the construction of the duplicate solutions. The actual solutions can be reconstructed quickly from the implicit compact representation used. We have applied the methods on a few test domains, some of them are synthetic while the others are based on well known problems including the search space of the 5-*peg Tower of Hanoi* problem, the *matrix-chain multiplication* problem and the problem of finding *secondary structure of RNA*. Experimental results show the efficacy of the proposed algorithms over the existing approach. Our proposed algorithms have potential use in various domains ranging from knowledge based frameworks to service composition, where the AND/OR structure is widely used for representing problems.

## 1. Introduction

The use of AND/OR structures for modeling and solving complex problems efficiently has attracted a significant amount of research effort over the last few decades. Initially, AND/OR search spaces were mostly used in problem reduction search for solving complex problems, logical reasoning and theorem proving, etc., where the overall problem can be hierarchically decomposed into conjunction and disjunction of subproblems (Pearl, 1984; Nilsson, 1980). Subsequently, AND/OR structures were also applied in a variety of domains, e.g., for representing assembly plans (Homem de Mello & Sanderson, 1990), generating VLSI floor-plans (Dasgupta, Sur-Kolay, & Bhattacharya, 1995), puzzle solving (Fuxi, Ming, & Yanxiang, 2003), etc. Traditionally the algorithm AO* (Pearl, 1984; Nilsson, 1980; Martelli & Montanari, 1978, 1973; Chang & Slagle, 1971) has been used for searching implicitly defined AND/OR structures. An empirical study of AO* can be found in Bonet and Geffner's (2005) work.

In the recent past there has been a renewed research interest towards the application of AND/OR structures. In various planning problems, including conditional planning to handle uncertainty, the AND/OR structure (Russell & Norvig, 2003) is a natural form for representation. The problem of generating solutions for such representations has been studied extensively (Hansen & Zilberstein, 2001; Jiménez & Torras, 2000; Chakrabarti, 1994). Dechter and Mateescu (2007) have presented the explicit AND/OR search space perspective for graphical models. Different search strategies (best first, branch and bound, etc.) over the AND/OR search spaces in graphical models are discussed by Marinescu and Dechter (2007b, 2006). AND/OR search spaces are also used for solving mixed integer linear programming (Marinescu & Dechter, 2005), 0/1 integer Programming (Marinescu & Dechter, 2007a), combinatorial optimization in graphical models (Marinescu & Dechter, 2009a, 2009b). AND/OR Multivalued Decision Diagrams (AOMDD), which combine the idea of Multi-Valued Decision Diagrams(MDD) and AND/OR structures, is presented by Mateescu, Dechter, and Marinescu (2008) and further research along this direction can be found in the work of Mateescu and Dechter (2008). AND/OR search spaces are also applied for solution sampling and counting (Gogate & Dechter, 2008). Smooth Deterministic Decomposable Negative Normal Forms (sd-DNNF) (Darwiche, 2001) exhibit explicit AND/OR DAG structure and have been used for various applications including compiling knowledge (Darwiche, 1999), estimating belief states (Elliott & Williams, 2006), etc.

Apart from the domains of planning, constraint satisfaction, knowledge based reasoning, etc., AND/OR structure based techniques are also widely used for various application based domains, e.g., web service composition (Gu, Xu, & Li, 2010; Shin, Jeon, & Lee, 2010; Gu, Li, & Xu, 2008; Ma, Dong, & He, 2008; Yan, Xu, & Gu, 2008; Lang & Su, 2005), vision and graphics tasks (Chen, Xu, Liu, & Zhu, 2006), etc. Lang and Su (2005) have described an AND/OR graph search algorithm for composing web services for user requirements. Ma et al. (2008) have advocated the use of AND/OR trees to capture dependencies between the inputs and outputs of the component web services and propose a top-down search algorithm to generate solutions of the AND/OR tree. Further research that uses AND/OR structures in the context of web service composition can be found in the works of Gu et al. (2010, 2008), Shin et al. (2010) and Yan et al. (2008). Chen et al. (2006) have applied explicit AND/OR structures for cloth modeling and recognition which is an important problem in vision and graphics tasks.

Such recent adoption of AND/OR search spaces for a wide variety of AI problems warrants further research towards developing suitable algorithms for searching AND/OR structures from different perspectives. In the general setting, the fundamental problem remains to find the minimum cost solution of AND/OR structures. For a given explicit AND/OR graph structure, the minimum cost solution is computed using either a top-down or a bottom-up approach. These approaches are based on the principle of dynamic programming and have complexity which is linear with respect to the size of the search space. Finding a minimum cost solution of an explicit AND/OR structure is a fundamental step for the approaches that use an implicit representation and systematically explore the search space. This is particularly the case for AO* (Nilsson, 1980) where the *potential solution graph (psg)* is recomputed every time from the current explicit graph after a node is expanded. In view of recent research where AND/OR structures are used and leveraged in a wide variety of problems ranging from planning domain to web service composition, the need for generating an ordered set of solutions of a given AND/OR structure becomes imminent. We briefly mention some areas where ordered solutions are useful.

Ordered set of solutions of an explicit AND/OR DAG can be used to develop useful variants of the AO* algorithm. Currently in AO*, only the minimum cost solution is computed whereas several variants of the A* algorithm exist, where solutions are often sought within a factor of cost of the optimal solution. These approaches (Ebendt & Drechsler, 2009; Pearl, 1984) were developed to adapt the A* algorithm for using inadmissible heuristics, leveraging multiple heuristics (Chakrabarti, Ghose, Pandey, & DeSarkar, 1989), generating solutions quickly within bounded sub-optimality, etc. Typically these techniques order the Open list using one evaluation function, and the next element for expansion is selected from an ordered subset of Open using some other criterion. Similar techniques can be developed for AO* search if ordered set of potential solutions are made available. That set can be used for node selection and expansion instead of expanding nodes only from the current best *psg*. This opens up an interesting area with significant research potential where the existing variations of the A* algorithm can be extended for AND/OR search spaces.

In the context of model based programming, the problem of finding ordered set of solutions has significant importance. Elliott (2007) has used valued sd-DNNFs to represent the problem and proposed an approach to generate $k$-best solutions. Since valued sd-DNNFs have an AND/OR structure, the proposed approach is possibly the earliest algorithm for generating ordered set of solutions of an AND/OR structure. The problem of finding ordered set of solutions for graphical models is studied by Flerova and Dechter (2011, 2010). However these techniques use alternative representations for the algorithm, where AND/OR search spaces can be constructed (Dechter & Mateescu, 2007) for graphical models. Recent research involving AOMDD based representation on weighted structures suggested future extensions towards generalizing Algebraic Decision Diagrams and introduces the notion of cost in AOMDDs. We envisage that ordered set of solutions finds useful applications in the context of research around AND/OR decision diagram based representation.

In the domain of service composition, the primary motivation behind providing a set of alternative solutions ordered by cost is to offer more choices, while trading off the specified cost criterion (to a limited extent) in favor of other 'unspecified' criteria (primarily from the standpoint of quality). Shiaa, Fladmark, and Thiell (2008) have presented an approach for generating a ranked set of solutions for the service composition problem. Typically the quality criteria are subjective in nature and difficult to express in terms of a single scalar cost function which is able to combine the cost/price and the quality aspects together. These aspects of quality are often encountered in the context of serving custom user requirements where the user prefers to minimize the cost/price of the solution while preserving his/her preferences. For example, for booking a holiday package for a specific destination, a travel service portal typically offers a list of packages with various combinations of attractions, hotel options and meal plans ordered by a single cost criterion, namely, the cost of the package. In general any product/solution that is composed of a number of components has a compositional flavor similar to service composition and it becomes important to present the user a set of alternative solutions ordered by cost so that he/she can select the best alternative according to his/her preferences.

Dynamic programming formulations typically have an underlying AND/OR DAG structure, which had been formally studied in the past (Martelli & Montanari, 1973). Besides classical problems like *matrix chain multiplication*, many other real world optimization problems offer *dynamic programming* formulations, where alternative solutions ordered by cost are useful in practice. One example of such a problem is finding the *secondary structure of RNA* (Mathews & Zuker, 2004) which is an important problem in Bioinformatics. RNAs may be viewed as sequences of bases belonging to the set {*Adenine(A), Cytocine(C), Guanine(G), Uracil(U)*}. RNA molecules tend to loop back and form base pairs with itself and the resulting shape is called the *secondary structure*. The primary factor that influences the secondary structure of RNA is the number of base pairings (higher number of base pairings generally implies more stable secondary structure). Under the well established rules for base pairings, the problem of maximizing the number of base pairings has an interesting dynamic programming formulation. However, apart from the number of base pairings, there are other factors that influence the stability, but these factors are typically evaluated experimentally. Therefore, for a given RNA sequence, it is useful to compute a pool of candidate secondary structures (in decreasing order of the number of base pairings) that may be subjected to further experimental evaluation in order to determine the most stable secondary structure.

The problem of generating ordered set of solutions is well studied in other domains. For discrete optimization problems, Lawler (1972) had proposed a general procedure for generating $k$-best solutions. A similar problem of finding $k$ most probable configurations in probabilistic expert systems is addressed by Nilsson (1998). Fromer and Globerson (2009) have addressed the problem of finding $k$ maximum probability assignments for probabilistic modeling using LP relaxation. In the context of ordinary graphs, Eppstein (1990) has studied the problem of finding $k$-smallest spanning trees. Subsequently, an algorithm for finding $k$-best shortest paths has been proposed in Eppstein's (1998) work. Hamacher and Queyranne (1985) have suggested an algorithm for $k$-best solutions to combinatorial optimization problems. Algorithms for generating $k$-best perfect matching are presented by Chegireddy and Hamacher (1987). Other researchers applied the $k$-shortest path problem to practical scenarios, such as, routing and transportation, and developed specific solutions (Takkala, Borndörfer, & Löbel, 2000; Subramanian, 1997; Topkis, 1988; Sugimoto & Katoh, 1985). However none of the approaches seems to be directly applicable for AND/OR structures. Recently some schemes related to ordered solutions to graphical models (Flerova & Dechter, 2011, 2010) and anytime AND/OR graph search (Otten & Dechter, 2011) have been proposed. Anytime algorithms for traditional OR search space (Hansen & Zhou, 2007) are well addressed by the research community.

In this paper, we address the problem of generating ordered set of solutions for explicit AND/OR DAG structure and present new algorithms. The existing method, proposed by Elliott (2007), works bottom-up by computing $k$-best solutions for the current node from the $k$-best solutions of its children nodes. We present a best first search algorithm, named *Alternative Solution Generation* (ASG) for generating ordered set of solutions. The proposed algorithm maintains a list of candidate solutions, initially containing only the optimal solution, and iteratively generates the next solution in non-decreasing order of cost by selecting the minimum cost solution from the list. In each iteration, this minimum cost solution is used to construct another set of candidate solutions, which is again added to the current list. We present two versions of the algorithm –

    a. Basic ASG (will be referred to as ASG henceforth) : This version of the algorithm may construct a particular candidate solution more than once;

> b. Lazy ASG or LASG : Another version of ASG algorithm that constructs every candidate solution only once.

In these algorithms, we use a compact representation, named *signature*, for storing the solutions. From the signature of a solution, the actual explicit form of that solution can be constructed through a top-down traversal of the given DAG. This representation allows the proposed algorithms to work in a top-down fashion starting from the initial optimal solution. Another salient feature of our proposed algorithms is that these algorithms work incrementally unlike the existing approach. Our proposed algorithms can be interrupted at any point of time during the execution and the set of ordered solutions obtained so far can be observed and subsequent solutions will be generated when the algorithms are resumed again. Moreover, if an upper limit estimate on the number of solutions required is known a priori, our algorithms can be further optimized using that estimate.

The rest of the paper is organised as follows. The necessary formalisms and definitions are presented in Section 2. In Section 3, we address the problem of generating ordered set of solutions for trees. Subsequently in Section 4, we address the problem of finding alternative solutions of explicit acyclic AND/OR DAGs in non-decreasing order of cost. We present two different solution semantics for AND/OR DAGs and discuss the existing approach as well as our proposed approach, along with a comparative analysis. Detailed experimental results, including the comparison of the performance of the proposed algorithms with the existing algorithm (Elliott, 2007), are presented in Section 5. We have used randomly constructed trees and DAGs as well as some well-known problem domains including the 5-*peg Tower of Hanoi* problem, the *matrix-chain multiplication* problem and the problem of finding the *secondary structure of RNA* as test domain. The time required and the memory used for generating a specific number of ordered solutions for different domains are reported in detail. In Section 6, we outline briefly about applying the proposed algorithms for implicitly specified AND/OR structures. Finally we present the concluding remarks in Section 7.

## 2. Definitions

In this section, we describe the terminology of AND/OR trees and DAGs followed by other definitions that are used in this paper. $G_{\alpha\beta} = \langle V, E \rangle$ is an AND/OR directed acyclic graph, where $V$ is the set of nodes and $E$ is the set of edges. Here $\alpha$ and $\beta$ in $G_{\alpha\beta}$ refer to the AND nodes and OR nodes in the DAG respectively. The direction of edges in $G_{\alpha\beta}$ is from the parent node to the child node. The nodes of $G_{\alpha\beta}$ with no successors are called *terminal* nodes. The *non-terminal* nodes of $G_{\alpha\beta}$ are of two types – *i*) OR nodes and *ii*) AND nodes . $V_\alpha$ and $V_\beta$ are the set of AND and OR nodes in $G_{\alpha\beta}$ respectively, and $n_{\alpha\beta} = |V|, n_\alpha = |V_\alpha|$, and $n_\beta = |V_\beta|$. The start (or root) node of $G_{\alpha\beta}$ is denoted by $v_R$. *OR edges* and *AND edges* are the edges that emanate from OR nodes and AND nodes respectively.

**Definition 2.a [Solution Graph]** A *solution graph*, $S(v_q)$, rooted at any node $v_q \in V$, is a finite sub-graph of $G_{\alpha\beta}$ defined as:

> a. $v_q$ is in $S(v_q)$;
> b. If $v'_q$ is an OR node in $G_{\alpha\beta}$ and $v'_q$ is in $S(v_q)$, then exactly one of its immediate successors in $G_{\alpha\beta}$ is in $S(v_q)$;
> c. If $v'_q$ is an AND node in $G_{\alpha\beta}$ and $v'_q$ is in $S(v_q)$, then all its immediate successors in $G_{\alpha\beta}$ are in $S(v_q)$;

d. Every maximal (directed) path in $S(v_q)$ ends in a terminal node;

e. No node other than $v_q$ or its successors in $G_{\alpha\beta}$ is in $S(v_q)$.

By a solution graph $S$ of $G_{\alpha\beta}$ we mean a solution graph with root $v_R$. □

**Definition 2.b [Cost of a Solution Graph]** In $G_{\alpha\beta}$, every edge $e_{qr} \in E$ from node $v_q$ to node $v_r$ has a finite non-negative cost $c_e(\langle v_q, v_r \rangle)$ or $c_e(e_{qr})$. Similarly every node $v_q$ has a finite non-negative cost denoted by $c_v(v_q)$. The cost of a solution $S$ is defined recursively as follows. For every node $v_q$ in $S$, the cost $C(S, v_q)$ is:

$$C(S, v_q) = \begin{cases} c_v(v_q), \text{if } v_q \text{ is a terminal node;} \\ c_v(v_q) + \big\{ C(S, v_r) + c_e(\langle v_q, v_r \rangle) \big\}, \text{ where } v_q \text{ is an OR node, and} \\ \qquad v_r \text{ is the successor of } v_q \text{ in } S; \\ c_v(v_q) + \sum \big\{ C(S, v_j) + c_e(\langle v_q, v_j \rangle) \big\}, \text{ where } 1 \le j \le k, \ v_q \text{ is an AND node} \\ \qquad \text{with degree } k, \text{ and } v_1, \dots, v_k \text{ are the immediate successors of } v_q \text{ in } S. \end{cases}$$

Therefore the cost of a solution $S$ is $C(S, v_R)$ which is also denoted by $C(S)$. We denote the optimal solution below every node $v_q$ as $opt(v_q)$. Therefore, the optimal solution of the entire AND/OR DAG $G_{\alpha\beta}$, denoted by $S_{opt}$, is $opt(v_R)$. The cost of the optimal solution rooted at every node $v_q$ in $G_{\alpha\beta}$ is $C_{opt}(v_q)$, which is defined recursively (for minimum cost objective functions) as follows:

$$C_{opt}(v_q) = \begin{cases} c_v(v_q), \text{ if } v_q \text{ is a terminal node;} \\ c_v(v_q) + min \big\{ C_{opt}(v_j) + c_e(\langle v_q, v_j \rangle) \big\}, \text{ where } 1 \le j \le k, \ v_q \text{ is an OR node} \\ \qquad \text{with degree } k, \text{ and } v_1, \dots, v_k \text{ are the immediate successors of } v_q \text{ in } G_{\alpha\beta}; \\ c_v(v_q) + \sum \big\{ C_{opt}(v_j) + c_e(\langle v_q, v_j \rangle) \big\}, \text{ where } 1 \le j \le k, \ v_q \text{ is an AND node} \\ \qquad \text{with degree } k, \text{ and } v_1, \dots, v_k \text{ are the immediate successors of } v_q \text{ in } G_{\alpha\beta}. \end{cases}$$

The cost of the optimal solution $S_{opt}$ of $G_{\alpha\beta}$ is denoted by $C_{opt}(v_R)$ or, alternatively, by $C_{opt}(S_{opt})$. When the objective function needs to be maximized, instead of the $min$ function, the $max$ function is used in the definition of $C_{opt}(v_q)$. □

It may be noted that it is possible to have more than one solution below an OR node $v_q$ to qualify to be the optimal one, i.e., when they have the same cost, and that cost is the minimum. Ties for the optimal solution below any such OR node $v_q$ are resolved arbitrarily and only one among the qualifying solutions (determined after tie-breaking) is marked as $opt(v_q)$.

An AND/OR tree, $T_{\alpha\beta} = \langle V, E \rangle$, is an AND/OR DAG and additionally satisfies the restrictions of a tree structure i.e., there can be at most one parent node for any node $v_q$ in $T_{\alpha\beta}$. In the context of AND/OR trees, we use $e_q$ to denote the edge that points to the vertex $v_q$. An *alternating* AND/OR tree, $\hat{T}_{\alpha\beta} = \langle V, E \rangle$, is an AND/OR tree with the restriction that there is an alternation between the AND nodes and the OR nodes. Every child of an AND node is either an OR node or a terminal node, and every children of an OR node is either an AND node or a terminal node. We use the term *solution tree* to denote the solutions of AND/OR trees.

We also discuss a different solution semantics, namely *tree based semantics*, for AND/OR DAGs. Every AND/OR DAG can be converted to an equivalent AND/OR tree by traversing the intermediate nodes in reverse topological order and replicating the subtree rooted at every node whenever the in-degree of the traversed node is more than 1. The details are shown in Procedure ConvertDAG. Suppose an AND/OR DAG $G_{\alpha\beta}$ is converted to an equivalent AND/OR tree $T_{\alpha\beta}$. We define the solutions of $T_{\alpha\beta}$ as the solutions of $G_{\alpha\beta}$ under *tree based semantics*.

---

**Procedure ConvertDAG($G_{\alpha\beta}$)**

    **input** : An AND/OR DAG $G_{\alpha\beta}$
    **output**: An equivalent AND/OR tree $T_{\alpha\beta}$

**1** Construct a list $M$, of non-terminal nodes of $G_{\alpha\beta}$, sorted in the reverse topological order;

**2** **while** $M$ *is not empty* **do**

**3**     $v_q \leftarrow$ Remove the first element of $M$;
       /* Suppose $E_{in}(v_q)$ is the list of incoming edges of $v_q$          */

**4**     **if** $InDegree(v_q) > 1$ **then**

**5**         **for** $i \leftarrow 2$ **to** $InDegree(v_q)$ **do**

**6**             $e_t \leftarrow E_{in}(v_q)[i]$;

**7**             Replicate the sub-tree rooted at $v_q$ with $v_q'$ as the root;

**8**             Modify the target node of $e_t$ from $v_q$ to $v_q'$;

**9**         **end**

**10**     **end**

**11** **end**

---

In this paper we use the solution semantics defined in Definition 2.a as the default semantics for the solutions of AND/OR DAGs. When the *tree based semantics* is used, it is explicitly mentioned.

## 2.1 Example

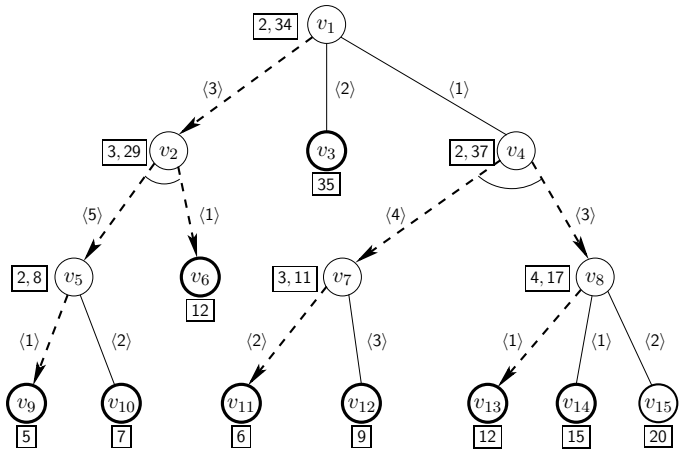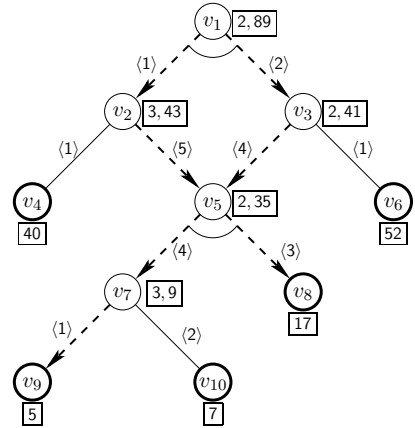

Figure 1: Alternating AND/OR Tree            Figure 2: AND/OR DAG

We present an example of an alternating AND/OR tree in Figure 1. In the figure, the terminal nodes are represented by a circle with thick outline. AND nodes are shown in the figures with their outgoing edges connected by a semi-circular curve in all the examples. The edge costs are shown by the side of each edge within an angled bracket. The cost of the terminal nodes are shown inside a box. For every non-terminal node $v_q$, the pair of costs, $c_v(v_q)$ and $C_{opt}(v_q)$, is shown inside a rectangle.

In Figure 1 the optimal solution below every node is shown using by thick dashed edges with an arrow head. The optimal solution of the AND/OR tree can be traced by following these thick dashed edges from node $v_1$. The cost of the optimal solution tree is 34. Also, Figure 2 shows an example of a DAG; the cost of the optimal solution DAG is 89.

## 3. Generating Ordered Solutions for AND/OR Trees

In this section we address the problem of generating ordered solutions for trees. We use the notion of alternating AND/OR trees, defined in Section 2, to present our algorithms. An alternating AND/OR tree presents a succinct representation and so the correctness proofs are much simpler for alternating AND/OR trees. In Appendix C we show that every AND/OR tree can be converted to an equivalent alternating AND/OR tree with respect to the solution space.

It is worth noting that the search space of some problems (e.g. the search space of multi-peg Tower of Hanoi problem) exhibit the alternating AND/OR tree structure. Moreover, the algorithms that are presented for alternating AND/OR trees work without any modification for general AND/OR trees. In this section, first we present the existing algorithm (Elliott, 2007) briefly, and then we present our proposed algorithms in detail.

### 3.1 Existing Bottom-Up Evaluation Based Method for Computing Alternative Solutions

We illustrate the working of the existing method that is proposed by Elliott (2007) for computing alternative solutions for trees using an example of an alternating AND/OR tree. This method (will be referred as $BU$ henceforth) computes the $k$-best solutions in a bottom-up fashion. At every node, $v_q$, $k$-best solutions are computed from the $k$-best solutions of the children of $v_q$. The overall idea is as follows.

    a. For an OR node $v_q$, a solution rooted at $v_q$ is obtained by selecting a solution of a child. Therefore $k$-best solutions of $v_q$ are computed by selecting the top $k$ solutions from the entire pool consisting of all solutions of all children.

    b. In the case of AND nodes, every child of an AND node $v_q$ will have at most $k$ solutions. A solution rooted at an AND node $v_q$ is obtained by combining one solution from every child of $v_q$. Different combinations of the solutions of the children nodes of $v_q$ generate different solutions rooted at $v_q$. Among those combinations, top $k$ combinations are stored for $v_q$.

In Figure 3 we show the working of the existing algorithm. At every intermediate node 2-best solutions are shown within rounded rectangle. At every OR node $v_q$, the $i^{th}$-best solution rooted at $v_q$ is shown as a triplet of the form – $\underbrace{i}$ : $\underbrace{< child, sol_{idx} >}$, $\underbrace{cost}$. For example, at node $v_1$ the second best solution is shown as $- \ 2 : \langle v_2, 2 \rangle, 37$; which means that the $2^{nd}$ best solution rooted at $v_1$ is obtained by selecting the $2^{nd}$ best solution of $v_2$. Similarly, at every AND node $v_q$, the $i^{th}$ solution rooted at $v_q$ is shown as a triplet of the form $- i : |sol\_vec|, cost$ triplets. Here $sol\_vec$ is a comma separated list of solution indices such that every element of $sol\_vec$ corresponds to a child of $v_q$. The $j^{th}$ element of $sol\_vec$ shows the index of the solution of $j^{th}$ child. For example, the $2^{nd}$ best solution rooted at $v_2$ is shown as $- 2 : |2, 1|, 32$. This means the $2^{nd}$ best solution rooted at $v_2$ is computed using the $2^{nd}$ best solution of the $1^{st}$ child (which is $v_5$) and the best solution ($1^{st}$) of the $2^{nd}$ child (which is $v_6$). Which index of $sol\_vec$ corresponds to which child is shown by placing the child node name above every index position.

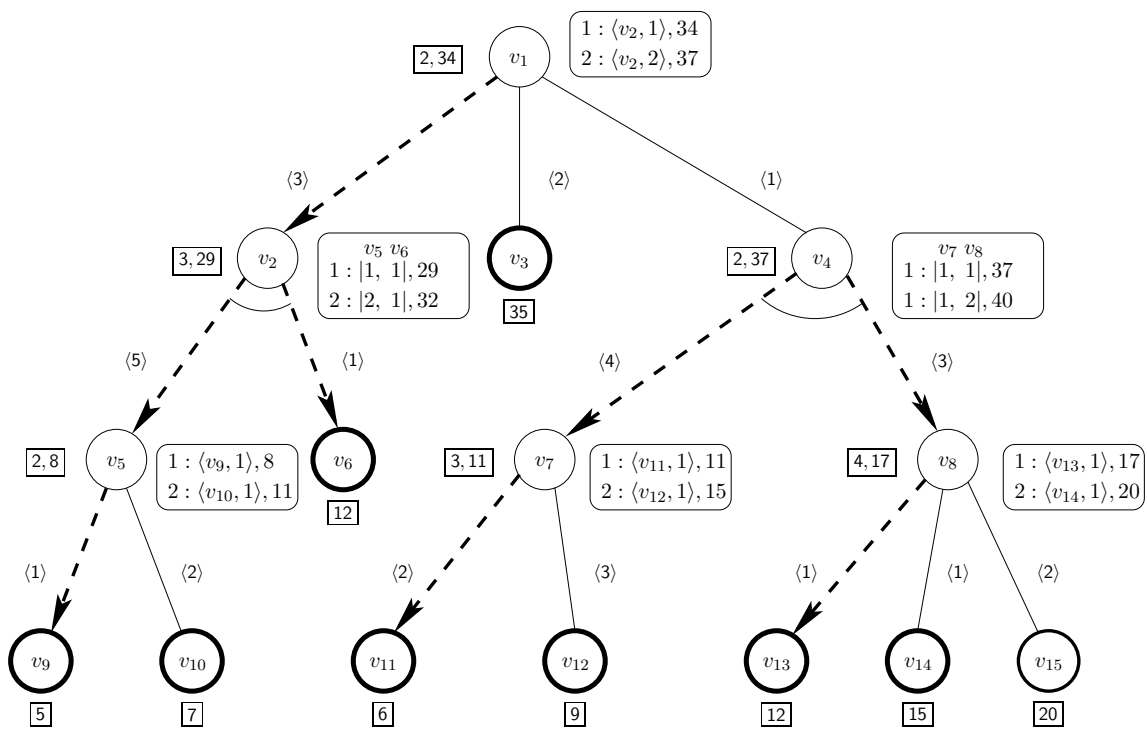

Figure 3: Example working of the existing algorithm

The existing method works with the input parameter $k$, i.e., the number of solutions to be generated have to be known a priori. Also this method is not inherently incremental in nature, thus does not perform efficiently when the solutions are needed on demand, e.g., at first, top 20 solutions are needed, then the next 10 solutions are needed. In this case the top 20 solutions will have to be recomputed while computing next 10 solutions, i.e., from the $21^{st}$ solution to the $30^{th}$ solution. Next we present our proposed top-down approach which does not suffer from this limitation.

## 3.2 Top-Down Evaluation Algorithms for Generating Ordered Solutions

So far we have discussed the existing approaches which primarily use bottom-up approach for computing ordered solutions. Now we propose a top-down approach for generating alternative solutions in the non-decreasing order of cost. It may be noted that the top-down approach is incremental in nature. We use an edge marking based algorithm, *Alternative Solution Generation* (*ASG*), to generate the next best solutions from the previously generated solutions. In the initial phase of the *ASG* algorithm, we compute the optimal solution for a given alternating AND/OR tree $\hat{T}_{\alpha\beta}$ and perform an initial marking of all *OR edges*. The following terminology and notions are used to describe the *ASG* algorithm. In the context of AND/OR trees, we use $e_q$ to denote the edge that points to the vertex $v_q$. We will use the following definitions for describing our proposed top-down approaches.

**Definition 3.c [Aggregated Cost]** In an AND/OR DAG $G_{\alpha\beta}$, the *aggregated cost*, $c_a$, for an edge $e_{ij}$ from node $v_i$ to node $v_j$, is defined as : $c_a(e_{ij}) = c_e(e_{ij}) + C_{opt}(v_j)$. □

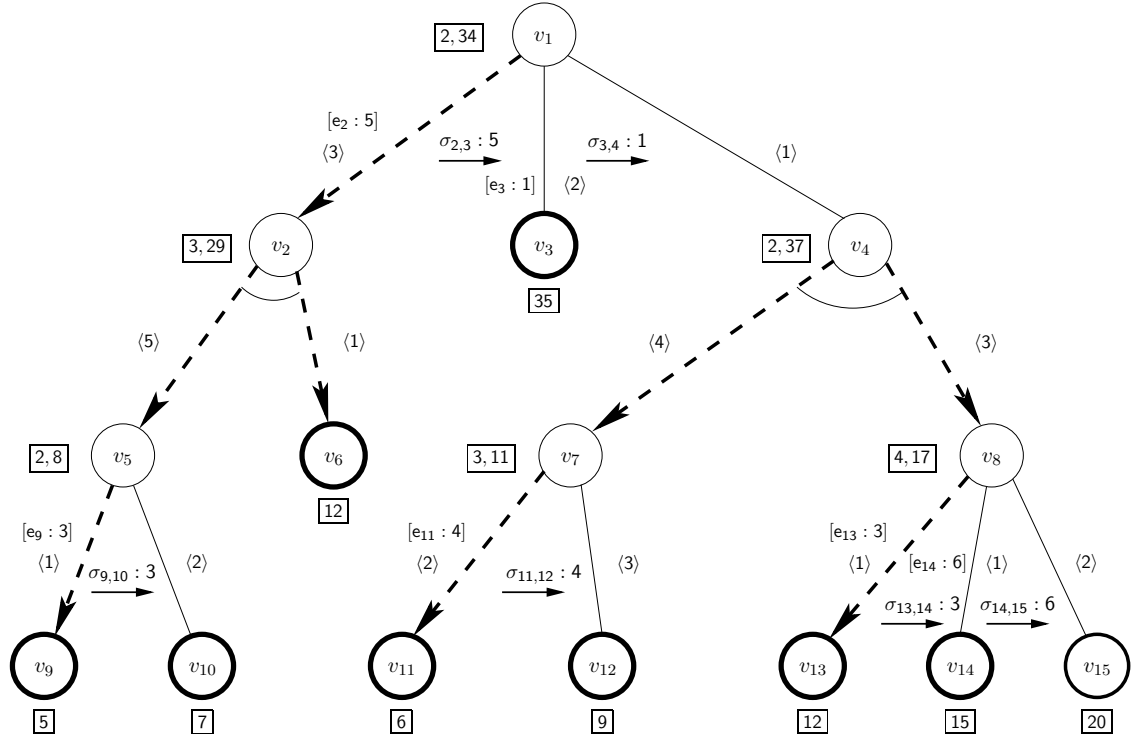

Figure 4: Example of OR-edge marking and swap option

**Marking of an OR edge :** The notion of marking an OR edge is as follows. For an OR node $v_q$, $L(v_q)$ is the list of OR edges of $v_q$ sorted in non-decreasing order of the *aggregated cost* of the edges. We define $\delta_{(i,i+1)}$ as the difference between the *cost* of OR edges, $e_i$ and $e_{i+1}$, such that $e_i$ and $e_{i+1}$ emanate from the same OR node $v_q$, and $e_{i+1}$ is the edge next to $e_i$ in $L(v_q)$. Procedure *MarkOR* describes the marking process for the OR edges of an OR node. Intuitively, a mark represents the cost increment incurred when the corresponding edge is replaced in a solution by its next best sibling. The OR edge having maximum aggregated cost is not marked.

Consider a solution, $S_{cur}$, containing the edge $e_i = (v_q, v_i)$, where $e_i \in E_{opt}(S_{cur})$. We mark $e_i$ with the cost increment which will be incurred to construct the next best solution from $S_{cur}$ by choosing another child of $v_q$. In Figure 4 the marks corresponding to OR edges $e_2$, $e_3$, $e_9$, $e_{11}$, $e_{13}$, and $e_{14}$ are $[e_2 : 5]$, $[e_3 : 1]$, $[e_9 : 3]$, $[e_{11} : 4]$, $[e_{13} : 3]$, and $[e_{14} : 6]$.

---

**Procedure** `MarkOR`($v_q$)

**1**    Construct $L(v_q)$ ;   /* List of OR edges of $v_q$ sorted in the non-decreasing order of $c_a$ values */

**2**    $count \leftarrow$ number of elements in $L(v_q)$ ;

**3**    **for** $i \leftarrow 1$ **to** $i = count - 1$ **do**

**4**        $e_c \leftarrow L(v_q)[i]$ ;

**5**        $e_n \leftarrow L(v_q)[i+1]$ ;

**6**        $\delta_{tmp} = (c_a(e_n) - c_a(e_c))$ ;

**7**        Mark $e_c$ with the pair $[e_n : \delta_{tmp}]$ ;

**8**    **end**

---

**Definition 3.d [Swap Option]** A *swap option* $\sigma_{ij}$ is defined as a three-tuple $\langle e_i, e_j, \delta_{ij} \rangle$ where $e_i$ and $e_j$ emanate from the same OR node $v_q$, $e_j$ is the edge next to $e_i$ in $L(v_q)$, and $\delta_{ij} = c_a(e_j) - c_a(e_i)$. Also, we say that the swap option $\sigma_{ij}$ *belongs* to the OR node $v_q$. □

Consider the OR node $v_q$ and the sorted list $L(v_q)$. It may be observed that in $L(v_q)$ every consecutive pair of edges forms a swap option. Therefore, if there are $k$ edges in $L(v_q)$, $k-1$ swap options will be formed. At node $v_q$, these swap options are *ranked* according to the rank of their *original edges* in $L(v_q)$. In Figure 4 the swap options are : $\sigma_{(2,3)} = \langle e_2, e_3, 5 \rangle$, $\sigma_{(3,4)} = \langle e_3, e_4, 1 \rangle$, $\sigma_{(9,10)} = \langle e_9, e_{10}, 3 \rangle$, $\sigma_{(11,12)} = \langle e_{11}, e_{12}, 4 \rangle$, $\sigma_{(13,14)} = \langle e_{13}, e_{14}, 3 \rangle$, and $\sigma_{(14,15)} = \langle e_{14}, e_{15}, 6 \rangle$. Consider the node $v_1$ where $L(v_1) = \langle e_2, e_3, e_4 \rangle$. Therefore, the swap options, $\sigma_{(2,3)}$ and $\sigma_{(3,4)}$, belong to $v_1$. At node $v_1$, the rank of $\sigma_{(2,3)}$ and $\sigma_{(3,4)}$ are 1 and 2 respectively.

**Definition 3.e [Swap Operation]** *Swap operation* is defined as the application of a swap option $\sigma_{ij} = \langle e_i, e_j, \delta_{ij} \rangle$ to a solution $S_m$ that contains the OR edge $e_i$ in the following way:

  a. Remove the subtree rooted at $v_i$ from $S_m$. Let the modified tree be $S'_m$. Edge $e_i$ is the *original edge* of $\sigma_{ij}$.

  b. Add the subtree $opt(v_j)$ to $S'_m$, which is constructed at the previous step. Let the newly constructed solution be $S''_m$. Edge $e_j$ is the *swapped edge* of $\sigma_{ij}$.

Intuitively, a swap operation $\sigma_{ij} = \langle e_i, e_j, \delta_{ij} \rangle$ constructs a new solution $S'_m$ from $S_m$ when $S_m$ contains the OR edge $e_i$. Moreover, the cost of $S'_m$ is increased by $\delta_{ij}$ compared to cost of $S_m$ if $C(S_m, v_i) = C_{opt}(v_i)$. □

Our proposed algorithms use a swap option based compact representation, named *signature*, for storing the solutions. Intuitively, any alternative solution can be described as a set of swap operations performed on the optimal solution $S_{opt}$. It is interesting to observe that while applying an ordered sequence of swap options, $\langle \sigma_1, \cdots, \sigma_k \rangle$, the application of each swap operation creates an intermediate alternative solution. For example, when the first swap option in the sequence, $\sigma_1$, is applied to the optimal solution, $S_{opt}$, a new solution, say $S_1$, is constructed. Then, when the $2^{nd}$ swap option, $\sigma_2$, is applied to $S_1$, yet another solution $S_2$ is constructed. Let $S_i$ denote the solution obtained by applying the swap options, $\sigma_1, \cdots, \sigma_i$, on $S_{opt}$ in this sequence. Although, an ordered sequence of swap options, like $\langle \sigma_1, \cdots, \sigma_k \rangle$, can itself be used as a compact representation of an alternative solution, the following key points are important to observe.

A. Among all possible sequences that generate a particular solution, we need to preclude those sequences which contain redundant swap options (those swap options whose orig-

285

inal edge is not present in the solution to which it is applied). This is formally defined later as *superfluous* swap options. Also the order of applying the swap options is another important aspect. There can be two swap options, $\sigma_i$ and $\sigma_j$ where $1 \leq i < j \leq k$ such that the source edge of $\sigma_j$ belongs to the sub-tree which is included in the solution $S_i$ only after applying $\sigma_i$ to $S_{i-1}$. In this case, if we apply $\sigma_j$ at the place of $\sigma_i$, i.e., apply $\sigma_j$ directly to $S_{i-1}$, it will have no effect as the source edge of $\sigma_j$ is not present in $S_{i-1}$, i.e., after swapping the location of $\sigma_i$ and $\sigma_j$ in the sequence, $\sigma_j$ becomes a redundant swap option and the solution constructed would be different for the swapped sequence from the original sequence. We formally define an order relation on a pair of swap options based on this observation in the later part of this section and formalize the compact representation of the solutions based on that order relation.

B. Suppose the swap option $\sigma_j$ belongs to a node $v_{p_j}$. Now it is important to observe that the application of $\sigma_j$ on $S_{j-1}$ to construct $S_j$, invalidates the application of all other swap options that belong to an OR edge in the path from the root node to $v_{p_j}$ in the solution $S_j$. This is because in $S_j$ the application of any such swap option which belongs to an OR edge in the path from the root node to $v_{p_j}$ would make the swap at $v_{p_j}$ redundant. In fact, for each swap option $\sigma_i$ belonging to node $v_{p_i}$, where $1 \leq i \leq j$, the application of all other swap options that belong to an OR edge in the path from the root node to $v_{p_i}$ is invalidated in the solution $S_j$ for the same reason. This condition restricts the set of swap options that can be applied on a particular solution.

C. Finally, there can be two swap options $\sigma_i$ and $\sigma_j$ for $1 \leq i < j \leq k$ such that $\sigma_i$ and $\sigma_j$ are independent of each other, that is, (a) applying $\sigma_i$ to $S_{j-1}$ and subsequently the application of $\sigma_j$ to $S_{j-1}$, and (b) applying $\sigma_j$ to $S_{i-1}$ and subsequently the application of $\sigma_i$ to $S_{j-1}$, ultimately construct the same solution. This happens only when the *original edges* of both $\sigma_i$ and $\sigma_j$ are present in $S_{i-1}$, thus application of one swap option does not influence the application of the other. However, it is desirable to use only one way to generate solution $S_j$. In Section 3.3, we propose a variation of the top-down approach (called LASG) which resolves this issue.

**Definition 3.f [Order Relation $\hat{\mathcal{R}}$]** We define an order relation, namely $\hat{\mathcal{R}}$, between a pair of swap options as follows.

a. If there is a path from $v_i$ to $v_r$ in $\hat{T}_{\alpha\beta}$, where $e_i$ and $e_r$ are OR edges, $\sigma_{qi}$ and $\sigma_{rj}$ are swap options, then $(\sigma_{qi}, \sigma_{rj}) \in \hat{\mathcal{R}}$. For example, in Figure 4 $(\sigma_{(3,4)}, \sigma_{(13,14)}) \in \hat{\mathcal{R}}$.

b. If $\sigma_{pq} = \langle e_p, e_q, \delta_{pq} \rangle$ and $\sigma_{rt} = \langle e_r, e_t, \delta_{rt} \rangle$ are two swap options such that $v_q = v_r$, then $(\sigma_{pq}, \sigma_{rt}) \in \mathcal{R}$. In Figure 4 $(\sigma_{(2,3)}, \sigma_{(3,4)}) \in \hat{\mathcal{R}}$. □

**Implicit Representation of the Solutions :** We use an *implicit* representation for storing every solution other than the optimal one. These other solutions can be constructed from the optimal solution by applying a set of swap options to the optimal solution in the following way. If $(\sigma_i, \sigma_j) \in \hat{\mathcal{R}}$, $\sigma_i$ has to be applied before $\sigma_j$. Therefore, every solution is represented as a sequence $\hat{\Sigma}$ of swap options, where $\sigma_i$ appears before $\sigma_j$ in $\hat{\Sigma}$ if $(\sigma_i, \sigma_j) \in \hat{\mathcal{R}}$. Intuitively the application of every swap option specifies that the swapped edge will be the part of the solution. Since the swap options are applied in the specific order $\hat{\mathcal{R}}$, it may so happen that an OR edge which had become the part of solution due to the application of an earlier swap option and may get swapped out due to the application of a later swap option.

**Definition 3.g [Superfluous Swap Option]** Consider a sequence of swap options $\hat{\Sigma} = \langle \sigma_1, \cdots, \sigma_m \rangle$ corresponding to a solution $S_m$. Clearly it is possible for a swap option, $\sigma_i$, where $1 \leq i \leq m$, to be present in the sequence such that the original edge of $\sigma_i$ is not present in the solution $S_{i-1}$ which is constructed by the successive applications of swap options $\sigma_1, \cdots, \sigma_{i-1}$ to solution $S_{opt}$. Now the application of $\sigma_i$ has no effect on $S_{i-1}$, i.e., solution $S_i$ is identical to solution $S_{i-1}$. Each such swap option $\sigma_i$ is a *superfluous* swap option with respect to the sequence $\hat{\Sigma}$ of swap options corresponding to solution $S_m$. ☐

**Property 3.1** *The sequence of swap options corresponding to a solution is* minimal, *if it has no superfluous swap option.*

This property follows from the definition of superfluous swap options and the notion of the implicit representation of a solution.

**Definition 3.h [Signature of a Solution]** The minimal sequence of swap options corresponding to a solution, $S_m$, is defined as the *signature*, $Sig(S_m)$, of that solution. It may be noted that for the optimal solution $S_{opt}$ of any alternating AND/OR tree $\hat{T}_{\alpha\beta}$, $Sig(S_{opt}) = \{\}$, i.e., an empty sequence. It is possible to construct more than one signature for a solution, as $\hat{\mathcal{R}}$ is a partial order. It is important to observe that all different signatures for a particular solution are of equal length and the sets of swap options corresponding to these different signatures are also equal. Therefore the set of swap options corresponding to a signature is a canonical representation of the signature. Henceforth we will use the set notation for describing the signature of a solution.

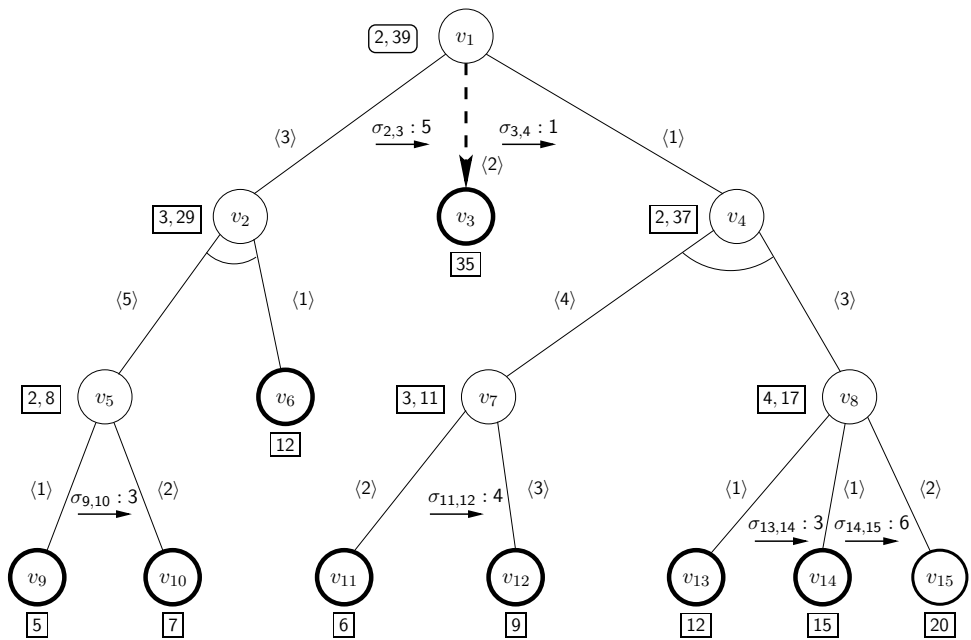

Figure 5: A solution, $S_2$, of the AND/OR tree shown in Figure 4

In Figure 5 we show a solution, say $S_2$, of the AND/OR tree shown in Figure 4. The solution is highlighted using thick dashed lines with arrow head. The pair, $c_v(v_q), C(S_2, v_q)$, is shown within rectangles beside each node $v_q$ in solution $S_2$, and we have used the rectangles with rounded corner whenever $C(S_2, v_q) \neq C_{opt}(v_q)$. Since $S_2$ is generated by applying the swap option $\sigma_{(2,3)}$ to solution $S_{opt}$, the signature of $S_2$, $Sig(S_2) = \langle \sigma_{(2,3)} \rangle$. Consider another sequence, $\hat{\Sigma}_2 = \langle \sigma_{(2,3)}, \sigma_{(9,10)} \rangle$, of swap options. It is worth noting that $\hat{\Sigma}_2$ also represents the solution $S_2$. Here the second swap option in $\hat{\Sigma}_2$, namely $\sigma_{9,10}$, can not be applied to the solution constructed by applying $\sigma_{(2,3)}$ to $S_{opt}$ as the source edge of $\sigma_{(9,10)}$, $e_9$, is not present in that solution. Hence $\sigma_{(9,10)}$ is a *superfluous swap option* for $\hat{\Sigma}_2$ .

**Definition 3.i [$V_{opt}$ and $E_{opt}$]** For any solution graph $S_m$ of an AND/OR DAG $G_{\alpha\beta}$, we define a set of nodes, $V_{opt}(S_m)$, and a set of OR edges, $E_{opt}(S_m)$, as:

    a. $V_{opt}(S_m) = \{ v_q \mid v_q$ in $S_m$ and solution graph $S_m(v_q)$ is identical to the solution graph $opt(v_q) \}$

    b. $E_{opt}(S_m) = \{ e_{pr} \mid$ OR edge $e_{pr}$ in $S_m$, and $v_r \in V_{opt}(S_m) \}$

Clearly, for any node $v_q \in V_{opt}(S_m)$, if $v_q$ is present in $S_{opt}$, then – (a) the solution graph $S_m(v_q)$ is identical to the solution graph $S_{opt}(v_q)$, and (b) $C(S_m, v_q) = C_{opt}(v_q)$     □

**Definition 3.j [Swap List]** The *swap list* corresponding to a solution $S_m$, $\mathcal{L}(S_m)$, is the list of swap options that are applicable to $S_m$. Let $Sig(S_m) = \{ \sigma_1, \cdots, \sigma_m \}$ and $\forall i, 1 \leq i \leq m$, each swap option $\sigma_i$ belongs to node $v_{p_i}$. The application of all other swap options that belong to the OR edges in the path from the root node to $v_{p_i}$ is invalidated in the solution $S_m$. Hence, only the remaining swap options that are not invalidated in $S_m$ can be applied to $S_m$ for constructing the successor solutions of $S_m$.

It is important to observe that for a swap option $\sigma_i$, if the *source edge* of $\sigma_i$ belongs to $E_{opt}(S_m)$, the application is not invalidated in $S_m$. Hence, for a solution $S_m$, we construct $\mathcal{L}(S_m)$ by restricting the swap operations only on the edges belonging to $E_{opt}(S_m)$. Moreover, this condition also ensures that the cost of a newly constructed solution can be computed directly form the cost of the parent solution and the $\delta$ value of the applied swap option. To elaborate, suppose solution $S'_m$ is constructed form $S_m$ by applying $\sigma_{jk}$. The cost of $S'_m$ can be computed directly form $C(S_m)$ and $\sigma_{jk}$ as : $C(S'_m) = C(S_m) + \delta_{jk}$ if $e_j \in E_{opt}(S_m)$. Procedure $ComputeSwapList(S_m)$ describes the details of computing swap options for a given solution $S_m$.     □

---

**Procedure** `ComputeSwapList(`$S_m$`)`

---

**1**    $\mathcal{L}(S_m) \leftarrow \varnothing$; Compute $E_{opt}(S_m)$;

**2**    **foreach** *OR edge $e_c$ in $E_{opt}(S_m)$* **do**

**3**        **if** *there exists a swap option on edge $e_c$* **then**

           /* Suppose $e_c$ emanates from OR node $v_q$ such that $e_c = L(v_q)[i]$. Also $e_c$ is marked with the pair $\langle \delta_{tmp}, e_n \rangle$, where $e_n = L(v_q)[i+1]$     */

**4**          $\sigma_{cn} \leftarrow \langle e_c, e_n, \delta_{tmp} \rangle$; Add $\sigma_{cn}$ to $\mathcal{L}(S_m)$;

**5**        **end**

**6**    **end**

---

The swap list of the optimal solution, $\mathcal{L}(S_{opt})$, in Figure 4, is $\{ \sigma_{(2,3)}, \sigma_{(9,10)} \}$. In the solution $S_1$, shown in Figure 6, $V_{opt} = \{ v_6, v_{10} \}$, because except node $v_6$ and $v_{10}$, for all other nodes $v_i$ in $S_1$, $opt(v_i) \neq S_1(v_i)$. Here also rectangles with rounded corner are used when $C(S_1, v_q) \neq C_{opt}(v_q)$. Therefore, $E_{opt} = \{ e_6, e_{10} \}$. Since there exists no swap option on the OR edges, $e_6$ and $e_{10}$, the swap list of solution $S_1$, $\mathcal{L}(S_1) = \varnothing$. Hence, for a solution $S_m$, $\mathcal{L}(S_m)$ may be empty, though $V_{opt}(S_m)$ can never be empty.

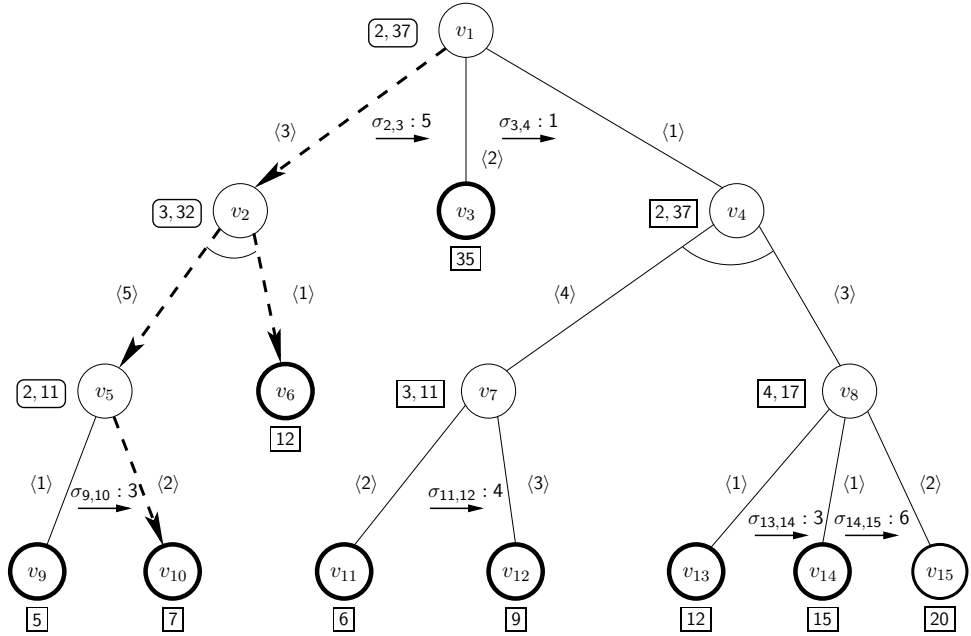

Figure 6: A solution, $S_1$, of the AND/OR tree shown in Figure 4

Although we use the notation $\sigma_{ij}$ to denote a swap option with edge $e_i$ as the *original edge* and edge $e_j$ as the *swapped edge*, for succinct representation, we also use $\sigma$ with a single subscript, such as $\sigma_3, \sigma_k, \sigma_{ij}$ etc., to represent a swap option. This alternative representation of swap options does not relate to any edge.

**Definition 3.k [Successors and Predecessors of a Solution]**  The set of successors and predecessors of a solution $S_m$ is defined as:

a. $Succ(S_m) = \{S'_m \mid S'_m$ can be constructed from $S_m$ by applying a swap option that belongs to the swap list of $S_m\}$

b. $Pred(S_m) = \{S''_m \mid S_m \in Succ(S''_m)\}$ □

**Property 3.2** *For any solution $S_m$ of an alternating AND/OR tree $\hat{T}_{\alpha\beta}$ the following statement holds:* $\forall S'_m \in Pred(S_m), C(S'_m) \leq C(S_m)$

The property follows from the definitions. One special case requires attention. Consider the case when $C(S'_m) = C(S_m)$ and $S'_m \in Pred(S_m)$. This case can only arise when a swap option of cost 0 is applied to $S_m$. This occurs in the case of a *tie*.

### 3.2.1 ASG Algorithm

We present ASG, a best first search algorithm, for generating solutions for an alternating AND/OR tree in non-decreasing order of costs. The overall idea of this algorithm is as follows. We maintain a list, **Open**, which initially contains only the optimal solution $S_{opt}$. At any point of time **Open** contains a set of candidate solutions from which the next best solution in the non-decreasing order of cost is selected. At each iteration the minimum cost solution ($S_{min}$) in Open is removed from Open and added to another list, named, Closed. The Closed list contains the set of ordered solutions generated so far. Then the successor set of $S_{min}$ is constructed and any successor solution which is not currently present in Open as well as is not already added to Closed is inserted to Open. However as a further optimization, we use a sublist of Closed, named TList, to store the relevant portion of Closed such that checking with respect to the solutions in TList is sufficient to figure out whether the successor solution is already added to Closed. It is interesting to observe that this algorithm can be interrupted at any time and the set of ordered solutions computed so far can be obtained. Also, the algorithm can be resumed if some more solutions are needed. The details of ASG algorithm are presented in Algorithm 4.

---

**Algorithm 4**: Alternative Solution Generation (ASG) Algorithm

**input** : An alternating AND/OR tree $\hat{T}_{\alpha\beta}$

**output**: Alternative solutions of $\hat{T}_{\alpha\beta}$ in the non-decreasing order of cost

1 Compute the optimal solution $S_{opt}$, perform OR edge marking and populate the swap options;

2 Create three lists, Open, Closed, and TList, that are initially empty;

3 Put $S_{opt}$ in Open;

4 $lastSolCost \leftarrow C(S_{opt})$;

5 **while** Open *is not empty* **do**

6    $S_{min} \leftarrow$ Remove the minimum cost solution from Open ;

7    **if** $lastSolCost < C(S_{min})$ **then**

8       Remove all the elements of TList;

9       $lastSolCost \leftarrow C(S_{min})$;

10    **end**

11    Add $S_{min}$ to Closed and TList;

12    Compute the swap list, $\mathcal{L}(S_{min})$, of $S_{min}$;

   /* Construct $Succ(S_{min})$ using $\mathcal{L}(S_{min})$ and add new solutions to Open       */

13    **foreach** $\sigma_{ij} \in \mathcal{L}(S_{min})$ **do**

14       Construct $S_m$ by applying $\sigma_{ij}$ to $S_{min}$;

15       Construct the signature of $S_m$, $Sig(S_m)$, by concatenating $\sigma_{ij}$ after $Sig(S_{min})$;

      /* Check whether $S_m$ is already present in Open or in TList       */

16       **if** ($S_m$ *not in* Open) *and* ($S_m$ *not in* TList) **then**

17          Add $S_m$ to Open;

18    **end**

19 **end**

20 Report the solutions in Closed;

---

The pseudo-code from Line-1 to Line-4 computes the optimal solution $S_{opt}$, performs the marking of OR edges, populates the swap options, and initializes Open, Closed and TList. The loop in Line-10 is responsible for generating a new solution every time it is executed as long as Open is not empty. In Line-6 of the ASG algorithm, the solution that is the current minimum cost solution in Open ($S_{min}$) is selected and removed from Open. The TList is populated and maintained from Line-7 to Line-10. The loop in Line-13 generates the successor solutions of $S_{min}$ one by one and adds the newly constructed solutions to Open if the newly constructed solution is not already present in Open as well as not added to TList (Line-16 does the checking). The proof of correctness of Algorithm 4 is presented in Appendix A. We discuss the following issues related to Algorithm 4.

**Checking for Duplication :** In order to check whether a particular solution $S_i$ is already present in Open or TList, the signature of $S_i$ is matched with the signatures of the solutions that are already present in Open and TList. It is sufficient to check the equality between the set of swap options in the respective signatures because that set is unique for a particular solution. It may be noted that TList is used as an optimization, which avoids searching the entire Closed list.

**Resolving Ties :** While removing the minimum cost solution from the Open list, a tie may be encountered among a set of solutions. Suppose there is a tie among the set $\mathcal{S}_{tie} = \{S_1, \cdots, S_k\}$. The ties are resolved in the favor of the predecessor solutions, that is,

$$\left(\forall S_i, S_j \in \mathcal{S}_{tie}\right), \ \left[\left(\text{If } S_i \text{ is the predecessor of } S_j\right) \Rightarrow \left(S_i \text{ is removed before } S_j\right)\right]$$

For all other cases the ties are resolved arbitrarily in the favor of the solution which was added to Open first.

### 3.2.2 Working of ASG Algorithm

We illustrate the working of the ASG algorithm on the example AND/OR tree shown in Figure 4. The contents of the different lists obtained after first few iterations of outermost *while* loop are shown in Table 1. We use the *signature* of a solution for representation purpose. The solutions that are already present in Open and also constructed by expanding the current $S_{min}$, are highlighted with under-braces.

| It. | $S_{min}$ | $\mathcal{L}(S_{min})$ | Open | Closed | TList |
|---|---|---|---|---|---|
| 1 | {} | $\sigma_{(2,3)}, \sigma_{(9,10)}$ | $\{\sigma_{(2,3)}\}, \{\sigma_{(9,10)}\}$ | {} | {} |
| 2 | $\{\sigma_{(9,10)}\}$ | $\varnothing$ | $\{\sigma_{(2,3)}\}$ | $\{\}, \{\sigma_{(9,10)}\}$ | $\{\sigma_{(9,10)}\}$ |
| 3 | $\{\sigma_{(2,3)}\}$ | $\sigma_{(3,4)}$ | $\sigma_{(2,3)}, \sigma_{(3,4)}$ | $\{\}, \{\sigma_{(9,10)}\}, \{\sigma_{(2,3)}\}$ | $\{\sigma_{(2,3)}\}$ |
| 4 | $\{\sigma_{(2,3)}, \sigma_{(3,4)}\}$ | $\sigma_{(11,12)}, \sigma_{(13,14)}$ | $\{\sigma_{(2,3)}, \sigma_{(3,4)}, \sigma_{(11,12)}\},$ $\{\sigma_{(2,3)}, \sigma_{(3,4)}, \sigma_{(13,14)}\}$ | $\{\}, \{\sigma_{(9,10)}\}, \{\sigma_{(2,3)}\},$ $\{\sigma_{(2,3)}, \sigma_{(3,4)}\}$ | $\{\sigma_{(2,3)}, \sigma_{(3,4)}\}$ |
| 5 | $\{\sigma_{(2,3)}, \sigma_{(3,4)},$ $\sigma_{(13,14)}\}$ | $\sigma_{(11,12)}, \sigma_{(14,15)}$ | $\{\sigma_{(2,3)}, \sigma_{(3,4)}, \sigma_{(11,12)}\},$ $\{\sigma_{(2,3)}, \sigma_{(3,4)}, \sigma_{(13,14)}, \sigma_{(11,12)}\}$ $\{\sigma_{(2,3)}, \sigma_{(3,4)}, \sigma_{(13,14)}, \sigma_{(14,15)}\}$ | $\{\}, \{\sigma_{(9,10)}\}, \{\sigma_{(2,3)}\},$ $\{\sigma_{(2,3)}, \sigma_{(3,4)}\}$ $\{\sigma_{(2,3)}, \sigma_{(3,4)}, \sigma_{(13,14)}\}$ | $\{\sigma_{(2,3)}, \sigma_{(3,4)},$ $\sigma_{(13,14)}\}$ |
| 6 | $\{\sigma_{(2,3)}, \sigma_{(3,4)},$ $\sigma_{(11,12)}\}$ | $\sigma_{(13,14)}$ | $\underbrace{\{\sigma_{(2,3)}, \sigma_{(3,4)}, \sigma_{(13,14)}, \sigma_{(11,12)}\}},$ $\{\sigma_{(2,3)}, \sigma_{(3,4)}, \sigma_{(13,14)}, \sigma_{(14,15)}\}$ | $\{\}, \{\sigma_{(9,10)}\}, \{\sigma_{(2,3)}\},$ $\{\sigma_{(2,3)}, \sigma_{(3,4)}\}$ $\{\sigma_{(2,3)}, \sigma_{(3,4)}, \sigma_{(13,14)}\}$ $\{\sigma_{(2,3)}, \sigma_{(3,4)}, \sigma_{(11,12)}\}$ | $\{\sigma_{(2,3)}, \sigma_{(3,4)},$ $\sigma_{(11,12)}\}$ |
| 7 | $\{\sigma_{(2,3)}, \sigma_{(3,4)},$ $\sigma_{(13,14)}, \sigma_{(11,12)}\}$ | $\sigma_{(14,15)}$ | $\{\sigma_{(2,3)}, \sigma_{(3,4)}, \sigma_{(13,14)}, \sigma_{(14,15)}\},$ $\{\sigma_{(2,3)}, \sigma_{(3,4)}, \sigma_{(13,14)},$ $\sigma_{(11,12)}, \sigma_{(14,15)}\}$ | $\{\}, \{\sigma_{(9,10)}\}, \{\sigma_{(2,3)}\},$ $\{\sigma_{(2,3)}, \sigma_{(3,4)}\}$ $\{\sigma_{(2,3)}, \sigma_{(3,4)}, \sigma_{(13,14)}\}$ $\{\sigma_{(2,3)}, \sigma_{(3,4)}, \sigma_{(11,12)}\}$ $\{\sigma_{(2,3)}, \sigma_{(3,4)}, \sigma_{(13,14)},$ $\sigma_{(11,12)}\}$ | $\{\sigma_{(2,3)}, \sigma_{(3,4)},$ $\sigma_{(13,14)}, \sigma_{(11,12)}\}$ |

Table 1: Working of ASG Algorithm

Before entering the outermost *while* loop (Line 5), ASG computes the optimal solution $S_{opt}$, populates the swap options, and inserts $S_{opt}$ to Open. Thus, at this point of time, Open contains only the optimal solution $S_{opt}$; Closed and TList are empty. In the first iteration $S_{opt}$ (the signature of $S_{opt}$ is {}) is selected and removed from Open. Then the swap list of $S_{opt}$, $\mathcal{L}(S_{opt})$, is computed. $\mathcal{L}(S_{opt})$, consists of two swap options, namely $\sigma_{(2,3)}$ and $\sigma_{(9,10)}$. ASG adds two new solutions $\{\sigma_{(2,3)}\}$ and $\{\sigma_{(9,10)}\}$ to Open. Then solution $S_{opt}$ is added to both Closed and TList.

In the next iteration, solution $\{\sigma_{(9,10)}\}$ which has the minimum cost among the solutions currently in Open, is selected and removed from Open, the swap list $\{\sigma_{(9,10)}\}$ is computed and subsequently $\{\sigma_{(9,10)}\}$ is added to Open and TList. As it happens, $\mathcal{L}(\{\sigma_{(9,10)}\}) = \varnothing$ (owing to the fact that $E_{opt} = \{e_6, e_{10}\}$ and there exists no swap option on the OR edges, $e_6$ and $e_{10}$), thus nothing else happens in this iteration. In the next iteration, solution $\{\sigma_{(2,3)}\}$ is removed from Open and ultimately solution $\{\sigma_{(2,3)}, \sigma_{(3,4)}\}$ is added to Open after adding $\{\sigma_{(2,3)}\}$ to Closed as well as to TList. Next two iterations proceed in a similar fashion. Now, consider the $6^{th}$ iteration. In this iteration, solution $\{\sigma_{(2,3)}, \sigma_{(3,4)}, \sigma_{(11,12)}\}$ is removed from Open, and its successor set has only one solution, $\{\sigma_{(2,3)}, \sigma_{(3,4)}, \sigma_{(11,12)}, \sigma_{(13,14)}\}$, which is already present in Open (inserted to Open in Iteration-5). Therefore, solution $\{\sigma_{(2,3)}, \sigma_{(3,4)}, \sigma_{(11,12)}, \sigma_{(13,14)}\}$ is not inserted to Open again. We have shown up to Iteration-7 in Table 1.

### 3.3 Technique for Avoiding the Checking for Duplicates in Open

In this section, we present a technique to avoid the checking done before adding a newly constructed solution $S_m$ to Open to determine whether $S_m$ is already present in Open. We first explain the scenario with an example, which is a portion of the previous example shown in Figure 4. In Figure 7-10, the solutions are shown using thick dashed line with arrow head. Also the rectangles with rounded corner are used to highlight the fact that the corresponding node in the marked solution does not belong to the $V_{opt}$ set of that solution.

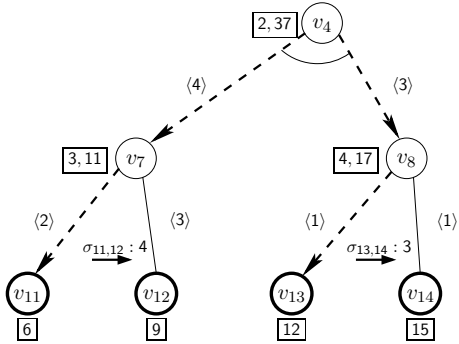 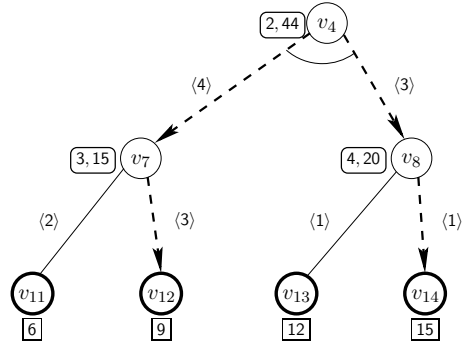

Figure 7: Running Example          Figure 8: Solution $S_3$

Consider the solutions $S_1$, $S_2$ and $S_3$ (shown in Figure 9, Figure 10 and Figure 8). Here
  (a) $\mathcal{L}(S_{opt}) = \{\sigma_{(11,12)}, \sigma_{(13,14)}\}$, (b) $Succ(S_{opt}) = \{S_1, S_2\}$,
  (c) $Sig(S_1) = \{\sigma_{(13,14)}\}$, (d) $Sig(S_2) = \{\sigma_{(11,12)}\}$, and (e) $Sig(S_3) = \{\sigma_{(13,14)}, \sigma_{(11,12)}\}$.
Algorithm 4 constructs the solution $S_3$ (shown in Figure 8) for adding to Open twice –
  (i) as a part of adding $Succ(S_1)$ to Open, and (ii) while adding $Succ(S_2)$ to Open.

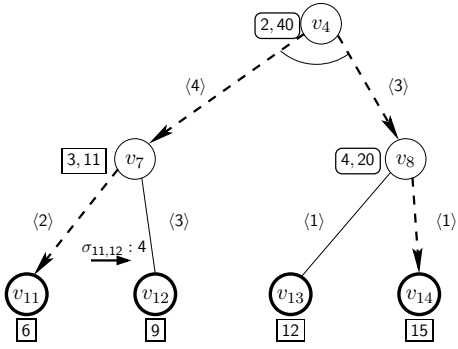

Figure 9: Solution $S_1$

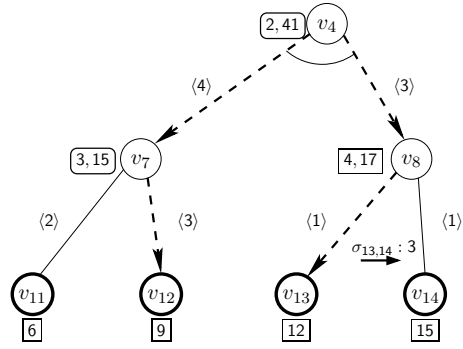

Figure 10: Solution $S_2$

We use the following definitions to describe another version of the ASG algorithm, which constructs the solutions in such a way that the check to find out whether a solution is already added to Open is avoided.

**Definition 3.l [Solution Space DAG(SSDAG)]** The *solution space DAG* of an alternating AND/OR tree $\hat{T}_{\alpha\beta}$ is a directed acyclic graph (DAG), $\mathcal{G}^s = \langle \mathcal{V}, \mathcal{E} \rangle$, where $\mathcal{V}$ is the set of all possible solutions of the AND/OR tree $\hat{T}_{\alpha\beta}$, and $\mathcal{E}$ is the set of edges which is defined as:

$$\mathcal{E} = \left\{ e_{pm}^s \ \middle| \ \begin{array}{l} S_p, S_m \in \mathcal{V}, \text{ and} \\ e_{pm}^s \text{ is a directed edge from node } S_p \text{ to } S_m, \text{ and} \\ S_m \in Succ(S_p) \end{array} \right\}$$

Clearly $S_{opt}$ is the root node of $\mathcal{G}^s$. $\qquad\qquad\square$

**Definition 3.m [Solution Space Tree and Completeness]** A *solution space tree* of an alternating AND/OR tree $\hat{T}_{\alpha\beta}$ is a tree $\mathcal{T}^s = \langle \mathcal{V}^t, \mathcal{E}^t \rangle$ where $\mathcal{V}^t \subseteq \mathcal{V}$, where $\mathcal{V}$ is the set of all possible solutions of the AND/OR tree $\hat{T}_{\alpha\beta}$, and $\mathcal{E}^t$ is the set of edges which is defined as:

$$\mathcal{E}^t = \left\{ e_{pm}^s \ \middle| \ \begin{array}{l} S_p, S_m \in \mathcal{V}^t, \text{ and} \\ e_{pm}^s \text{ is a directed edge from node } S_p \text{ to } S_m, \text{ and} \\ S_p \in Pred(S_m), \text{ and} \\ \forall S_p' \in Pred(S_m), \ \big((S_p \neq S_p') \Rightarrow \text{ there is no edge between } S_p' \text{ and } S_m\big). \end{array} \right\}$$

The *sibling set* for a solution $S_m$, is denoted using $Sib(\mathcal{T}^s, S_m)$. A solution space tree $\mathcal{T}^s$ for an AND/OR tree is *complete* if $\mathcal{V}^t = \mathcal{V}$. $\qquad\qquad\square$

It may be noted that the *complete solution space tree* of an alternating AND/OR tree is not necessarily unique. It is possible for an alternating AND/OR tree to have more than one complete solution space tree. However the solution space DAG for any AND/OR tree is unique.

**Definition 3.n [Native Swap Options of a Solution]** Consider a solution $S_m$ of an alternating AND/OR tree $\hat{T}_{\alpha\beta}$. Suppose $S_m$ is constructed by applying swap option $\sigma_{ij}$ to solution $S_p$. Since swap option $\sigma_{ij} = \langle e_i, e_j, \delta_{ij} \rangle$ is used to construct $S_m$, AND node $v_j$ is present in $S_m$. The native swap options of solution $S_m$ with respect to swap option $\sigma_{ij}$, $\mathcal{N}(S_m, \sigma_{ij})$, is a subset of $\mathcal{L}(S_m)$, and comprises of the following swap options :

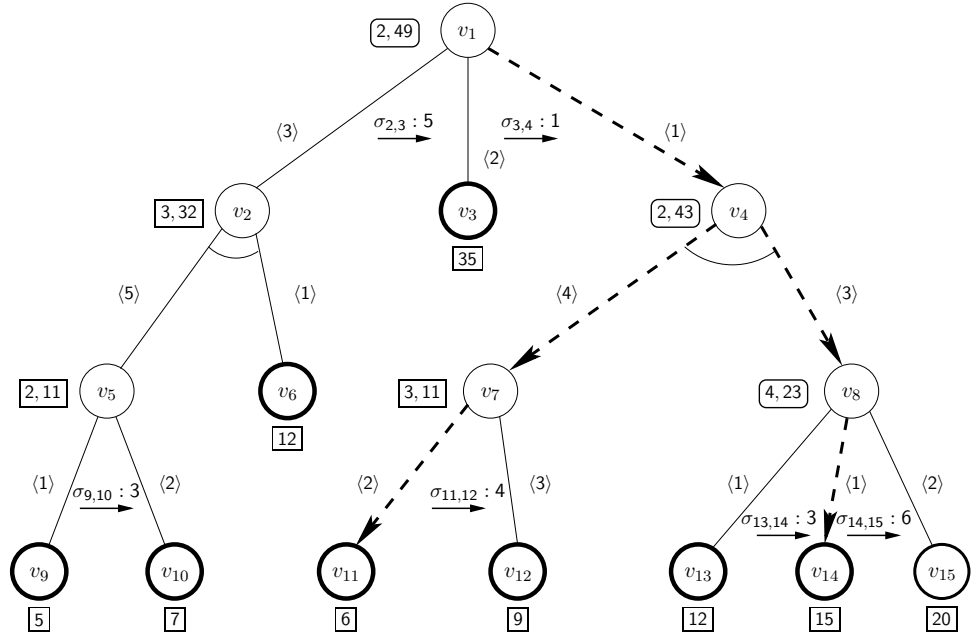

Figure 11: A solution, $S_4$, of the AND/OR tree shown in Figure 4

a. $\sigma_{jk}$, where $\sigma_{jk}$ is the swap option on the edge $e_j$

b. each $\sigma_t$, if $\sigma_t$ belongs to an OR node $v_q$ where $v_q$ is a node in $S_m(v_j)$

We use the term $\mathcal{N}(S_m)$ to denote the native swap options when $\sigma_{ij}$ is understood from the context. Intuitively the native swap options for solution $S_m$ are the swap options that become available immediately after applying $\sigma_{ij}$, but were not available in the predecessor solution of $S_m$. □

Consider the solution $S_4$ shown in Figure 11 where $Sig(S_4) = \{\sigma_{(2,3)}, \sigma_{(3,4)}, \sigma_{(13,14)}\}$. The solution is highlighted using thick dashed lines with arrow head. We have used the rectangles with rounded corner beside each node $v_q$ in solution $S_4$, where $C(S_4, v_q) \neq C_{opt}(v_q)$. Suppose $S_4$ is constructed form solution $S_3$ (where $Sig(S_3) = \{\sigma_{(2,3)}, \sigma_{(3,4)}\}$) using option $\sigma_{(13,14)}$. Here $\mathcal{N}(S_4, \sigma_{(13,14)}) = \{\sigma_{(14,15)}\}$ whereas $\mathcal{L}(S_4) = \{\sigma_{(11,12)}, \sigma_{(14,15)}\}$. Now consider solution $S_6$ where $Sig(S_6) = \{\sigma_{(2,3)}, \sigma_{(3,4)}, \sigma_{(11,12)}, \sigma_{(13,14)}\}$. It is worth observing that applying only the native swap options to $S_4$ instead of all swap options in $\mathcal{L}(S_4)$ prevents the construction of solution $S_6$ from solution $S_4$. $S_6$ can also be constructed by applying $\sigma_{(13,14)}$ to solution $S_5$, where $Sig(S_5) = \{\sigma_{(2,3)}, \sigma_{(3,4)}, \sigma_{(11,12)}\}$. However, it may be noted that $\sigma_{(13,14)}$ is not a native swap option of solution $S_5$.

### 3.3.1 Lazy ASG Algorithm

The intuition behind the other version of the ASG algorithm is as follows. For a newly constructed solution $S_m$, we need to check whether $S_m$ is already present in **Open** because $S_m$ can be constructed as a part of computing the successor set of multiple solutions. Instead of using the entire swap list of a solution to construct all successors at once and then add those solutions to **Open**, using the native swap options for constructing a subset of the successor set ensures the following. The subset constructed using native swap options consists of only those solutions that are currently not present in Open and thus can be added to Open without comparing with the existing entries in Open. The construction of each remaining successor solution $S'_m$ of $S_m$ and then insertion to Open is delayed until every other predecessor solution of $S'_m$ is added to Closed.

---

**Algorithm 5**: Lazy ASG (LASG) Algorithm

**input** : An alternating AND/OR tree $\hat{T}_{\alpha\beta}$

**output**: Alternative solutions of $\hat{T}_{\alpha\beta}$ in the non-decreasing order of cost

1   Compute the optimal solution $S_{opt}$, perform OR edge marking and populate the swap options;

2   Create two lists, Open and Closed, that are initially empty;

3   Put $S_{opt}$ in the Closed list;

4   Create a solution space tree $\mathcal{T}^s$ with $S_{opt}$ as root;

5   Compute the swap list, $\mathcal{L}(S_{opt})$, of $S_{opt}$;

6   Construct $Succ(S_{opt})$ using $\mathcal{L}(S_{opt})$;

7   **forall** $S_m \in Succ(S_{opt})$ **do**

8     Add $S_m$ to Open;

9   **end**

10   **while** Open *is not empty* **do**

11     $S_{min} \leftarrow$ Remove the minimum cost solution from Open ;

     /* Suppose $S_{min}$ is constructed from $S_m$ applying swap option $\sigma_{ij}$           */

12     Add a node corresponding to $S_{min}$ in $\mathcal{T}^s$ and connect that node using an edge from $S_m$ ;

13     Compute the swap list $\mathcal{L}(S_{min})$ and the list of native swap options $\mathcal{N}(S_{min}, \sigma_{ij})$;

     /* Expansion using native swap options                  */

14     **foreach** $\sigma_{tmp} \in \mathcal{N}(S_{min}, \sigma_{ij})$ **do**

15       Construct $S_{tmp}$ from $S_{min}$ by applying $\sigma_{tmp}$;

16       Construct the signature of $S_{tmp}$, $Sig(S_{tmp})$, by concatenating $\sigma_{tmp}$ after $Sig(S_{min})$;

17       Add $S_{tmp}$ to Open;

18     **end**

     /* Lazy Expansion                                  */

19     **forall** $S_p \in Sib(\mathcal{T}^s, S_{min})$ **do**

20       **if** $\sigma_{ij} \in \mathcal{L}(S_p)$ **then**

21         Construct $S'_p$ from $S_p$ using $\sigma_{ij}$;

22         Construct the signature of $S'_p$, $Sig(S'_p)$, by concatenating $\sigma_{ij}$ after $Sig(S_p)$;

23         Add $S'_p$ to Open;

24       **end**

25     **end**

26     Add $S_{min}$ to Closed;

27   **end**

28   Report the solutions in Closed;

---

The solution space tree $\mathcal{T}^s$ is maintained throughout the course of the algorithm to determine when every other predecessor of $S'_m$ is added to Closed. Based on this idea we present a *lazy* version of ASG algorithm, named LASG. After selecting the minimum cost solution from **Open**, the algorithm explores the successor set of the current minimum cost solution in a lazy fashion. For a solution $S_m$, at first a subset of $Succ(S_m)$ is constructed using only the native swap options of $S_m$. The other solutions that belong to $Succ(S_m)$ are explored as late as possible as described above. For resolving ties, LASG algorithm uses the same strategy which is used by ASG algorithm. The details of LASG algorithm are presented in Algorithm 5. The proof of correctness of this algorithm is presented in Appendix B.

Consider the example tree shown in Figure 7 and solutions $S_1$ and $S_2$ (shown in Figure 9 and Figure 10). Initially the **Open** will contain only $S_{opt}$ and $\mathcal{N}(S_{opt}) = \{\sigma_{(11,12)}, \sigma_{(13,14)}\}$. When $S_{opt}$ is selected from **Open**, both $S_1$ and $S_2$ is added to **Open**. Next $S_1$ will be selected followed by $S_2$. Since, $\mathcal{N}(S_1) = \varnothing$ and $\mathcal{N}(S_2) = \varnothing$, after selecting $S_1$ or $S_2$ no successor solutions are constructed using the native swap list. Among the predecessors of $S_3$, $S_2$ is added last to **Closed**. After selecting and removing $S_2$ from **Open**, solution $S_3$ is constructed from the previously selected predecessor $S_1$ using the swap option $\sigma_{(11,12)}$ which is used to construct solution $S_2$ from $S_{opt}$.

### 3.3.2 Working of LASG Algorithm (on AND/OR tree in Figure 4)

Before entering the outermost *while* loop (Algorithm 5, Line 10), LASG computes the optimal solution $S_{opt}$ and constructs $Succ(S_{opt})$. Then the solutions in $Succ(S_{opt})$ are added to **Open** and the contents of the **Open** becomes $\{\{\sigma_{(2,3)}\}, \{\sigma_{(9,10)}\}\}$. The contents of the different lists when a solution is added to **Closed** are shown in Table 2. The solutions are represented using their signatures. The solutions that are added to **Open** as a result of lazy expansion, are highlighted using under-brace.

| Iteration | $S_{min}$ | $\mathcal{N}(S_{min})$ | Open | Closed |
|---|---|---|---|---|
| - | {} | $\sigma_{(2,3)}, \sigma_{(9,10)}$ | $\{\sigma_{(2,3)}\}, \{\sigma_{(9,10)}\}$ | {} |
| 1 | $\{\sigma_{(9,10)}\}$ | $\varnothing$ | $\{\sigma_{(2,3)}\}$ | $\{\}, \{\sigma_{(9,10)}\}$ |
| 2 | $\{\sigma_{(2,3)}\}$ | $\sigma_{(3,4)}$ | $\{\sigma_{(2,3)}, \sigma_{(3,4)}\}$ | $\{\}, \{\sigma_{(9,10)}\}, \{\sigma_{(2,3)}\}$ |
| 3 | $\{\sigma_{(2,3)}, \sigma_{(3,4)}\}$ | $\sigma_{(11,12)}, \sigma_{(13,14)}$ | $\{\sigma_{(2,3)}, \sigma_{(3,4)}, \sigma_{(11,12)}\},$ $\{\sigma_{(2,3)}, \sigma_{(3,4)}, \sigma_{(13,14)}\}$ | $\{\}, \{\sigma_{(9,10)}\}, \{\sigma_{(2,3)}\},$ $\{\sigma_{(2,3)}, \sigma_{(3,4)}\}$ |
| 4 | $\{\sigma_{(2,3)}, \sigma_{(3,4)}, \sigma_{(13,14)}\}$ | $\sigma_{(14,15)}$ | $\{\sigma_{(2,3)}, \sigma_{(3,4)}, \sigma_{(11,12)}\},$ $\{\sigma_{(2,3)}, \sigma_{(3,4)}, \sigma_{(13,14)}, \sigma_{(14,15)}\}$ | $\{\}, \{\sigma_{(9,10)}\}, \{\sigma_{(2,3)}\},$ $\{\sigma_{(2,3)}, \sigma_{(3,4)}\}$ $\{\sigma_{(2,3)}, \sigma_{(3,4)}, \sigma_{(13,14)}\}$ |
| 5 | $\{\sigma_{(2,3)}, \sigma_{(3,4)}, \sigma_{(11,12)}\}$ | $\varnothing$ | $\{\sigma_{(2,3)}, \sigma_{(3,4)}, \sigma_{(13,14)}, \sigma_{(14,15)}\},$ $\underbrace{\{\sigma_{(2,3)}, \sigma_{(3,4)}, \sigma_{(13,14)}, \sigma_{(11,12)}\}}$ | $\{\}, \{\sigma_{(9,10)}\}, \{\sigma_{(2,3)}\},$ $\{\sigma_{(2,3)}, \sigma_{(3,4)}\}$ $\{\sigma_{(2,3)}, \sigma_{(3,4)}, \sigma_{(13,14)}\}$ $\{\sigma_{(2,3)}, \sigma_{(3,4)}, \sigma_{(11,12)}\}$ |
| 6 | $\{\sigma_{(2,3)}, \sigma_{(3,4)}, \sigma_{(13,14)}, \sigma_{(11,12)}\}$ | $\varnothing$ | $\{\sigma_{(2,3)}, \sigma_{(3,4)}, \sigma_{(13,14)}, \sigma_{(14,15)}\},$ | $\{\}, \{\sigma_{(9,10)}\}, \{\sigma_{(2,3)}\},$ $\{\sigma_{(2,3)}, \sigma_{(3,4)}\}$ $\{\sigma_{(2,3)}, \sigma_{(3,4)}, \sigma_{(13,14)}\}$ $\{\sigma_{(2,3)}, \sigma_{(3,4)}, \sigma_{(11,12)}\}$ $\{\sigma_{(2,3)}, \sigma_{(3,4)}, \sigma_{(13,14)}, \sigma_{(11,12)}\}$ |

Table 2: Working of LASG Algorithm

While generating the first four solutions, the contents of the different lists for LASG are identical to the contents of the corresponding lists of ASG (shown in Table 1). For each of these soltuions, the native swap list is equal to the actual swap list of that solution. It is worth noting that, unlike ASG, for LASG the outermost *while* loop starts after generating the optimal solution $S_{opt}$, thus while generating the same solution the iteration number for LASG is less than that of ASG by 1. In the $4^{th}$ iteration, for solution $S_4 = \{\sigma_{(2,3)}, \sigma_{(3,4)}, \sigma_{(13,14)}\}$ the native swap list is not equal to the swap list as described previously. The same holds true for solution $S_5 = \{\sigma_{(2,3)}, \sigma_{(3,4)}, \sigma_{(11,12)}\}$ and solution $S_6 = \{\sigma_{(2,3)}, \sigma_{(3,4)}, \sigma_{(13,14)}, \sigma_{(11,12)}\}$. It is important to observe that LASG adds the solution $S_6 = \{\sigma_{(2,3)}, \sigma_{(3,4)}, \sigma_{(13,14)}, \sigma_{(11,12)}\}$ to Open after the generation of solution $S_5 = \{\sigma_{(2,3)}, \sigma_{(3,4)}, \sigma_{(11,12)}\}$ as a part of *lazy expansion* (highlighted using under-brace in Table 2). Whereas, the ASG algorithm adds $S_6$ to Open after generating solution $S_4 = \{\sigma_{(2,3)}, \sigma_{(3,4)}, \sigma_{(13,14)}\}$.

## 3.4 Complexity Analysis and Comparison among ASG, LASG and BU

In this section we present a complexity analysis of ASG and LASG and compare them with BU. We will use the following parameters in the analysis.

    a. $n_{\alpha\beta}$ and $n_\beta$ denote the total number of nodes and the number of OR nodes in an alternating AND/OR tree.

    b. $\mathfrak{d}$ denotes the out degree of the OR node having maximum number of children.

    c. $\mathfrak{m}$ denotes the maximum number of OR edges in a solution.

    d. $\mathfrak{o}$ denotes the maximum size of Open. We will present the complexity analysis for generating $\mathfrak{c}$ solutions. Therefore the size of Closed is $O(\mathfrak{c})$.

### 3.4.1 Complexity of ASG

**Time Complexity :** The time complexity of the major steps of Algorithm 4 are as follows.

    a. Computing the first solution can be done in bottom-up fashion, thus requiring $O(n_{\alpha\beta})$ steps. The edges emanating from an OR node are sorted in the non-decreasing order of aggregated cost to compute the marks of the OR edges, the marking process takes $O(n_\beta.\mathfrak{d}.\log\mathfrak{d})$. Since the value of $\mathsf{d}$ is not very large in general (can be upper bounded by a constant), $O(n_\beta.\mathfrak{d}.\log\mathfrak{d}) = O(n_{\alpha\beta})$.

    b. The number of swap options available to a solution can be at most equal to the number of OR edges in that solution. Thus, the swap list for every solution can be built in $O(\mathfrak{m})$ time. For $\mathfrak{c}$ solutions, generating swap options take $O(\mathfrak{c}.\mathfrak{m})$.

    c. Since the size of the successor set of a solution can be $\mathfrak{m}$ at most, the size of Open, $\mathfrak{o}$ can at most be $\mathfrak{c}.\mathfrak{m}$. Also the size of the TList can at most be equal to $\mathfrak{c}$ (the size of Closed).

    d. The Open list can be implemented using Fibonacci heap. Individual *insert* and *delete* operation on Open take $O(1)$(amortized) and $O(\lg\mathfrak{o})$ time respectively. Hence, for inserting in the Open and deleting from Open altogether takes $O(\mathfrak{o}.\lg\mathfrak{o})$ time which is $O(\mathfrak{c}.\mathfrak{m}.\log(\mathfrak{c}.\mathfrak{m}))$.

    e. The *checking for duplicates* requires scanning the entire Open and TList. Since the length of TList can be at most $\mathfrak{c}$, for a newly constructed solution this checking takes $O(\mathfrak{c}+\mathfrak{o})$ time and at most $O(\mathfrak{c}+\mathfrak{o})$ solutions are generated. Since $O(\mathfrak{c}+\mathfrak{o})$ is actually $O(\mathfrak{o})$, for generating $\mathfrak{c}$ solutions, this step takes $O(\mathfrak{o})^2$ time. Also, the maximum value

of $\mathfrak{o}$ can be $O(\mathfrak{c}.\mathfrak{m})$. Thus, the time complexity of this step is $O(\mathfrak{c}.\mathfrak{m})^2$. Clearly this step dominates $O(\mathfrak{o}.\lg\mathfrak{o})$ which is the total time taken for all insertions into the Open and deletions from Open.

However, this time bound can be further improved if we maintain a *hash map* of the solutions in the Open and TList, and in this case the *checking for duplicates* can be done in $O(\mathfrak{o})$ time. In that case $O(\mathfrak{o}.\lg\mathfrak{o})$ (total time taken for all insertions into the Open and deletions from Open) becomes dominant over the time required for checking for duplicates.

f. An upper limit estimate of $\mathfrak{m}$ could be made by estimating the size of a solution tree which is $\sqrt{n_{\alpha\beta}}$ for regular and complete alternating AND/OR trees. It is important to observe that the value of $\mathfrak{m}$ is independent of the average out degree of a node in $\hat{T}_{\alpha\beta}$.

Combining the above factors together we get the time complexity of ASG algorithm as :
$$O\left(n_{\alpha\beta} + \mathfrak{o}^2\right) = O\left(n_{\alpha\beta} + (\mathfrak{c}.\mathfrak{m})^2\right) = O\left(n_{\alpha\beta} + \mathfrak{c}^2.n_{\alpha\beta}\right) = O(\mathfrak{c}^2.n_{\alpha\beta})$$
However if the additional hash map is used the time complexity is further reduced to :
$$O\left(n_{\alpha\beta} + \mathfrak{o}.\lg\mathfrak{o}\right) = O\left(n_{\alpha\beta} + \mathfrak{c}.\sqrt{n_{\alpha\beta}}.\lg(\mathfrak{c}.n_{\alpha\beta})\right) = O\left(n_{\alpha\beta} + \sqrt{n_{\alpha\beta}}.(\mathfrak{c}.\lg\mathfrak{c} + \mathfrak{c}.\lg n_{\alpha\beta})\right)$$

**Space Complexity:**  The following data-structures primarily contribute to the space complexity of ASG algorithm.

a. Three lists, namely, Open, Closed, and TList are maintained throughout the course of the running ASG. This contributes a $O(\mathfrak{o} + \mathfrak{c})$ factor, which is $O(\mathfrak{o})$.

b. Since the number of swap options is upper bounded by the total number of OR edges, constructing the swap list contributes the factor, $O(n_\beta.\mathfrak{d})$ to the space complexity. Also marking a solution requires putting a mark at every OR node of the AND/OR tree, thus adding another $O(n_\beta)$ space which is clearly dominated by the previous $O(n_\beta.\mathfrak{d})$ factor.

c. Since the signature of a solution is essentially a set of swap options, the size of a signature is upper bounded by the total number of swap options available. Combining the Open and Closed list, altogether $(\mathfrak{c} + \mathfrak{o})$ solutions need to be stored. Since $(\mathfrak{c} + \mathfrak{o})$ is $O(\mathfrak{o})$, total space required for storing the solutions is $O(\mathfrak{o}.n_\beta.\mathfrak{d})$ .

Combining the above factors together we get the space complexity of ASG algorithm as :
$$O\left(\mathfrak{o} + n_\beta.\mathfrak{d} + \mathfrak{o}.n_\beta.\mathfrak{d}\right) = O(\mathfrak{o}.n_\beta.\mathfrak{d})$$

When an additional hash map is used to improve the time complexity, another additional $O(\mathfrak{o}.n_\beta.\mathfrak{d})$ space is required for maintaining the hash map. Although the exact space requirement is doubled, asymptotically the space complexity remains same.

### 3.4.2 Complexity of LASG

**Time Complexity :**  Compared to Algorithm 4, Algorithm 5 does not check for the duplicates and adds the solution to Open only when it is required. Therefore the other terms in the complexity remain the same except the term corresponding to the checking for duplicates. However, here $\mathcal{T}^s$ is created and maintained during the course of Algorithm 5. Creating and maintaining the tree require $O(\mathfrak{c})$ time. Also during the *lazy expansion* the swap list of the previously generated sibling solutions are searched (Line 19 and Line 20 of Algorithm 5). The size of the swap list of any solution is $O(\mathfrak{m})$, where $\mathfrak{m}$ is the maximum number of OR edges in a solution. Also there can be at most $O(\mathfrak{m})$ sibling solutions for a solution. Therefore the complexity of the *lazy expansion* is $O(\mathfrak{c}.\mathfrak{m}^2)$. Since $O(\mathfrak{c}.\mathfrak{m}^2)$ is the dominant factor, the time complexity of LASG is $O(\mathfrak{c}.\mathfrak{m}^2) = O(\mathfrak{c}.n_{\alpha\beta})$.

**Space Complexity :** Compared to ASG algorithm, LASG algorithm does not maintain the TList. However LASG maintains the solution space tree $\mathcal{T}^s$ whose size is equal to the Closed list, thus adding another $O(\mathfrak{c})$ factor to the space complexity incurred by ASG algorithm. It is interesting to observe that the worst case space complexity remains $O(\mathfrak{o} + n_{\beta}.\mathfrak{d} + \mathfrak{o}.n_{\beta}.\mathfrak{d}) = O(\mathfrak{o}.n_{\beta}.\mathfrak{d})$ which is equal to the space complexity of ASG algorithm.

### 3.4.3 Comparison with BU

The time complexity of generating the $\mathfrak{c}$ best solutions for an AND/OR tree is $O(n_{\alpha\beta}.\mathfrak{c}.\log\mathfrak{c})$ and the space complexity is $O(n_{\alpha\beta}.\mathfrak{c})$. The detailed analysis can be found in the work of Elliott (2007). Since, $n_{\beta}.\mathfrak{d} = O(n_{\alpha\beta})$, the space complexity of both ASG and LASG algorithm reduces to $O(n_{\alpha\beta}.\mathfrak{c})$ and the time complexity of LASG is $\log\mathfrak{c}$ factor better than BU whereas the time complexity of ASG is quadratic with respect to $\mathfrak{c}$ compared to the $(\mathfrak{c}.\log\mathfrak{c})$ factor of BU. When an additional hash-map is used to reduce the time overhead of duplicate checking, ASG beats both LASG and BU both in terms time complexity, as both $O(n_{\alpha\beta})$ and $O(\sqrt{n_{\alpha\beta}}.(\mathfrak{c}.\lg\mathfrak{c} + \mathfrak{c}.\lg n_{\alpha\beta}))$ is asymptotically lower than $O(n_{\alpha\beta}.\mathfrak{c}.\log\mathfrak{c})$.

However this worst case complexity is only possible for AND/OR trees where no duplicate solution is generated. Empirical results show that the length of Open, $\mathfrak{o}$ hardly reaches $O(\mathfrak{c}.\mathfrak{m})$.

## 4. Ordered Solution Generation for AND/OR DAGs

In this section, we present the problem of generating solutions in non-decreasing order of cost for a given AND/OR DAG. We present the working of the existing algorithm for generating solution for both *tree based semantics* and *default semantics*. Next we present the modifications in ASG and LASG for handling DAG.

### 4.1 Existing Bottom-Up Algorithm

Figure 12 shows an example working of the existing bottom-up approach, BU, on the AND/OR DAG in Figure 2. We use the notations that are used in Figure 3 to describe different solutions in Figure 12 and the generation of the top 2 solutions under *tree-based semantics* is shown.

It is important to notice that although BU correctly generates alternative solutions of an AND/OR DAGs under *tree based semantics*, BU may generate some solutions which are invalid under *default semantics*. In Figure 13 we present a solution of the AND/OR DAG in Figure 2. This solution is an example of such a solution which is correct under *tree-based semantics* but is invalid under *default semantics*. The solution DAG (highlighted using thick dashed lines with arrow heads) in Figure 13 will be generated as the $3^{rd}$ solution of the AND/OR DAG in Figure 2 while running BU. At every non-terminal node, the entry (within rectangle) corresponding to the $3^{rd}$ solution is highlighted using **bold face**. It may be noted that the terminal nodes, $v_9$ and $v_{10}$, are included in the solution DAG though both of them emanate from the same parent OR node. Therefore, this solution is not a valid one under *default semantics*.

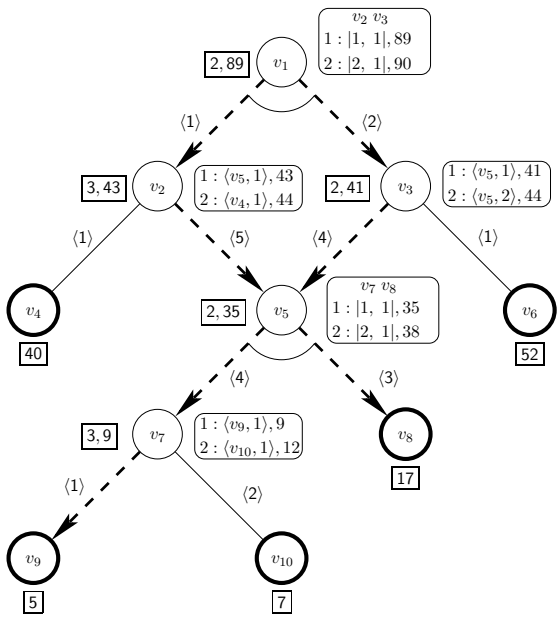

Figure 12: BU approach for AND/OR DAG

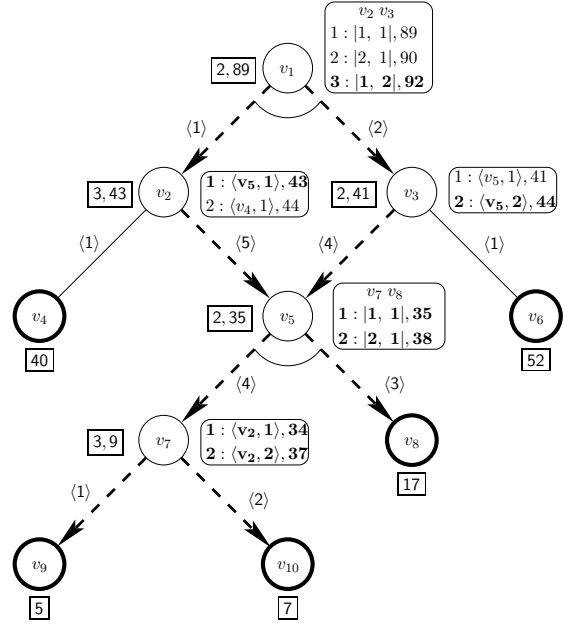

Figure 13: A solution (tree based semantics)

**Proposed Extension of BU to Generate Alternative Solutions under Default Semantics :**  We propose a simple top-down traversal and pruning based extension of BU to generate alternative solutions under *default semantics*. While generating the ordered solutions at any AND node $v_q$ by combining the solutions of the children, we do the following. For each newly constructed solution rooted at $v_q$, a top-down traversal of that solution starting from $v_q$ is done to check whether more than two edges of an OR node are present in that particular solution (a violation of the *default semantics*). If such a violation of the *default semantics* is detected, that solution is pruned from the list of alternative solutions rooted at $v_q$. Therefore, at every AND node, when a new solution is constructed, an additional top-down traversal is used to detect the semantics violation.

## 4.2 Top-Down Method for DAGs

The proposed top-down approaches (ASG and LASG) are also applicable for AND/OR DAGs to generate alternative solution DAGs under *default semantics*. Only the method of computing the cost increment after the application of a swap option needs to be modified to incorporate the fact that an OR node may be included in a solution DAG through multiple paths from the root node. We use the notion of *participation count* for computing the cost increment.

**Participation Count :**  The notion of *participation count* is applicable to the intermediate nodes of a solution DAG as follows. In a solution DAG, the *participation count* of an intermediate node, $v_q$, is the total number of distinct paths connecting the root node, $v_R$, and $v_q$. For example, in Figure 14, the optimal solution DAG is shown using thick dashed lines with arrow heads, and the *participation count* for every intermediate OR nodes are shown within a circle beside the node.

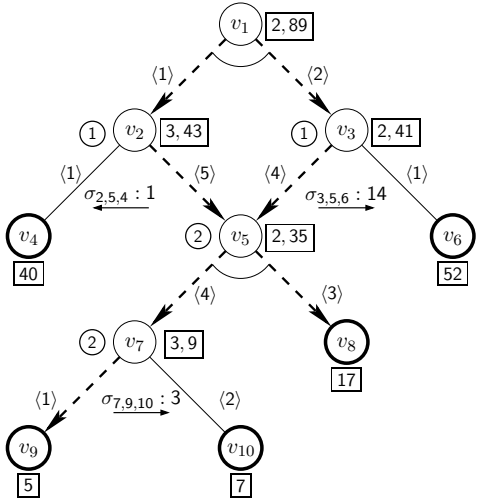

Figure 14: AND/OR DAG

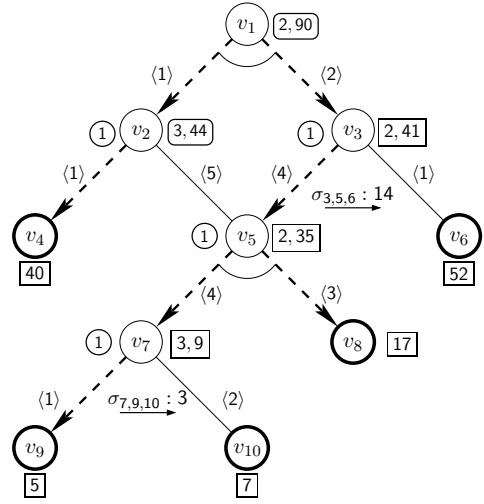

Figure 15: Solution DAG $S_1$

We use the notation $\sigma_{ijk}$ to denote a swap option in the context of AND/OR DAGs, where swap option $\sigma_{ijk}$ belongs to node $v_i$, the source edge of the swap option is $e_{ij}$ from node $v_i$ to node $v_j$, and the destination edge is $e_{ik}$ from node $v_i$ to node $v_k$.

### 4.2.1 Modification in the Proposed Top-Down Approach

The ASG algorithm is modified for handling AND/OR DAGs in the following way. The computation of the successor solution in Line 14 of Algorithm 4 is modified to incorporate the *participation count* of the OR node to which the applied swap option belongs. The overall method is shown in Algorithm 6(in the next page).

In order to apply LASG on AND/OR DAGs, apart from using the above mentioned modification for computing the cost of a newly generated solution, another modification is needed for computing the native swap options for a given solution. The modification is explained with an example. Consider the solution, $S_1$, shown in Figure 15. $S_1$ is highlighted using thick dashed lines with arrow heads. The pair, $c_v(v_q), C(S_1, v_q)$, is shown within rectangles beside each node $v_q$; rectangles with rounded corner are used when $C(S_1, v_q) \neq C_{opt}(v_q)$. Swap option $\sigma_{(2,5,4)}$ was applied to $S_{opt}$ to generate $S_1$. After the application of swap option $\sigma_{(2,5,4)}$, the participation count of node $v_5$ is decremented to 1. Therefore in $S_1$ there is a path from the root node to node $v_5$ and so node $v_5$ is still present in $S_1$. As a result, the swap option $\sigma_{(7,9,10)}$ is available to $S_1$ with a participation count equal to 1 for node $v_7$, whereas $\sigma_{(7,9,10)}$ is available to its parent solution $S_{opt}$ with participation count 2 for node $v_7$. In other words, $\sigma_{(7,9,10)}$ is not available to $S_1$ and its parent solution $S_{opt}$ with the same value of participation count for node $v_7$. Therefore $\sigma_{(7,9,10)}$ becomes the native swap option of $S_1$. The generalized definition of native swap options for a solution is presented below.

**Definition 4.o [Native Swap Options of a Solution]** Consider a solution $S_m$ of an AND/OR DAG $G_{\alpha\beta}$, where $S_m$ is constructed by applying swap option $\sigma_{hij}$ to solution $S_p$. Since swap option $\sigma_{hij} = \langle e_{hi}, e_{hj}, \delta_{hij} \rangle$ is used to construct $S_m$, AND node $v_j$ belongs to $S_m$. Similarly, if the participation count of node $v_i$ remains *greater than zero* after applying $\sigma_{hij}$ to $S_m$, node $v_i$ belongs to $S_m$. The native swap options of solution $S_m$ with respect to swap option $\sigma_{hij}$, $\mathcal{N}(S_m, \sigma_{hij})$, a subset of $\mathcal{L}(S_m)$, comprises of the following swap options :

    a. $\sigma_{hjk}$, where $\sigma_{hjk}$ is the swap option on the edge $e_{hj}$

    b. each $\sigma_t$, if $\sigma_t$ belongs to an OR node $v_q$ where $v_q$ is a node in $S_m(v_j)$

    c. each $\sigma_t'$, if node $v_i$ is present in $S_m$ and $\sigma_t'$ belongs to an OR node $v_q$ where $v_q$ is a node in $S_m(v_i)$.

We use the term $\mathcal{N}(S_m)$ to denote the native swap options when $\sigma_{hij}$ is understood from the context. Intuitively the native swap options for solution $S_m$ are the swap options that become available immediately after applying $\sigma_{hij}$, but were not available in the predecessor solution of $S_m$.     □

---

**Algorithm 6**: ASG Algorithm for AND/OR DAGs

    **input** : An AND/OR DAG $G_{\alpha\beta}$

    **output**: Alternative solutions of $G_{\alpha\beta}$ in the non-decreasing order of cost

**1** Compute the optimal solution $S_{opt}$, perform OR edge marking and populate the swap options;

**2** Create three lists, Open, Closed, and TList, that are initially empty;

**3** Put $S_{opt}$ in Open;

**4** $lastSolCost \leftarrow C(S_{opt})$;

**5** **while** Open *is not empty* **do**

**6**     $S_{min} \leftarrow$ Remove the minimum cost solution from Open;

**7**     **if** $lastSolCost < C(S_{min})$ **then**

**8**         Remove all the elements of TList;

**9**         $lastSolCost \leftarrow C(S_{min})$;

**10**     **end**

**11**     Add $S_{min}$ to Closed and TList;

**12**     Compute the swap list, $\mathcal{L}(S_{min})$, of $S_{min}$;

        /* Construct $Succ(S_{min})$ using $\mathcal{L}(S_{min})$ and add new solutions to Open    */

**13**     **foreach** $\sigma_{ij} \in \mathcal{L}(S_{min})$ **do**

**14**         Construct $S_m$ by applying $\sigma_{ij}$ to $S_{min}$;

**15**         Construct the signature of $S_m$, $Sig(S_m)$, by concatenating $\sigma_{ij}$ after $Sig(S_{min})$;

**16**         Let $\sigma_{ij}$ belongs to OR node $v_q$, $p$ is the participation count of $v_q$, and $\delta$ is the cost increment for $\sigma_{ij}$;

**17**         $C(S_m) = C(S_m) + p \times \delta$;

        /* Check whether $S_m$ is already present in Open or in TList    */

**18**         **if** ($S_m$ *not in* Open) *and* ($S_m$ *not in* TList) **then**

**19**             Add $S_m$ to Open;

**20**     **end**

**21** **end**

**22** Report the solutions in Closed;

---

It is worth noting that Definition 4.o of native swap option is a generalization of the earlier definition of native swap option (Definition 3.n), defined in the context of trees. In the case of trees, the participation count of any node can be at maximum 1. Therefore, after the application of a swap option to a solution, the participation count of the node, to which the original edge of the swap option points to, becomes 0. Therefore the third condition is never applicable for trees.

LASG (Algo. 5) can be applied on AND/OR DAGs, with the mentioned modification for computing the cost of a newly generated solution and the general definition of *native swap option* to generate ordered solutions under *default semantics*.

### 4.2.2 Working of ASG and LASG Algorithm on AND/OR DAG

We describe the working of ASG algorithm on the example DAG shown in Figure 2. Before entering the outermost *while* loop, TList and Closed are empty, and Open contains the optimal solution $S_{opt}$. The contents of the different lists obtained after first few cycles of outermost *while* loop are shown in Table 3. Each solution is represented by its *signature*. The solutions that are already present in Open and also constructed by expanding the current $S_{min}$, are highlighted with under-braces. For example, the solution $\{\sigma_{(2,5,4)}, \sigma_{(3,5,6)}\}$ which is added to Open in Iteration 2 (while constructing the successor solutions of $\{\sigma_{(2,5,4)}\}$) constructed again in Iteration 5 while expanding solution $\{\sigma_{(3,5,6)}\}$.

| It. | $S_{min}$ | $\mathcal{L}(S_{min})$ | Open | Closed |
|---|---|---|---|---|
| 1 | {} | $\sigma_{(2,5,4)}, \sigma_{(3,5,6)}, \sigma_{(7,9,10)}$ | $\{\sigma_{(2,5,4)}\}, \{\sigma_{(3,5,6)}\}, \{\sigma_{(7,9,10)}\}$ | {} |
| 2 | $\{\sigma_{(2,5,4)}\}$ | $\sigma_{(3,5,6)}, \sigma_{(7,9,10)}$ | $\{\sigma_{(3,5,6)}\}, \{\sigma_{(7,9,10)}\}, \{\sigma_{(2,5,4)}, \sigma_{(3,5,6)}\},$ $\{\sigma_{(2,5,4)}, \sigma_{(7,9,10)}\}$ | $\{\}, \{\sigma_{(2,5,4)}\}$ |
| 3 | $\{\sigma_{(2,5,4)}, \sigma_{(7,9,10)}\}$ | $\varnothing$ | $\{\sigma_{(3,5,6)}\}, \{\sigma_{(7,9,10)}\}, \{\sigma_{(2,5,4)}, \sigma_{(3,5,6)}\},$ | $\{\}, \{\sigma_{(2,5,4)}\},$ $\{\sigma_{(2,5,4)}, \sigma_{(7,9,10)}\}$ |
| 4 | $\{\sigma_{(7,9,10)}\}$ | $\varnothing$ | $\{\sigma_{(3,5,6)}\}, \{\sigma_{(2,5,4)}, \sigma_{(3,5,6)}\},$ | $\{\}, \{\sigma_{(2,5,4)}\},$ $\{\sigma_{(2,5,4)}, \sigma_{(7,9,10)}\},$ $\{\sigma_{(7,9,10)}\}$ |
| 5 | $\{\sigma_{(3,5,6)}\}$ | $\sigma_{(2,5,4)}, \sigma_{(7,9,10)}$ | $\underbrace{\{\sigma_{(2,5,4)}, \sigma_{(3,5,6)}\}}, \{\sigma_{(3,5,6)}, \sigma_{(7,9,10)}\}$ | $\{\}, \{\sigma_{(2,5,4)}\},$ $\{\sigma_{(2,5,4)}, \sigma_{(7,9,10)}\},$ $\{\sigma_{(7,9,10)}\}, \{\sigma_{(3,5,6)}\}$ |

Table 3: Example Working of ASG Algorithm on the DAG shown in Figure 2

Now we illustrate the working of LASG algorithm on the example DAG shown in Figure 2. The contents of the different lists when a solution is added to Closed are shown in Table 4. It is worth noting that for solution $S_1 = \{\sigma_{2,5,4}\}$, the swap list $\mathcal{L}(S_1) = \{\sigma_{(3,5,6)}, \sigma_{(7,9,10)}\}$ whereas the native swap list $\mathcal{N}(S_1) = \{\sigma_{(7,9,10)}\}$. The solutions that are added to Open as a result of lazy expansion, are highlighted using under-brace. For example, in Iteration 7 LASG adds the solution $S_5 = \{\sigma_{(2,5,4)}, \sigma_{(3,5,6)}\}$ to Open after the generation of solution $S_4 = \{\sigma_{3,5,6}\}$ as a part of *lazy expansion*, whereas the ASG algorithm adds $S_5$ to Open after generating solution $S_1 = \{\sigma_{2,5,4}\}$.

### 4.2.3 Generating Solutions under Tree Based Semantics

Unlike the *default semantics*, ASG or LASG does not have any straight forward extension for generating solutions under *tree based semantics*. In Figure 13 we show an example solution which is valid under *tree based semantics*, but invalid under *default semantics*, because both OR edges emanating form the OR node $v_7$, namely $e_{(7,9)}$ and $e_{(7,10)}$, are present in this solution. These two OR edges are included in the solution through two different paths emanating form the root node, $v_1$. As the existing bottom-up approach stores the alternative solutions at each node in terms of the solutions of the children of that node, this representation allows these different paths to be stored explicitly, thus making BU amenable for generating alternative solutions under *tree-based* semantics.

| It. | $S_{min}$ | $\mathcal{N}(S_{min})$ | Open | Closed |
|-----|-----------|------------------------|------|--------|
| - | {} | $\sigma_{(2,5,4)}, \sigma_{(3,5,6)}, \sigma_{(7,9,10)}$ | $\{\sigma_{(2,5,4)}\}, \{\sigma_{(3,5,6)}\}, \{\sigma_{(7,9,10)}\}$ | {} |
| 1 | $\{\sigma_{(2,5,4)}\}$ | $\sigma_{(7,9,10)}$ | $\{\sigma_{(3,5,6)}\}, \{\sigma_{(7,9,10)}\}, \{\sigma_{(3,5,4)}, \sigma_{(7,9,10)}\}$ | $\{\}, \{\sigma_{(2,5,4)}\}$ |
| 2 | $\{\sigma_{(2,5,4)}, \sigma_{(7,9,10)}\}$ | $\varnothing$ | $\{\sigma_{(3,5,6)}\}, \{\sigma_{(7,9,10)}\},$ | $\{\}, \{\sigma_{(2,5,4)}\}$ $\{\sigma_{(2,5,4)}, \sigma_{(7,9,10)}\}$ |
| 3 | $\{\sigma_{(7,9,10)}\}$ | $\varnothing$ | $\{\sigma_{(3,5,6)}\}$ | $\{\}, \{\sigma_{(2,5,4)}\},$ $\{\sigma_{(2,5,4)}, \sigma_{(7,9,10)}\},$ $\{\sigma_{(7,9,10)}\}$ |
| 4 | $\{\sigma_{(3,5,6)}\}$ | $\sigma_{(7,9,10)}$ | $\{\sigma_{(3,5,6)}, \sigma_{(7,9,10)}\},$ $\{\sigma_{(2,5,4)}, \sigma_{(3,5,6)}\}$ | $\{\}, \{\sigma_{(2,5,4)}\},$ $\{\sigma_{(2,5,4)}, \sigma_{(7,9,10)}\},$ $\{\sigma_{(7,9,10)}\}, \{\sigma_{(3,5,6)}\}$ |

Table 4: Example Working of LASG Algorithm on the DAG shown in Figure 2

On the contrary, our approach works top-down using a compact representation (signature) for storing the solutions. In this signature based representation, it is currently not possible to store the fact that a particular OR node is included in the solution through two different paths which may select different child of that OR node. If we use the equivalent tree constructed from the given graph, our compact representation will work correctly, because in that case, each node would be reachable from the root node through at most one path. An AND/OR DAG can be converted to its equivalent AND/OR tree representation using procedure ConvertDAG (described in Section 2) and then ASG or LASG can be applied on the equivalent tree representation in order to generate the alternative solutions correctly under *tree-based semantics*. However, in the worst case, procedure ConvertDAG incurs a space explosion which will blow up the worst case complexity of both ASG and LASG algorithms. Using our compact representations to generate the ordered solutions under *tree-based semantics* for a given AND/OR DAG while containing the space explosion such that the worst case complexity of our algorithms remain comparable with BU turns out to be an interesting open problem.

## 5. Experimental Results and Observations

To obtain an idea of the performance of the proposed algorithms and to compare with the existing approach, we have implemented the ASG, LASG and BU (existing bottom-up approach) and tested on the following test domains.

a. A set of synthetically generated AND/OR trees;
b. *Tower of Hanoi*(TOH) problem;
c. A set of synthetically generated AND/OR DAGs;
d. *Matrix-chain multiplication* problem; and
e. The problem of determining the *secondary structure of RNA* sequences.

It may noted that in our implementation of the ASG algorithm, we have implemented the more space efficient version of ASG algorithm (without a separate hash-map for storing the solutions in Open and Closed, thereby incurring an extra overhead in time for duplication checking). Another important point is that for every test case the reported running time of ASG and LASG for generating a particular number of solutions includes the time required for constructing the optimal solution graph. The details of the different test domains are as follows.

## 5.1 Complete Trees

We have generated a set of complete $d$-ary alternating AND/OR trees by varying – $(a)$ the degree of the non-terminal nodes (denoted by $d$), and $(b)$ the height (denoted by $h$).

| $(d, h)$ | 100 solutions | | | 300 solutions | | | 500 solutions | | |
|---|---|---|---|---|---|---|---|---|---|
| | ASG | LASG | BU | ASG | LASG | BU | ASG | LASG | BU |
| $(2, 7)$ | 0.027 | 0.005 | 0.004 | 0.086 | 0.014 | 0.009 | 0.186 | 0.023 | 0.020 |
| $(2, 9)$ | 0.216 | 0.010 | 0.015 | 1.448 | 0.035 | 0.046 | 4.137 | 0.060 | 0.097 |
| $(2, 11)$ | 1.170 | 0.031 | 0.068 | 10.098 | 0.094 | 0.184 | 27.354 | 0.216 | 0.407 |
| $(2, 13)$ | 6.072 | 0.124 | 0.257 | 57.757 | 0.348 | 0.777 | 158.520 | 0.524 | 1.641 |
| $(2, 15)$ | 30.434 | 0.517 | 1.180 | 278.453 | 1.433 | 3.917 | 766.201 | 2.806 | 7.257 |
| $(2, 17)$ | 130.746 | 2.265 | 4.952 | T | 6.443 | 13.277 | T | 10.306 | 29.703 |
| $(3, 5)$ | 0.046 | 0.006 | 0.005 | 0.196 | 0.015 | 0.018 | 0.459 | 0.026 | 0.042 |
| $(3, 7)$ | 0.528 | 0.017 | 0.037 | 4.764 | 0.060 | 0.153 | 10.345 | 0.088 | 0.457 |
| $(3, 9)$ | 5.812 | 0.106 | 0.343 | 55.170 | 0.290 | 1.733 | 156.158 | 0.494 | 4.913 |
| $(3, 11)$ | 66.313 | 1.552 | 3.973 | 620.996 | 3.712 | 14.323 | T | 6.607 | 33.923 |
| $(3, 13)$ | 636.822 | 12.363 | 31.043 | T | 34.150 | 128.314 | T | 55.510 | 303.785 |
| $(4, 5)$ | 0.144 | 0.011 | 0.033 | 1.041 | 0.025 | 0.092 | 2.610 | 0.042 | 0.123 |
| $(4, 7)$ | 2.916 | 0.056 | 0.573 | 25.341 | 0.181 | 1.561 | 69.596 | 0.264 | 2.107 |
| $(4, 9)$ | 58.756 | 1.266 | 7.698 | 544.989 | 3.327 | 27.063 | T | 5.172 | 38.606 |
| $(5, 5)$ | 0.334 | 0.012 | 0.081 | 2.792 | 0.036 | 0.400 | 7.374 | 0.062 | 0.930 |
| $(5, 7)$ | 12.227 | 0.177 | 2.066 | 102.577 | 0.443 | 11.717 | 283.689 | 0.827 | 26.994 |
| $(6, 5)$ | 0.699 | 0.022 | 0.161 | 5.384 | 0.071 | 1.418 | 15.133 | 0.134 | 2.235 |
| $(6, 7)$ | 32.620 | 0.654 | 7.464 | 288.257 | 1.566 | 37.758 | 832.235 | 2.594 | 90.465 |
| $(7, 5)$ | 1.306 | 0.030 | 0.287 | 12.006 | 0.092 | 1.833 | 29.870 | 0.179 | 4.322 |
| $(7, 7)$ | 81.197 | 1.786 | 15.892 | 785.160 | 4.284 | 102.431 | T | 6.890 | 241.064 |

Table 5: Comparison of running time (in seconds) for generating 100, 300, and 500 solutions for complete alternating AND/OR trees ($T$ denotes the timeout after 15 minutes)

These trees can be viewed as the search space for a *gift packing problem*, where
(a) the terminal nodes represent the cost of elementary items,
(b) the OR nodes model a choice among the items (elementary or composite in nature) represented by the children, and
(c) the AND nodes model the repackaging of the items returned by each of the children.
Every packaging incurs a cost which is modeled by the cost of the intermediate AND nodes. Here the objective is to find the alternative gifts in the order of non-decreasing cost.

Table 5 shows the time required for generating 100, 300, and 500 solutions for various complete alternating AND/OR trees. We have implemented the ASG, LASG and the existing bottom-up algorithm and the corresponding running time is shown in the column with the heading ASG, LASG and BU, respectively. We have used a time limit of 15 minutes and the entries marked with $T$ denotes that the time-out occurred for those test cases. The space required for generating 100, 300, and 500 solutions is reported in Table 6. It can be observed that in terms of both time and space required, LASG outperforms both ASG and BU. Between ASG and BU, for most of the test cases BU performs better than ASG with respect to the time required for generating a specific number of solutions. The space requirement of ASG and BU for generating a specific number of solutions has an interesting correlation with the $degree(d)$ and $height(h)$ parameter of the tree. For low numerical values of the $d$ and the $h$ parameter, e.g., $(d,h)$ combinations like $(2,7)$, $(3,5)$ etc., BU performs better than ASG. On the contrary, for the other combinations, where at least one of these $d$ and $h$ parameters has a high value, e.g., $(d,h)$ combinations like $(2,17)$, $(7,5)$, $(4,9)$ etc., ASG outperforms BU.

| $(d,h)$ | 100 solutions | | | 300 solutions | | | 500 solutions | | |
|---|---|---|---|---|---|---|---|---|---|
| | ASG | LASG | BU | ASG | LASG | BU | ASG | LASG | BU |
| $(2,7)$ | 12.633 | 13.168 | 11.047 | 28.105 | 32.293 | 14.266 | 41.676 | 49.832 | 16.609 |
| $(2,9)$ | 52.770 | 26.152 | 48.484 | 144.730 | 75.355 | 69.953 | 230.168 | 128.934 | 87.922 |
| $(2,11)$ | 116.582 | 63.254 | 198.234 | 341.227 | 165.824 | 292.703 | 566.766 | 269.766 | 373.172 |
| $(2,13)$ | 287.898 | 173.730 | 797.234 | 832.562 | 399.445 | 1183.703 | 1396.758 | 612.184 | 1514.172 |
| $(2,15)$ | 664.789 | 413.855 | 3193.234 | 1767.867 | 804.801 | 4747.703 | 2942.629 | 1197.266 | 6078.172 |
| $(2,17)$ | 1785.156 | 1257.387 | 12777.234 | T | 2047.859 | 19003.703 | T | 2849.617 | 24334.172 |
| $(3,5)$ | 17.270 | 17.258 | 11.688 | 47.531 | 49.230 | 14.812 | 76.270 | 80.980 | 17.938 |
| $(3,7)$ | 82.609 | 48.086 | 111.438 | 235.855 | 134.102 | 152.062 | 393.113 | 219.555 | 192.688 |
| $(3,9)$ | 335.301 | 184.375 | 1009.188 | 926.004 | 376.766 | 1387.312 | 1507.973 | 577.766 | 1765.438 |
| $(3,11)$ | 1474.477 | 1071.352 | 9088.938 | 3234.523 | 1656.844 | 12504.562 | T | 2238.152 | 15920.188 |
| $(3,13)$ | 9139.312 | 7872.055 | 81806.688 | T | 9565.598 | 112559.812 | T | 11251.035 | 143312.938 |
| $(4,5)$ | 40.285 | 24.469 | 47.453 | 121.336 | 67.102 | 112.609 | 199.254 | 116.535 | 129.016 |
| $(4,7)$ | 213.816 | 128.629 | 767.453 | 559.734 | 284.922 | 1826.359 | 917.824 | 451.223 | 2105.266 |
| $(4,9)$ | 1563.770 | 1158.582 | 12287.453 | 3209.145 | 1699.191 | 29246.359 | T | 2240.012 | 33725.266 |
| $(5,5)$ | 64.879 | 40.355 | 88.281 | 182.270 | 110.480 | 225.781 | 305.891 | 179.801 | 363.281 |
| $(5,7)$ | 529.738 | 343.254 | 2217.188 | 1254.715 | 596.957 | 5675.000 | 2008.344 | 858.852 | 9132.812 |
| $(6,5)$ | 97.703 | 58.191 | 151.047 | 270.027 | 148.453 | 372.141 | 443.656 | 245.227 | 593.234 |
| $(6,7)$ | 1264.828 | 862.332 | 5449.797 | 2747.238 | 1273.641 | 13433.391 | 4203.957 | 1695.684 | 21416.984 |
| $(7,5)$ | 137.527 | 90.703 | 242.219 | 369.086 | 205.914 | 576.594 | 606.133 | 317.492 | 910.969 |
| $(7,7)$ | 2628.461 | 1995.195 | 11882.781 | 4869.551 | 2627.211 | 28295.281 | T | 3273.703 | 44707.781 |

Table 6: Comparison of space required (in KB) for generating 100, 300, and 500 solutions for complete alternating AND/OR trees

### 5.1.1 Experimentation with Queue with Bounded Length

Since the Open can grow very rapidly, both ASG and LASG incur a significant overhead in terms of time as well as space to maintain the Open list when the number of solutions to be generated is not known a priori. In fact, for ASG checking for duplicates in Open is actually the primary source of time complexity and storing the solutions in Open is a major contributing factor in space complexity. If the number of solutions that have to generated is known a priori, the proposed top-down approach can leverage the fact by using a bounded length queue for implementing Open. When a bounded length queue is used, the time requirement along with space requirement decreases significantly.

| (d, h) | 100 solutions | | | 300 solutions | | | 500 solutions | | |
|--------|------|------|------|------|------|------|------|------|------|
| | ASG | LASG | BU | ASG | LASG | BU | ASG | LASG | BU |
| (2, 7) | 0.011 | 0.008 | 0.004 | 0.003 | 0.002 | 0.009 | 0.005 | 0.004 | 0.020 |
| (2, 9) | 0.030 | 0.011 | 0.015 | 0.008 | 0.006 | 0.046 | 0.014 | 0.008 | 0.097 |
| (2, 11) | 0.051 | 0.031 | 0.068 | 0.020 | 0.011 | 0.184 | 0.023 | 0.017 | 0.407 |
| (2, 13) | 0.125 | 0.103 | 0.257 | 0.043 | 0.059 | 0.777 | 0.065 | 0.058 | 1.641 |
| (2, 15) | 0.473 | 0.421 | 1.180 | 0.168 | 0.164 | 3.917 | 0.254 | 0.346 | 7.257 |
| (2, 17) | 2.129 | 2.199 | 4.952 | 0.766 | 1.005 | 13.277 | 1.146 | 1.492 | 29.703 |
| (3, 5) | 0.012 | 0.009 | 0.005 | 0.003 | 0.002 | 0.018 | 0.005 | 0.004 | 0.042 |
| (3, 7) | 0.031 | 0.018 | 0.037 | 0.012 | 0.006 | 0.153 | 0.019 | 0.010 | 0.457 |
| (3, 9) | 0.133 | 0.102 | 0.343 | 0.048 | 0.043 | 1.733 | 0.071 | 0.061 | 4.913 |
| (3, 11) | 1.246 | 1.143 | 3.973 | 0.477 | 0.636 | 14.323 | 0.693 | 0.905 | 33.923 |
| (3, 13) | 10.713 | 10.313 | 31.043 | 4.160 | 5.555 | 128.314 | 6.013 | 7.890 | 303.785 |
| (4, 5) | 0.019 | 0.008 | 0.033 | 0.006 | 0.004 | 0.092 | 0.010 | 0.006 | 0.123 |
| (4, 7) | 0.071 | 0.055 | 0.573 | 0.026 | 0.023 | 1.561 | 0.038 | 0.033 | 2.107 |
| (4, 9) | 1.099 | 0.998 | 7.698 | 0.443 | 0.552 | 27.063 | 0.641 | 0.808 | 38.606 |
| (5, 5) | 0.025 | 0.013 | 0.081 | 0.009 | 0.031 | 0.400 | 0.015 | 0.008 | 0.930 |
| (5, 7) | 0.201 | 0.161 | 2.066 | 0.083 | 0.078 | 11.717 | 0.116 | 0.153 | 26.994 |
| (6, 5) | 0.036 | 0.018 | 0.161 | 0.014 | 0.011 | 1.418 | 0.021 | 0.010 | 2.235 |
| (6, 7) | 0.543 | 0.460 | 7.464 | 0.240 | 0.325 | 37.758 | 0.326 | 0.431 | 90.465 |
| (7, 5) | 0.042 | 0.029 | 0.287 | 0.020 | 0.013 | 1.833 | 0.025 | 0.022 | 4.322 |
| (7, 7) | 1.940 | 1.705 | 15.892 | 0.807 | 0.843 | 102.431 | 0.870 | 1.125 | 241.064 |

Table 7: Comparison of running time (in seconds) for generating 100, 300, and 500 solutions for complete alternating AND/OR trees with bounded length **Open** queue for ASG and LASG

| (d, h) | 100 solutions | | | 300 solutions | | | 500 solutions | | |
|--------|------|------|------|------|------|------|------|------|------|
| | ASG | LASG | BU | ASG | LASG | BU | ASG | LASG | BU |
| (2, 7) | 10.109 | 2.383 | 11.047 | 27.781 | 5.508 | 14.266 | 45.789 | 8.633 | 16.609 |
| (2, 9) | 23.875 | 4.883 | 48.484 | 64.430 | 8.008 | 69.953 | 104.117 | 11.133 | 87.922 |
| (2, 11) | 54.609 | 14.883 | 198.234 | 141.203 | 18.008 | 292.703 | 225.969 | 21.133 | 373.172 |
| (2, 13) | 135.477 | 54.883 | 797.234 | 317.445 | 58.008 | 1183.703 | 497.508 | 61.133 | 1514.172 |
| (2, 15) | 361.859 | 214.883 | 3193.234 | 738.992 | 218.008 | 4747.703 | 1114.422 | 221.133 | 6078.172 |
| (2, 17) | 1071.258 | 854.883 | 12777.234 | 1845.562 | 858.008 | 19003.703 | 2615.656 | 861.133 | 24334.172 |
| (3, 5) | 12.008 | 2.617 | 11.688 | 34.609 | 5.742 | 14.812 | 57.617 | 8.867 | 17.938 |
| (3, 7) | 39.469 | 11.160 | 111.438 | 101.320 | 14.285 | 152.062 | 163.102 | 17.410 | 192.688 |
| (3, 9) | 169.469 | 88.047 | 1009.188 | 353.477 | 91.172 | 1387.312 | 537.328 | 94.297 | 1765.438 |
| (3, 11) | 971.930 | 780.027 | 9088.938 | 1529.031 | 783.152 | 12504.562 | 2085.367 | 786.277 | 15920.188 |
| (3, 13) | 7075.109 | 7007.852 | 81806.688 | 8763.023 | 7010.977 | 112559.812 | 10457.797 | 7014.102 | 143312.938 |
| (4, 5) | 20.664 | 5.016 | 47.453 | 56.703 | 8.141 | 112.609 | 93.031 | 11.266 | 129.016 |
| (4, 7) | 116.609 | 57.016 | 767.453 | 247.320 | 60.141 | 1826.359 | 377.922 | 63.266 | 2105.266 |
| (4, 9) | 1082.633 | 889.016 | 12287.453 | 1607.859 | 892.141 | 29246.359 | 2132.516 | 895.266 | 33725.266 |
| (5, 5) | 33.344 | 10.195 | 88.281 | 84.422 | 13.320 | 225.781 | 135.812 | 16.445 | 363.281 |
| (5, 7) | 324.258 | 217.715 | 2217.188 | 565.531 | 220.840 | 5675.000 | 806.797 | 223.965 | 9132.812 |
| (6, 5) | 51.742 | 19.773 | 151.047 | 121.031 | 22.898 | 372.141 | 190.758 | 26.023 | 593.234 |
| (6, 7) | 825.859 | 657.648 | 5449.797 | 1227.742 | 660.773 | 13433.391 | 1628.797 | 663.898 | 21416.984 |
| (7, 5) | 78.141 | 35.742 | 242.219 | 169.297 | 38.867 | 576.594 | 260.406 | 41.992 | 910.969 |
| (7, 7) | 1919.805 | 1677.051 | 11882.781 | 2542.047 | 1680.176 | 28295.281 | 3163.438 | 1683.301 | 44707.781 |

Table 8: Comparison of space required (in KB) for generating 100, 300, and 500 solutions for complete alternating AND/OR trees with bounded length **Open** queue for ASG and LASG

We show the effect of using bounded length queue to implement Open in Table 7 (reporting the time requirement) and in Table 8 (reporting the memory usage) for generating 100, 300, and 500 solutions, where the number of solutions to be generated are known beforehand. Table 7 and Table 8 show that in this case both ASG and LASG outperforms BU in terms of time as well as space requirements. Particularly, ASG performs very well in this setting, outperforming LASG in some cases.

### 5.1.2 Experimentation to Compare the Incremental Nature

The proposed top-down algorithms are incremental in nature whereas the existing bottom-up approach is not incremental. After generating a specified number of ordered solutions, our methods can generate the next solution incrementally without needing to restart itself, whereas the existing approach needs to be restarted. For example, after generating the first 10 ordered solutions, ASG and LASG generate the $11^{th}$ solution directly from the data structures maintained so far by these algorithms and perform necessary updates to these data structures. Whereas, BU needs to be restarted with input parameter 11 for generating the $11^{th}$ solution. In Table 9 we compare the time needed to generate the subsequent $11^{th}$ solution and $12^{th}$ solution incrementally after generating first 10 solutions. In order to have more clarity in the comparison among the running times of the respective algorithms, we have used higher precision (upto the $6^{th}$ decimal place) while reporting the running time in Table 9. Clearly, both ASG and LASG outperform BU for generating the $11^{th}$ and $12^{th}$ solution in terms of the time requirement.

| $(d, h)$ | ASG | | | LASG | | | BU | | |
|---|---|---|---|---|---|---|---|---|---|
| | first 10 | $11^{th}$ | $12^{th}$ | first 10 | $11^{th}$ | $12^{th}$ | first 10 | $11^{th}$ | $12^{th}$ |
| $(2, 7)$ | 0.002403 | 0.000201 | 0.000201 | 0.001003 | 0.000240 | 0.000123 | 0.001344 | 0.001359 | 0.001397 |
| $(2, 9)$ | 0.009111 | 0.001957 | 0.001302 | 0.003023 | 0.000714 | 0.000629 | 0.003596 | 0.003696 | 0.003895 |
| $(2, 11)$ | 0.028519 | 0.003311 | 0.003533 | 0.006700 | 0.001250 | 0.001346 | 0.014628 | 0.015046 | 0.015521 |
| $(2, 13)$ | 0.097281 | 0.014776 | 0.015929 | 0.025877 | 0.004113 | 0.004918 | 0.059326 | 0.061393 | 0.062717 |
| $(2, 15)$ | 0.396460 | 0.063641 | 0.059229 | 0.102493 | 0.014490 | 0.020031 | 0.238418 | 0.246042 | 0.251746 |
| $(2, 17)$ | 1.561020 | 0.251839 | 0.277763 | 0.446899 | 0.061422 | 0.082366 | 0.962635 | 0.989777 | 1.015848 |
| $(3, 5)$ | 0.001692 | 0.000158 | 0.000151 | 0.000683 | 0.000176 | 0.000112 | 0.001055 | 0.001101 | 0.001133 |
| $(3, 7)$ | 0.012097 | 0.001542 | 0.001572 | 0.004084 | 0.000583 | 0.000959 | 0.009507 | 0.009931 | 0.010336 |
| $(3, 9)$ | 0.097356 | 0.013046 | 0.014405 | 0.031159 | 0.003948 | 0.004604 | 0.085610 | 0.089379 | 0.093419 |
| $(3, 11)$ | 0.934389 | 0.127943 | 0.156579 | 0.311128 | 0.033169 | 0.047594 | 0.778298 | 0.811176 | 0.846578 |
| $(3, 13)$ | 7.898530 | 1.082319 | 1.194090 | 2.811539 | 0.282836 | 0.387715 | 7.037050 | 7.313715 | 7.608619 |
| $(4, 5)$ | 0.005833 | 0.000650 | 0.000671 | 0.002143 | 0.000303 | 0.000582 | 0.004181 | 0.004434 | 0.004725 |
| $(4, 7)$ | 0.051598 | 0.006956 | 0.007196 | 0.017046 | 0.002209 | 0.003115 | 0.044913 | 0.047867 | 0.050940 |
| $(4, 9)$ | 0.813028 | 0.110205 | 0.124750 | 0.294561 | 0.027612 | 0.037281 | 0.727766 | 0.775950 | 0.823442 |
| $(5, 5)$ | 0.051530 | 0.001327 | 0.001641 | 0.004638 | 0.000753 | 0.000652 | 0.005963 | 0.006358 | 0.006782 |
| $(5, 7)$ | 0.172475 | 0.024262 | 0.024438 | 0.059751 | 0.006116 | 0.007197 | 0.152285 | 0.162527 | 0.173191 |
| $(6, 5)$ | 0.053422 | 0.002701 | 0.003092 | 0.005282 | 0.000636 | 0.001087 | 0.010895 | 0.011604 | 0.012556 |
| $(6, 7)$ | 0.502939 | 0.061417 | 0.069727 | 0.184584 | 0.017116 | 0.024042 | 0.406947 | 0.435398 | 0.465301 |
| $(7, 5)$ | 0.033831 | 0.003706 | 0.003846 | 0.012862 | 0.001266 | 0.001282 | 0.018185 | 0.019567 | 0.020896 |
| $(7, 7)$ | 1.198354 | 0.156145 | 0.166501 | 0.466560 | 0.038792 | 0.061305 | 0.929941 | 0.989326 | 1.052566 |

Table 9: Comparison of running time (in seconds) for generating for first 10 solutions and then the $11^{th}$ solution and $12^{th}$ solution incrementally for complete alternating AND/OR trees

## 5.2 Multipeg Tower of Hanoi Problem

Consider the problem of Multipeg Tower of Hanoi (Majumdar, 1996; Gupta, Chakrabarti, & Ghose, 1992). In this problem, $\rho$ pegs are fastened to a stand. Initially $\gamma$ disks rest on the source peg $\mathcal{A}$ with small disk on large disk ordering. The objective is to transfer all $\gamma$ disks from $\mathcal{A}$ to the destination peg $\mathcal{B}$ with minimum *legal* moves. In a legal move, the topmost disk from any tower can be transferred to any other peg with a larger disk as the topmost disk. The problem of multi-peg tower of Hanoi can be solved recursively as follows.

   a. Move recursively the topmost $k$ ($k$ varies from 1 to $\gamma - 1$) disks from $\mathcal{A}$ to some intermediate peg, $\mathcal{I}$, using all the pegs.

   b. Transfer the remaining $\gamma - k$ disks from $\mathcal{A}$ to $\mathcal{B}$ recursively, using the $(\rho - 1)$ pegs available.

   c. Recursively move $k$ disks that were transferred to $\mathcal{I}$ previously, from the intermediate peg $\mathcal{I}$ to $\mathcal{B}$, using all the $\rho$ pegs.

It may be noted that there is a choice for the value of $k$, which may take any value from 1 to $\gamma - 1$. Solutions with different values of $k$ may take different number of moves, and the solution which incurs minimum number of moves is the optimal solution. This choice of the value of $k$ is modeled as an OR node, and for every such choice, the problem is divided into three sub-problems. This decomposition into sub-problems is modeled as an AND node. Therefore, the search spaces of the multi-peg Tower of Hanoi problem correspond to alternating AND/OR trees.

| #disks | 100 solutions | | | 300 solutions | | | 500 solutions | | | #Opt. No. of Moves |
|---|---|---|---|---|---|---|---|---|---|---|
| | ASG | LASG | BU | ASG | LASG | BU | ASG | LASG | BU | |
| 8 | 0.034 | 0.030 | 0.069 | 0.104 | 0.084 | 0.252 | 0.200 | 0.138 | 0.577 | 23 |
| 9 | 0.119 | 0.116 | 0.264 | 0.314 | 0.289 | 0.942 | 0.590 | 0.458 | 2.183 | 27 |
| 10 | 0.479 | 0.635 | 1.310 | 1.706 | 1.658 | 3.305 | 2.303 | 2.829 | 7.592 | 31 |
| 11 | 2.421 | 2.178 | 3.171 | 6.573 | 6.161 | 12.998 | 10.651 | 9.678 | 29.242 | 39 |
| 12 | 7.453 | 7.448 | 11.437 | 21.232 | 21.081 | 43.358 | 35.825 | 35.663 | 99.593 | 47 |
| 13 | 25.379 | 25.115 | 38.458 | 68.574 | 67.170 | 140.392 | 112.411 | 112.470 | 332.113 | 55 |

Table 10: Comparison of running time (in seconds) for alternating AND/OR trees corresponding to the search spaces of 5-peg Tower of Hanoi problem with different number of disks

| #disks | 100 solutions | | | 300 solutions | | | 500 solutions | | |
|---|---|---|---|---|---|---|---|---|---|
| | ASG | LASG | BU | ASG | LASG | BU | ASG | LASG | BU |
| 8 | 36.664 | 43.008 | 416.312 | 64.516 | 80.734 | 660.062 | 105.039 | 117.008 | 903.812 |
| 9 | 96.211 | 111.320 | 1471.656 | 131.266 | 154.266 | 2359.156 | 166.789 | 197.859 | 3246.656 |
| 10 | 295.672 | 341.000 | 5074.219 | 326.352 | 383.453 | 8161.719 | 373.453 | 427.766 | 11249.219 |
| 11 | 957.336 | 1113.508 | 17197.312 | 999.602 | 1158.797 | 27728.562 | 1039.367 | 1204.719 | 38259.812 |
| 12 | 3155.086 | 3664.117 | 57512.812 | 3198.156 | 3719.352 | 92906.562 | 3247.547 | 3767.617 | 128300.312 |
| 13 | 10339.078 | 12022.883 | 190297.969 | 10412.242 | 12078.914 | 307872.969 | 10483.570 | 12137.242 | 425447.969 |

Table 11: Comparison of space required (in KB) for alternating AND/OR trees corresponding to the search spaces of 5-peg Tower of Hanoi problem with different number of disks

We have used the search space of 5 peg Tower of Hanoi problem with different number of disks, $\gamma$, and generated alternative solutions in non-decreasing order of cost using ASG and LASG algorithms. Here the cost function expresses the number of legal moves. The value of $\gamma$ is varied from 8 to 13, and in Table 10 and in Table 11, we report the time required and space required, respectively, for generating 100, 300, and 500 solutions for every test cases. Experimental results show that the performance of ASG is similar to the performance of LASG with respect to both space and time. However ASG as well as LASG outperforms BU with respect to both time and space requirements.

## 5.3 Randomly Constructed AND/OR DAGs

We have constructed a set of randomly generated AND/OR DAGs and evaluated the ASG, LASG, and BU algorithm for generating solutions under default semantics. We have used the proposed extension to the BU algorithm for generating solutions under default semantics.

| $n_{\alpha\beta}$ | $d$ | 100 solutions | | | 300 solutions | | | 500 solutions | | |
|---|---|---|---|---|---|---|---|---|---|---|
| | | ASG | LASG | BU | ASG | LASG | BU | ASG | LASG | BU |
| 60 | 2 | 0.027 | 0.006 | 0.039 | 0.089 | 0.021 | 0.158 | 0.172 | 0.033 | 0.282 |
| 220 | 2 | 0.060 | 0.009 | 0.096 | 0.281 | 0.030 | 1.100 | 0.594 | 0.051 | 3.665 |
| 920 | 2 | 0.363 | 0.020 | 0.106 | 2.485 | 0.059 | 0.266 | 6.163 | 0.100 | 0.528 |
| 33 | 3 | 0.020 | 0.006 | 0.019 | 0.123 | 0.021 | 0.098 | 0.280 | 0.032 | 0.245 |
| 404 | 3 | 0.203 | 0.018 | 0.067 | 1.483 | 0.048 | 0.257 | 4.043 | 0.083 | 0.541 |
| 2124 | 3 | 3.550 | 0.045 | 0.730 | 30.302 | 0.126 | 1.681 | 85.863 | 0.215 | 2.766 |
| 9624 | 3 | 26.659 | 0.201 | 14.620 | 257.605 | 0.612 | 33.382 | 710.708 | 1.194 | 52.406 |
| 144 | 4 | 0.065 | 0.008 | 0.034 | 0.348 | 0.027 | 0.217 | 0.817 | 0.049 | 2.251 |
| 744 | 4 | 0.877 | 0.025 | 0.400 | 6.910 | 0.069 | 0.994 | 18.823 | 0.118 | 1.365 |
| 8844 | 4 | 7.422 | 0.160 | 26.683 | 69.097 | 0.449 | 66.558 | 194.452 | 0.927 | 109.076 |
| 40884 | 4 | T | 1.972 | T | T | 5.819 | T | T | 9.426 | T |

Table 12: Comparison of running time (in seconds) for generating 100, 300, and 500 solutions for AND/OR DAGs ($T$ denotes the timeout after 15 minutes)

| $n_{\alpha\beta}$ | $d$ | 100 solutions | | | 300 solutions | | | 500 solutions | | |
|---|---|---|---|---|---|---|---|---|---|---|
| | | ASG | LASG | BU | ASG | LASG | BU | ASG | LASG | BU |
| 60 | 2 | 11.609 | 8.875 | 8.125 | 32.852 | 30.797 | 10.906 | 54.094 | 50.035 | 13.250 |
| 220 | 2 | 23.141 | 16.219 | 31.312 | 62.516 | 46.711 | 49.562 | 100.555 | 74.379 | 65.188 |
| 920 | 2 | 74.082 | 39.000 | 106.875 | 220.648 | 105.852 | 172.562 | 371.344 | 168.375 | 230.375 |
| 33 | 3 | 13.914 | 10.492 | 8.172 | 46.117 | 32.539 | 11.297 | 77.445 | 54.602 | 14.422 |
| 404 | 3 | 48.867 | 35.445 | 66.938 | 151.363 | 101.168 | 98.188 | 262.816 | 163.273 | 129.438 |
| 2124 | 3 | 229.820 | 118.707 | 389.844 | 705.809 | 312.246 | 621.094 | 1200.336 | 507.762 | 852.344 |
| 9624 | 3 | 772.441 | 339.676 | 1996.875 | 2245.938 | 825.984 | 3321.875 | 3732.523 | 1327.406 | 4646.875 |
| 144 | 4 | 30.648 | 17.332 | 29.609 | 85.781 | 53.961 | 73.359 | 140.312 | 93.539 | 86.641 |
| 744 | 4 | 121.535 | 65.578 | 287.109 | 381.133 | 168.305 | 737.891 | 659.434 | 275.594 | 883.984 |
| 8844 | 4 | 471.625 | 266.078 | 2729.297 | 1183.379 | 550.477 | 6945.703 | 1927.961 | 843.484 | 8419.922 |
| 40884 | 4 | 2722.938 | 1256.535 | T | T | 2353.562 | T | T | 3447.809 | T |

Table 13: Comparison of space required (in KB) for generating 100, 300, and 500 solutions for AND/OR DAGs

Table 12 and Table 13 compare the time required and space required for running ASG, LASG and BU for generating 100, 300, and 500 solutions for every test cases. The first and second columns of every row provide the size ($n_{\alpha\beta}$) and the average out-degree ($d$) of the DAG. The results obtained for this test domain are similar to the results for randomly constructed AND/OR trees. It may be noted that in terms of both time and space required, LASG outperforms both ASG and BU. Between ASG and BU, for most of the test cases BU performs better than ASG with respect to the time required for generating a specific number of solutions. Whereas, the space requirement of ASG and BU for generating a specific number of solutions has an interesting co-relation with the average *degree*($d$) and the size ($n_{\alpha\beta}$) parameter of the DAG. For low numerical values of the $d$ and the $n_{\alpha\beta}$ parameter, e.g., ($n_{\alpha\beta}, d$) combinations like (60, 2), (33, 3) etc., BU performs better than ASG. On the contrary, for the other combinations, where at least one of these $n_{\alpha\beta}$ and $d$ parameter has a high value, e.g., ($n_{\alpha\beta}, d$) combinations like (920, 2), (9624, 3), (40884, 4) etc., ASG outperforms BU.

## 5.4 Matrix-Chain Multiplication Problem

We have also used the well-known *matrix-chain multiplication* (Cormen, Stein, Rivest, & Leiserson, 2001) problem for experimentation. The search space of the popular dynamic programming formulation of this problem correspond to AND/OR DAG.

| #matrices | DAG Cnstr. Time (Sec) | $S_{opt}$ Cnstr. Time (Sec) | 10 solutions | | | 15 solutions | | | 20 solutions | | |
|---|---|---|---|---|---|---|---|---|---|---|---|
| | | | ASG | LASG | BU | ASG | LASG | BU | ASG | LASG | BU |
| 20 | 0.033 | 0.001 | 0.003 | 0.002 | 0.206 | 0.004 | 0.003 | 0.288 | 0.005 | 0.004 | 0.373 |
| 30 | 0.200 | 0.003 | 0.009 | 0.008 | 2.785 | 0.012 | 0.010 | 4.087 | 0.015 | 0.012 | 5.406 |
| 40 | 0.898 | 0.008 | 0.019 | 0.018 | 15.580 | 0.024 | 0.024 | 23.414 | 0.030 | 0.030 | 31.112 |
| 50 | 3.033 | 0.016 | 0.047 | 0.048 | 93.267 | 0.062 | 0.065 | 140.513 | 0.079 | 0.081 | 187.227 |
| 60 | 8.335 | 0.029 | 0.088 | 0.090 | 342.212 | 0.118 | 0.120 | 509.906 | 0.148 | 0.151 | 678.718 |
| 70 | 19.591 | 0.046 | 0.140 | 0.142 | 862.387 | 0.187 | 0.190 | T | 0.235 | 0.238 | T |
| 80 | 41.960 | 0.071 | 0.209 | 0.212 | T | 0.280 | 0.282 | T | 0.351 | 0.354 | T |
| 90 | 82.578 | 0.101 | 0.296 | 0.300 | T | 0.396 | 0.398 | T | 0.496 | 0.499 | T |
| 100 | 151.814 | 0.143 | 0.409 | 0.412 | T | 0.546 | 0.548 | T | 0.688 | 0.683 | T |

Table 14: Comparison of time required (in seconds) for AND/OR DAGs corresponding to the search spaces of *matrix-chain multiplication* with different number of matrices, ($T$ denotes the timeout after 15 minutes)

| #matrices | 10 solutions | | | 15 solutions | | | 20 solutions | | |
|---|---|---|---|---|---|---|---|---|---|
| | ASG | LASG | BU | ASG | LASG | BU | ASG | LASG | BU |
| 20 | 19.641 | 20.203 | 160.918 | 20.543 | 21.227 | 234.305 | 21.914 | 22.773 | 303.973 |
| 30 | 66.367 | 69.273 | 555.684 | 67.809 | 70.695 | 821.902 | 69.516 | 72.523 | 1081.902 |
| 40 | 156.559 | 160.227 | 1317.637 | 157.738 | 161.785 | 1960.281 | 158.758 | 162.852 | 2594.207 |
| 50 | 308.984 | 315.012 | 2563.965 | 310.277 | 316.543 | 3825.223 | 311.551 | 318.145 | 5075.262 |
| 60 | 537.383 | 545.117 | 4411.855 | 538.930 | 546.512 | 6592.508 | 539.914 | 547.551 | 8759.441 |
| 70 | 859.844 | 869.160 | 6978.496 | 862.133 | 870.867 | T | 863.977 | 872.219 | T |
| 80 | 1290.117 | 1301.406 | T | 1293.148 | 1303.426 | T | 1295.852 | 1305.090 | T |
| 90 | 1843.828 | 1857.480 | T | 1847.602 | 1859.812 | T | 1851.164 | 1861.789 | T |
| 100 | 2537.582 | 2556.883 | T | 2542.746 | 2560.043 | T | 2549.352 | 2566.992 | T |

Table 15: Comparison of space required (in KB) for AND/OR DAGs corresponding to the search spaces of *matrix-chain multiplication* with different number of matrices

Given a sequence of matrices, $A_1, A_2, \cdots, A_n$, of $n$ matrices where matrix $A_i$ has dimension $p_{i-1} \times p_i$, in this problem the objective is to find the most efficient way to multiply these matrices. The classical dynamic programming approach works as follows. Suppose $A_{[i,j]}$ denotes matrix that results from evaluating the product, $A_i A_{i+1} \cdots A_j$, and $m[i,j]$ is the minimum number of scalar multiplications required for computing the matrix $A_{[i,j]}$. Therefore, the cost of optimal solution is denoted by $m[i,j]$ which can be recursively defined as :

$$m[i,j] = \begin{cases} 0, & \text{if } i = j; \\ \min_{i \leq k < j} \left\{ m[i,k] + m[k+1,j] + p_{i-1} \times p_k \times p_j \right\}, & \text{if } i < j. \end{cases}$$

The choice of the value of $k$ is modeled as OR node and for every such choice, the problem is divided into three sub-problems. This decomposition into sub-problems is modeled as an AND node. It is worth noting that unlike the search space of 5-*peg ToH* problem, the search space of the *matrix-chain multiplication* problem corresponds to AND/OR DAG. We have used the search space for different matrix sequences having varying length and generated alternative solutions in the order of non-decreasing cost. In Table 14, we report the time required and in Table 15, we report the memory used for generating 10, 15, and 20 solutions for every test cases.

In Table 14, for each test case, we also report the time required for constructing the *explicit* AND/OR DAG from the recursive formulation in the $2^{nd}$ column, and the *optimal solution* construction time in the $3^{rd}$ column. It is interesting to observe that the relative performance of ASG and LASG for this search space is very similar to that obtained for 5-*peg ToH* search space though this search space for this domain is AND/OR DAG. Both ASG and LASG perform approximately the same with respect to time and space requirement. However, the advantage of ASG as well as LASG over BU with respect to both time and space requirement is more significant in this domain.

## 5.5 Generating Secondary Structure for RNA

Another relevant problem where the alternative solutions play an important role is the computation of the secondary structure of RNA. RNA molecules can be viewed as strings of bases, where each base belongs to the set $\{Adenine, Cytocine, Guanine, Uracil\}$ (also denoted as $\{A, C, G, U\}$). RNA molecules tend to loop back and form base pairs with itself and the resulting shape is called *secondary structure* (Mathews & Zuker, 2004). The stability of the secondary structure largely depends on the number of base pairings (in general, larger number of base pairings implies more stable secondary structure). Although there are other factors that influence the secondary structure, it is often not possible to express these other factors using a cost function and they are typically evaluated empirically. Therefore, it is useful to generate a set of possible alternative secondary structures ordered by decreasing numbering of base pairings for a given RNA which can be further subjected to experimental evaluation.

The computation of the optimal secondary structure considering the underlying principle of maximizing the number of base-pairings has a nice dynamic programming formulation (Kleinberg & Tardos, 2005). Given an RNA molecule $B = \langle b_1 b_2 \cdots b_n \rangle$ where each $b_i \in \{A, C, G, U\}$, the secondary structure on $B$ is a set of base pairings, $\mathcal{D} = \{(i,j)\}$, where $i, j \in \{1, 2, \cdots n\}$, that satisfies the following conditions:

| Test Case | Organism Name | # Bases |
|-----------|---------------|---------|
| TC1 | Anaerorhabdus Furcosa | 114 |
| TC2 | Archaeoglobus Fulgidus | 124 |
| TC3 | Chlorobium Limicola | 111 |
| TC4 | Desulfurococcus Mobilis | 129 |
| TC5 | Haloarcula Japonica | 122 |
| TC6 | Halobacterium Sp. | 120 |
| TC7 | Mycoplasma Genitalium | 104 |
| TC8 | Mycoplasma Hyopneumoniae | 105 |
| TC9 | Mycoplasma Penetrans | 103 |
| TC10 | Pyrobaculum Aerophilum | 131 |
| TC11 | Pyrococcus Abyssi | 118 |
| TC12 | Spiroplasma Melliferum | 107 |
| TC13 | Sulfolobus Acidocaldarius | 126 |
| TC14 | Symbiobacterium Thermophilum | 110 |

Table 16: Details of the RNA sequences used for Experimentation

a. if $(i, j) \in \mathcal{D}$, then $i + 4 < j$ : This condition states that the ends of each pair in $\mathcal{D}$ are separated by at least four intermediate bases.

b. The elements of any pair in $\mathcal{D}$ consists of either $\{A, U\}$ or $\{C, G\}$ (in either order).

c. No base appears in more than one pairings, i.e., $\mathcal{D}$ is a matching.

d. If $(i, j)$ and $(k, l)$ are two pairs in $\mathcal{D}$, then it is not possible to have $i < k < l < j$, i.e., no two pairings can cross each other.

| Test Case | DAG Cnstr. Time (Sec) | $S_{opt}$ Cnstr. Time (Sec) | 5 solutions | | | 10 solutions | | | 15 solutions | | |
|-----------|-----------------------|------------------------------|------|------|------|------|------|------|------|------|------|
| | | | ASG | LASG | BU | ASG | LASG | BU | ASG | LASG | BU |
| TC1 | 34.464 | 0.042 | 0.094 | 0.095 | 449.916 | 0.145 | 0.148 | 893.682 | 0.197 | 0.202 | 1359.759 |
| TC2 | 57.999 | 0.057 | 0.126 | 0.128 | 823.493 | 0.193 | 0.198 | T | 0.271 | 0.277 | T |
| TC3 | 26.423 | 0.038 | 0.084 | 0.089 | 363.421 | 0.135 | 0.133 | 718.326 | 0.183 | 0.186 | 1077.094 |
| TC4 | 83.943 | 0.065 | 0.144 | 0.152 | 1089.462 | 0.230 | 0.227 | T | 0.314 | 0.317 | T |
| TC5 | 51.290 | 0.051 | 0.114 | 0.116 | 681.429 | 0.176 | 0.180 | 1349.181 | 0.239 | 0.245 | T |
| TC6 | 46.508 | 0.047 | 0.107 | 0.108 | 598.419 | 0.166 | 0.175 | T | 0.226 | 0.238 | T |
| TC7 | 16.766 | 0.029 | 0.068 | 0.069 | 210.806 | 0.101 | 0.103 | 410.817 | 0.136 | 0.144 | 621.792 |
| TC8 | 22.775 | 0.033 | 0.077 | 0.078 | 284.455 | 0.120 | 0.122 | 559.318 | 0.153 | 0.165 | 836.359 |
| TC9 | 18.831 | 0.031 | 0.068 | 0.072 | 233.999 | 0.109 | 0.111 | 458.290 | 0.144 | 0.148 | 683.411 |
| TC10 | 91.419 | 0.073 | 0.167 | 0.170 | T | 0.249 | 0.263 | T | 0.347 | 0.355 | T |
| TC11 | 47.660 | 0.047 | 0.111 | 0.109 | 627.744 | 0.173 | 0.171 | 1253.034 | 0.220 | 0.240 | T |
| TC12 | 22.649 | 0.034 | 0.078 | 0.079 | 288.520 | 0.116 | 0.123 | 573.602 | 0.165 | 0.167 | 849.134 |
| TC13 | 67.913 | 0.061 | 0.140 | 0.141 | 962.641 | 0.206 | 0.218 | T | 0.290 | 0.288 | T |
| TC14 | 28.911 | 0.038 | 0.087 | 0.085 | 366.693 | 0.134 | 0.137 | 724.113 | 0.182 | 0.186 | 1072.552 |

Table 17: Comparison of time required (in seconds) for AND/OR DAGs corresponding to the search spaces of *RNA secondary structure* with different number of bases ($T$ denotes the timeout after 30 minutes)

Under the above mentioned conditions the dynamic programming formulation is as follows. Suppose $P(i, j)$ denotes the maximum number of base pairings in a secondary structure on $b_i \cdots b_j$. $P(i, j)$ can be recursively defined as :

$$P[i, j] = \begin{cases} 0, & \text{if } i + 4 \geq j, \\ max \Big\{ P[i, j-1], \ \max_{i \leq k < j} \big\{ 1 + P[i, k-1] + P[k+1, j-1] \big\}, \Big\} & \text{if } i + 4 < j. \end{cases}$$

Here, a choice of the value of $k$ is modeled as an OR node and for every such choice, the problem is divided into three sub-problems. This decomposition into sub-problems is modeled as an AND node. We have experimented with the search space of this problem for the set of RNA molecule sequences obtained from the test-cases developed by Szymanski, Barciszewska, Barciszewski, and Erdmann (2005). The details of the test cases are shown in Table 16.

For every test cases, we report the time required in Table 17 for generating 5, 10, and 15 solutions. For the same setting, the space required is reported in Table 18. In Table 17, for each test case, we also report the time required for constructing the *explicit* AND/OR DAG from the recursive formulation in the $2^{nd}$ column, and the time required for constructing the *optimal solution* time in the $3^{rd}$ column. We use a high value of time-out (1800 seconds) in order to gather the running time required by BU. We limit the maximum solutions generated at 15 because for generating higher number of solutions, BU is timed out for most of the test cases. It is worth noting that the result obtained for this domain is very similar to the result obtained for the *matrix-chain multiplication* problem domain. Both space and time wise ASG and LASG perform similarly and they outperform BU significantly with respect to time as well as space requirement.

| Test | 5 solutions | | | 10 solutions | | | 15 solutions | | |
|------|------|------|------|------|------|------|------|------|------|
| Case | ASG | LASG | BU | ASG | LASG | BU | ASG | LASG | BU |
| TC1 | 1647.555 | 1694.688 | 7409.336 | 1651.273 | 1697.797 | 14656.469 | 1654.367 | 1700.492 | 21846.156 |
| TC2 | 2254.531 | 2310.008 | 9902.953 | 2258.773 | 2315.258 | T | 2262.492 | 2318.008 | T |
| TC3 | 1473.852 | 1516.922 | 6629.891 | 1477.492 | 1521.750 | 13103.492 | 1480.555 | 1526.797 | 19518.625 |
| TC4 | 2606.242 | 2665.820 | 11358.945 | 2610.875 | 2671.711 | T | 2615.719 | 2675.633 | T |
| TC5 | 2045.930 | 2097.414 | 9021.273 | 2049.844 | 2101.836 | 17875.430 | 2052.867 | 2106.000 | T |
| TC6 | 1912.227 | 1963.367 | 8499.570 | 1916.422 | 1968.305 | T | 1921.117 | 1972.172 | T |
| TC7 | 1101.125 | 1138.633 | 5087.680 | 1104.422 | 1142.023 | 10036.938 | 1108.047 | 1144.109 | 14924.820 |
| TC8 | 1293.812 | 1333.336 | 5855.547 | 1297.750 | 1338.070 | 11560.203 | 1302.242 | 1342.484 | 17211.406 |
| TC9 | 1170.094 | 1207.633 | 5352.477 | 1173.023 | 1211.523 | 10562.766 | 1176.352 | 1213.906 | 15718.617 |
| TC10 | 2984.773 | 3047.539 | T | 2990.211 | 3053.977 | T | 2994.773 | 3059.781 | T |
| TC11 | 1974.695 | 2022.906 | 8641.422 | 1979.344 | 2030.922 | 17119.820 | 1983.664 | 2038.461 | T |
| TC12 | 1295.141 | 1335.883 | 5924.664 | 1297.273 | 1339.516 | 11701.695 | 1299.805 | 1341.914 | 17420.719 |
| TC13 | 2438.898 | 2496.469 | 10657.945 | 2442.961 | 2502.625 | T | 2447.172 | 2506.703 | T |
| TC14 | 1475.477 | 1517.828 | 6627.844 | 1478.555 | 1521.352 | 13099.055 | 1482.234 | 1525.344 | 19519.742 |

Table 18: Comparison of space required (in KB) for AND/OR DAGs corresponding to the search spaces of *RNA secondary structure* with different number of bases

## 5.6 Observations

The experimental data shows that the LASG algorithm generally outperforms the ASG algorithm and the existing bottom-up approach in terms of the running time for *complete* alternating AND/OR trees and AND/OR DAGs. Whereas, for the other problem domains, i.e., the 5-*peg Tower of Hanoi* problem, the *matrix-chain multiplication* problem, and the problem of determining *secondary structure of RNA sequences*, the overall performance of the ASG algorithm is similar to the performance of the LASG algorithm. This behavior can be explained from the average and maximum length statistics of Open list, reported in Table 19 - Table 23, for these above mentioned test domains.

In the case of *complete* trees and random DAGs, for ASG algorithm, the average as well as the maximum size of **Open** grows much faster than that of LASG algorithm (Table 19 and Table 20), with the increase in the size of the tree/DAG.

| $(d, h)$ | 100 solutions | | | | 300 solutions | | | | 500 solutions | | | |
|---|---|---|---|---|---|---|---|---|---|---|---|---|
| | *ASG* | | *LASG* | | *ASG* | | *LASG* | | *ASG* | | *LASG* | |
| | avg. | max. | avg. | max. | avg. | max. | avg. | max. | avg. | max. | avg. | max. |
| $(2, 7)$ | 235 | 383 | 75 | 159 | 435 | 629 | 179 | 289 | 545 | 792 | 236 | 372 |
| $(2, 9)$ | 994 | 1894 | 73 | 120 | 2657 | 4931 | 220 | 528 | 4103 | 7569 | 449 | 1069 |
| $(2, 11)$ | 2427 | 4709 | 156 | 306 | 6935 | 13537 | 483 | 1005 | 11251 | 21843 | 851 | 1771 |
| $(2, 13)$ | 5546 | 10947 | 524 | 1149 | 16266 | 32076 | 1550 | 2726 | 26748 | 52724 | 2261 | 3844 |
| $(2, 15)$ | 11744 | 23291 | 384 | 523 | 34836 | 69160 | 677 | 1121 | 57673 | 114367 | 983 | 1824 |
| $(2, 17)$ | 24264 | 48333 | 655 | 841 | T | T | 1087 | 1611 | T | T | 1527 | 2819 |
| $(3, 5)$ | 304 | 549 | 120 | 242 | 740 | 1323 | 341 | 652 | 1107 | 1972 | 539 | 1007 |
| $(3, 7)$ | 1561 | 3015 | 172 | 346 | 4359 | 8400 | 579 | 1260 | 7026 | 13588 | 1012 | 2084 |
| $(3, 9)$ | 5496 | 10899 | 191 | 289 | 16272 | 32244 | 387 | 661 | 26904 | 53271 | 622 | 1368 |
| $(3, 11)$ | 17336 | 34542 | 486 | 691 | 51954 | 103549 | 956 | 1754 | T | T | 1460 | 2672 |
| $(3, 13)$ | 53139 | 106155 | 1138 | 1216 | T | T | 1267 | 1569 | T | T | 1432 | 1776 |
| $(4, 5)$ | 734 | 1427 | 103 | 176 | 2062 | 4006 | 256 | 503 | 3322 | 6375 | 452 | 1065 |
| $(4, 7)$ | 3748 | 7383 | 194 | 381 | 10928 | 21489 | 678 | 1467 | 17932 | 35222 | 1265 | 2837 |
| $(4, 9)$ | 16282 | 32451 | 422 | 488 | 48786 | 97196 | 687 | 1131 | T | T | 1025 | 1807 |
| $(5, 5)$ | 1216 | 2352 | 146 | 307 | 3407 | 6555 | 496 | 1053 | 5508 | 10694 | 852 | 1742 |
| $(5, 7)$ | 7261 | 14446 | 249 | 335 | 21652 | 42972 | 470 | 888 | 35850 | 71054 | 832 | 1781 |
| $(6, 5)$ | 1781 | 3489 | 141 | 276 | 5089 | 9911 | 507 | 1126 | 8250 | 16035 | 971 | 2164 |
| $(6, 7)$ | 12362 | 24651 | 297 | 342 | 36868 | 73323 | 461 | 789 | 61221 | 121958 | 749 | 1573 |
| $(7, 5)$ | 2433 | 4765 | 261 | 508 | 7072 | 13910 | 747 | 1483 | 11595 | 22809 | 1204 | 2273 |
| $(7, 7)$ | 19311 | 38435 | 450 | 529 | 57754 | 115116 | 687 | 961 | T | T | 984 | 1922 |

Table 19: Average and maximum length of **Open** while generating 100, 300, and 500 solutions for complete alternating AND/OR trees

| $n_{\alpha\beta}$ | $d$ | 100 solutions | | | | 300 solutions | | | | 500 solutions | | | |
|---|---|---|---|---|---|---|---|---|---|---|---|---|---|
| | | *ASG* | | *LASG* | | *ASG* | | *LASG* | | *ASG* | | *LASG* | |
| | | avg. | max. | avg. | max. | avg. | max. | avg. | max. | avg. | max. | avg. | max. |
| 60 | 2 | 181 | 338 | 39 | 63 | 428 | 768 | 131 | 282 | 643 | 1138 | 219 | 411 |
| 220 | 2 | 479 | 854 | 77 | 133 | 1144 | 2058 | 210 | 417 | 1721 | 3139 | 329 | 612 |
| 920 | 2 | 1530 | 2957 | 116 | 227 | 4289 | 8278 | 332 | 639 | 6902 | 13305 | 512 | 946 |
| 33 | 3 | 202 | 409 | 58 | 102 | 604 | 1193 | 154 | 281 | 978 | 1875 | 234 | 422 |
| 404 | 3 | 1001 | 1969 | 236 | 447 | 2958 | 5799 | 675 | 1256 | 4874 | 9781 | 1013 | 1810 |
| 2124 | 3 | 5008 | 9911 | 374 | 626 | 14803 | 29314 | 851 | 1569 | 24442 | 48357 | 1337 | 2527 |
| 9624 | 3 | 14422 | 28666 | 394 | 491 | 43087 | 85825 | 746 | 1339 | 71547 | 142327 | 1254 | 2756 |
| 144 | 4 | 510 | 990 | 56 | 101 | 1374 | 2563 | 187 | 458 | 2140 | 3996 | 376 | 868 |
| 744 | 4 | 2407 | 4760 | 253 | 485 | 7166 | 14204 | 590 | 1018 | 11874 | 23558 | 885 | 1655 |
| 8844 | 4 | 7522 | 14931 | 258 | 437 | 22254 | 44062 | 847 | 1831 | 36743 | 72740 | 1565 | 3493 |
| 40884 | 4 | T | T | 749 | 804 | T | T | 852 | 1004 | T | T | 961 | 1215 |

Table 20: Average and maximum length of **Open** while generating 100, 300, and 500 solutions for randomly constructed AND/OR DAGs

Since ASG algorithm checks for the presence of duplicates while expanding a solution, the time required for duplication checking grows rapidly for these test domains. Hence, the overall time required for generating a specific number of solutions also increases rapidly (faster than both BU and LASG) with the increase in the size of the tree/DAG. As a result, BU outperforms ASG with respect to the time requirement for trees and DAGs. However the memory used for generating a specific number of solutions increases moderately (slower than BU) with the increase in the size of the tree/DAG. Therefore with respect to space requirement, ASG outperforms BU for larger trees and DAGs.

Between LASG and BU, the time as well as the memory requirement of BU increases faster than that of LASG when the degree of the AND/OR tree or DAG increases. This happens because, for BU, the time taken for merging the sub-solutions at the AND nodes and memory required for storing alternative solutions that are rooted at different nodes increases rapidly with the increase in the degree of that node.

On the contrary, for the other test domains, 5-*peg Tower of Hanoi* problem, *matrix-chain multiplication* problem, and the probelm of finding *secondary structure of RNA sequences*, the average and the maximum size of Open for both ASG and LASG are comparable (Table 21, Table 22 and Table 23). Therefore, for the LASG algorithm, the time saved by avoiding the *duplication checking* is compensated by the extra overhead of maintaining the *solution space tree* and the checks required for *lazy expansion*. Hence the running time as well as the space requirement are almost same for both algorithms for these three above mentioned problem domains.

Moreover, due to the low values of the average and the maximum size of Open, ASG outperforms BU with respect to both time requirement and memory used for these three test domains. For these three domains also, between LASG and BU, the time as well as the memory requirement of BU increases faster than that of LASG when the size of the search space (AND/OR tree or DAG) increases.

## 6. Ramifications on Implicitly Specified AND/OR Structures

In this section, we briefly discuss use of our proposed algorithms for generation of alternative solutions in the non-decreasing order of cost for implicit AND/OR search spaces. One possible way is to extend the standard $AO^*$ for generating a given number of solutions, say $k$, as follows. Instead of keeping only one *potential solution graph(psg)*, at any stage $k$ psgs can be computed on the explicitly constructed search space and instead of expanding one node, $k$ nodes, (that is, one node from each psg), can be expanded at once. After expanding the nodes, $k$ psgs are recomputed once again. Since the cost of the nodes are often recomputed after expanding nodes, the swap options associated with any such node have to be updated after every such recomputation.

Another possible approach could be to run $AO^*$ until it generates the optimal solution. At this point of time the swap options can be computed on the explicit portion of the graph and swap option with minimum cost can be applied to the optimal solution. Then the resulting psg is again expanded further resulting in the expansion of the explicit graph. The swap options are re-evaluated to incorporate the cost update. Again the next best psg is computed. This process continues till the second best solution is derived. Now among the remaining successor psgs of the first solution and the successor psgs of second solution, the most promising psg is selected and expanded. This process continues till the third solution is found. Then the successor psgs are also added to the already existing pool of candidate psgs. These two broad steps, (a) selecting the next best psg from the pool of candidate psgs, and then (b) keeping on expanding the explicit graph till the next best solution is found, is continued till $k$ solutions are found.

| # disks | 100 solutions | | | | 300 solutions | | | | 500 solutions | | | |
|---|---|---|---|---|---|---|---|---|---|---|---|---|
| | *ASG* | | *LASG* | | *ASG* | | *LASG* | | *ASG* | | *LASG* | |
| | avg. | max. | avg. | max. | avg. | max. | avg. | max. | avg. | max. | avg. | max. |
| 8 | 55 | 92 | 41 | 68 | 111 | 186 | 91 | 174 | 174 | 375 | 135 | 235 |
| 9 | 66 | 122 | 42 | 71 | 163 | 331 | 119 | 252 | 265 | 484 | 198 | 382 |
| 10 | 109 | 183 | 53 | 79 | 216 | 367 | 142 | 283 | 345 | 693 | 234 | 447 |
| 11 | 132 | 218 | 76 | 140 | 296 | 611 | 177 | 373 | 486 | 882 | 291 | 558 |
| 12 | 219 | 385 | 85 | 147 | 473 | 776 | 234 | 492 | 668 | 1200 | 404 | 724 |
| 13 | 259 | 482 | 118 | 200 | 675 | 1240 | 252 | 437 | 1016 | 1828 | 377 | 697 |

Table 21: Average and maximum length of **Open** while generating 100, 300, and 500 solutions for 5-peg Tower of Hanoi problem with different number of disks

| # matrices | 10 solutions | | | | 15 solutions | | | | 20 solutions | | | |
|---|---|---|---|---|---|---|---|---|---|---|---|---|
| | *ASG* | | *LASG* | | *ASG* | | *LASG* | | *ASG* | | *LASG* | |
| | avg. | max. | avg. | max. | avg. | max. | avg. | max. | avg. | max. | avg. | max. |
| 20 | 46 | 87 | 25 | 39 | 68 | 121 | 34 | 59 | 90 | 176 | 46 | 95 |
| 30 | 84 | 162 | 71 | 126 | 123 | 230 | 94 | 157 | 160 | 293 | 116 | 192 |
| 40 | 73 | 123 | 58 | 90 | 98 | 182 | 73 | 129 | 125 | 226 | 90 | 152 |
| 50 | 86 | 151 | 75 | 126 | 120 | 211 | 100 | 169 | 151 | 266 | 123 | 205 |
| 60 | 91 | 144 | 76 | 112 | 118 | 189 | 94 | 137 | 151 | 267 | 108 | 160 |
| 70 | 136 | 234 | 85 | 122 | 188 | 329 | 103 | 147 | 243 | 437 | 117 | 170 |
| 80 | 181 | 324 | 94 | 132 | 258 | 469 | 112 | 157 | 335 | 607 | 127 | 180 |
| 90 | 226 | 414 | 103 | 142 | 328 | 609 | 122 | 167 | 427 | 777 | 136 | 190 |
| 100 | 307 | 576 | 167 | 259 | 445 | 823 | 216 | 337 | 583 | 1145 | 262 | 477 |

Table 22: Average and maximum length of **Open** while generating 10, 15, and 20 solutions for *matrix-chain multiplication* problems

| Test case | 5 solutions | | | | 10 solutions | | | | 15 solutions | | | |
|---|---|---|---|---|---|---|---|---|---|---|---|---|
| | *ASG* | | *LASG* | | *ASG* | | *LASG* | | *ASG* | | *LASG* | |
| | avg. | max. | avg. | max. | avg. | max. | avg. | max. | avg. | max. | avg. | max. |
| TC1 | 45 | 84 | 41 | 74 | 93 | 176 | 75 | 125 | 135 | 249 | 95 | 143 |
| TC2 | 50 | 95 | 50 | 95 | 100 | 192 | 94 | 170 | 146 | 266 | 125 | 197 |
| TC3 | 47 | 90 | 46 | 89 | 90 | 168 | 82 | 142 | 132 | 244 | 115 | 210 |
| TC4 | 50 | 93 | 49 | 90 | 101 | 194 | 87 | 155 | 152 | 292 | 119 | 197 |
| TC5 | 47 | 86 | 45 | 74 | 98 | 186 | 87 | 149 | 140 | 246 | 114 | 184 |
| TC6 | 49 | 93 | 47 | 84 | 105 | 200 | 95 | 168 | 155 | 294 | 127 | 206 |
| TC7 | 42 | 81 | 42 | 80 | 83 | 157 | 73 | 119 | 121 | 231 | 92 | 138 |
| TC8 | 46 | 89 | 44 | 84 | 97 | 188 | 86 | 159 | 144 | 277 | 120 | 214 |
| TC9 | 40 | 77 | 39 | 73 | 80 | 147 | 70 | 119 | 115 | 214 | 93 | 146 |
| TC10 | 59 | 116 | 59 | 113 | 128 | 251 | 116 | 212 | 189 | 350 | 161 | 280 |
| TC11 | 55 | 106 | 54 | 105 | 115 | 225 | 110 | 211 | 171 | 317 | 166 | 321 |
| TC12 | 33 | 64 | 31 | 51 | 67 | 116 | 55 | 98 | 95 | 172 | 78 | 135 |
| TC13 | 51 | 98 | 51 | 97 | 103 | 193 | 100 | 185 | 149 | 276 | 140 | 239 |
| TC14 | 41 | 78 | 40 | 73 | 82 | 154 | 69 | 112 | 120 | 231 | 97 | 176 |

Table 23: Average and maximum length of **Open** while generating 5, 10, and 15 solutions for generating *secondary structure of RNA sequences*

It is important to observe that both methods heavily depend on incorporating the updates in the explicit DAG like adding nodes, increase in the cost, etc., and recomputing the associated swap options along with the signatures that use those swap options. Handling dynamic updates in the DAG efficiently and its use in implicit AND/OR search spaces remains an interesting future direction.

## 7. Conclusion

In our work we have presented a top-down algorithm for generating solutions of a given weighted AND/OR structure (DAG) in non-decreasing order of cost. Ordered solutions for AND/OR DAGs are useful for a number of areas including model based programming, developing new variants of AO*, service composition based on user preferences, real life problems having dynamic programming formulation, etc. Our proposed algorithm has two advantages – (a) it works incrementally, i.e., after generating a specific number of solutions, the next solution is generated quickly, (b) if the number of solutions to be generated is known a priori, our algorithm can leverage that to generate solutions faster. Experimental results show the efficacy of our algorithm over the state-of-the-art. This also opens up several interesting research problems and development of applications.

## 8. Acknowledgments

We thank the anonymous reviewers and the editor, Prof. Hector Geffner, for their valuable comments which have enriched the presentation of the paper significantly. We also thank Prof. Abhijit Mitra, International Institute of Information Technology, Hyderabad, India, for his valuable inputs regarding the test domain involving *secondary structure of RNA*. We thank Aritra Hazra and Srobona Mitra, Research Scholar, Department of Comp. Sc. & Engg., Indian Institute of Technology Kharagpur, India, for proof reading the paper.

## Appendix A. Proof of Correctness of Algorithm 4

**Lemma A.1** *Every solution other than the optimal solution $S_{opt}$ can be constructed from $S_{opt}$ by applying a sequence of swap options according to the order $\hat{\mathcal{R}}$.*

Proof: **[Lemma A.1]** Every solution other than $S_{opt}$ of an alternating AND/ OR tree $\hat{T}_{\alpha\beta}$ is constructed by choosing some non optimal edges at some OR nodes. Consider any other solution $S_m$, corresponding to which the set of non-optimal OR edges is $\mathcal{S}_\beta$ and suppose $|\mathcal{S}_\beta| = m$. We apply the relation $\mathcal{R}$ to $\mathcal{S}_\beta$ to obtain an ordered sequence $\Sigma$ of OR edges where $\forall e_1, e_2 \in \Sigma$, $e_1$ appears before $e_2$ in $\Sigma$ if $(e_1, e_2) \in \mathcal{R}$. We show that there exists a sequence $\hat{\Sigma}$ of swap options that can be constructed for $\mathcal{S}_\beta$. For every OR edge $e_{i_j}$ of $\Sigma$ (here $e_{i_j}$ is the $i^{th}$ edge of $\Sigma$ and $1 \leq i \leq m$), we append the subsequence of OR edges $e_{i_1}, \ldots, e_{i_{j-1}}$ before $e_{i_j}$, where $e_{i_1}, \ldots, e_{i_j}$ are the OR edges that emanate from the same parent $v_q$, and $e_{i_1}, \ldots, e_{i_{j-1}}$ are the first $i_j - 1$ edges in $L(v_q)$.

We get a sequence of OR edges $\Sigma_{aug}$ from $\Sigma$ by the above mentioned augmentation. $\Sigma_{aug}$ is basically a concatenation of subsequences $\tau_1, \ldots, \tau_m$, where $\tau_i$ is a sequence of edges $e_{i_1}, \ldots, e_{i_j}$ such that $e_{i_1}, \ldots, e_{i_j}$ are the OR edges that emanate from the same parent $v_q$, and $e_{i_1}, \ldots, e_{i_j}$ are the first $i_j$ edges in $L(v_q)$. We construct $\hat{\Sigma}$ from $\Sigma_{aug}$ as follows. From every $\tau_i$, we construct $\hat{\tau}_i = \langle \sigma_{i_1, i_2}, \ldots, \sigma_{i_{j-1}, i_j} \rangle$, where $\sigma_{i_k, i_{k+1}} = \langle e_{i_k}, e_{i_{k+1}}, \delta_{i_k, i_{k+1}} \rangle$ and $i_1 \leq i_k \leq (i_j - 1)$. $\hat{\Sigma}$ is constructed by concatenating every individual $\hat{\tau}_i$. Hence there exists a sequence of swap options $\hat{\Sigma}$ corresponding to every other solution $S_m$. □

**Definition A.p [Default Path]** From Lemma A.1, every non-optimal solution $S_m$ can be constructed from the initial optimal solution by applying a sequence of swap options, $\hat{\Sigma}(S_m)$, according to the order $\hat{\mathcal{R}}$. The sequence of solutions that is formed following $\hat{\Sigma}(S_m)$ corresponds to a path from $S_{opt}$ to $S_m$ in SSDAG $\mathcal{G}^s$. This path is defined as the *default path*, $\mathcal{P}_d(S_m)$, for $S_m$.

**Lemma A.2** *The* SSDAG *of an alternating AND/OR tree $\hat{T}_{\alpha\beta}$ contains every alternative solution of $\hat{T}_{\alpha\beta}$.*

Proof: **[Lemma A.2]** We prove this by induction on the length of the default path $\mathcal{P}_d$ of the solutions.

**[Basis (n = 1) :]** Consider the swap list of $S_{opt}$. The solutions whose default path length is equal to 1 form the Succ($S_{opt}$). Therefore these solutions are present in $\mathcal{G}$.

**[Inductive Step :]** Suppose the solutions whose default path length is less than or equal to $n$ are present in $\mathcal{G}$. We prove that the solutions having default path length equal to $n+1$ are also present in $\mathcal{G}$. Consider any solution $S_m$ where $\mathcal{P}_d(S_m) = n+1$. Let $\hat{\Sigma}(S_m) = \langle \sigma_1, \cdots, \sigma_n, \sigma_{n+1} \rangle$. Consider the solution $S'_m$ where $\hat{\Sigma}(S'_m) = \langle \sigma_1, \cdots, \sigma_n \rangle$. Since $\mathcal{P}_d(S'_m) = n$, $S'_m \in \mathcal{V}$, and swap option $\sigma_{n+1} \in \mathcal{L}(S'_m)$, there is a directed edge from $S'_m$ to $S_m$ in $\mathcal{G}^s$. Hence every solution having a default path length equal to $n+1$ is also present in $\mathcal{G}$. □

**Lemma A.3** *For any alternating AND/OR tree $\hat{T}_{\alpha\beta}$, Algorithm 4 adds solutions to* Closed *(at Line 11) in non-decreasing order of cost.*

Proof: [**Lemma A.3**] Consider the following invariants of Algorithm 4 that follow from the description of Algorithm 4.

    a. The minimum cost solution from Open is always removed at Line 6 of Algorithm 4.

    b. The cost of the solutions that are added in Open, while exploring the successor set of a solution $S_m$ (at Line 13 of Algorithm 4), are greater than or equal to $C(S_m)$.

From these two invariants it follows that Algorithm 4 adds solutions to Closed (at Line 11) in non-decreasing order of cost.

**Lemma A.4** *For any alternating AND/OR tree $\hat{T}_{\alpha\beta}$, for every node of the SSDAG of $\hat{T}_{\alpha\beta}$, Agorithm 4 generates the solution corresponding to that node.*

Proof: **[Lemma A.4]** From Lemma A.3 it follows that Algorithm 4 generates the solutions in the non-decreasing order of cost. By *generating* a solution $S_m$, we mean adding $S_m$ to Closed (at line 11 of Algorithm 4). For the purpose of proof by contradiction, let us assume that Algorithm 4 does not generate solution $S_m$. Also let $S_m$ be the first occurrence of this scenario while generating solutions in the mentioned order. According to Lemma A.1, there exists a sequence of swap options $\hat{\Sigma} = \sigma_1, \ldots, \sigma_k$ corresponding to $S_m$. Also consider the solution $S'_m$ whose sequence of swap options is $\hat{\Sigma}' = \sigma_1, \ldots, \sigma_{k-1}$. According to Property 3.2, $C(S'_m) \leq C(S_m)$. Consider the following two cases:

a. $C(S'_m) < C(S_m)$: Since $S_m$ is the first instance of the incorrect scenario, and Algorithm 4 generates the solutions in the non-decreasing order of cost, $S'_m$ is generated prior to $S_m$.

b. $C(S'_m) = C(S_m)$: Since Algorithm 4 resolves the tie in the favor of the parent solution, and $S_m$ is the first instance of the incorrect scenario – in this case also $S'_m$ will be generated prior to $S_m$.

The swap option $\sigma_k$ belongs to the swap list of $S'_m$. When $S'_m$ was generated by Algorithm 4, that is, when $S'_m$ was added to Closed, $S'_m$ was also expanded and the solutions which can be constructed from $S'_m$ applying one swap option, were added to the Open list. Since $S_m$ was constructed from $S'_m$ applying one swap option $\sigma_k$, $S_m$ was also added to the Open while exploring the successors of $S'_m$. Therefore $S_m$ will also be eventually generated by Algorithm 4 - a contradiction. □

**Lemma A.5** *For any alternating AND/OR tree $\hat{T}_{\alpha\beta}$, Algorithm 4 does not add any solution to* Closed *(at Line 11 of Algorithm 4) more than once.*

Proof: [**Lemma A.5**]  For the purpose of contradiction, let us assume that $S_m$ is the first solution that is added to Closed twice. Therefore $S_m$ must have been added to Open twice. Consider the following facts.

a. When $S_m$ was added to Closed for the first time, the value of $lastSolCost$ was $C(S_m)$, and $S_m$ was added to TList.

b. From the description of Algorithm 4 it follows that the contents of TList are deleted only when the value of $lastSolCost$ increases.

c. From Lemma A.3 it follows that Algorithm 4 generates the solutions in non-decreasing order of cost. Hence, when $S_m$ was generated for the second time, the value of $lastSolCost$ did not change from $C(S_m)$.

From the above facts it follow that $S_m$ was present in TList when $S_m$ was added to Open for the second time. Since, while adding a solution to Open, Algorithm 4 checks whether it is present in TList (at Line 16 of Algorithm 4); Algorithm 4 must had done the same while adding $S_m$ to Open for the second time. Therefore $S_m$ could not be added Open for the second time – a contradiction. □

**Theorem A.1** *$\forall S_j \in \mathcal{V}$, $S_j$ is generated (at Line 11) by Algorithm 4 only once and in the non-decreasing order of costs while ties among the solutions having same costs are resolved as mentioned before.*

Proof: [**Theorem A.1**]  Follows from Lemma A.2, Lemma A.3, Lemma A.4 and Lemma A.5.
□

## Appendix B. Proof of Correctness of Algorithm 5

**Definition B.q [Reconvergent Paths in Solution Space DAG]** Two paths, (*i*) $p_1 = S_{i_1}^1 \to \cdots \to S_{i_n}^1$ and (*ii*) $p_2 = S_{i_1}^2 \to \cdots \to S_{i_m}^2$, in the SSDAG $\mathcal{G}^s$ of an alternating AND/OR tree $\hat{T}_{\alpha\beta}$ are reconvergent if the following holds:

a. $S_{i_1}^1 = S_{i_1}^2$, i.e. the paths start from the same node;

b. $S_{i_n}^1 = S_{i_m}^1$, i.e. the paths ends at the same node;

c. $(\forall j \in [2, n-1])(\forall k \in [2, m-1]), \left(S_{i_j}^1 \neq S_{i_k}^2\right)$; i.e. the paths do not have any common intermediate node.

**Definition B.r [Order on Generation Time]** In the context of Algorithm 5, we define an order relation, $\prec_t \subset \mathcal{V} \times \mathcal{V}$, where $(S_p, S_q) \in \prec_t$ if $S_p$ is generated by Algorithm 5 before $S_q$. Here $\mathcal{V}$ is set of vertices in SSDAG $\mathcal{G}^s$ of an alternating AND/OR tree $\hat{T}_{\alpha\beta}$.

**Lemma B.1** *Algorithm 5 adds the solutions to the* Closed *list in the non-decreasing order of costs.*

**Proof: [Lemma B.1]** Consider the following invariants of Algorithm 5 that follow from the description of Algorithm 5.

a. The minimum cost solution from Open is always removed at line 11 of Algorithm 4.

b. Algorithm 5 expands any solution, say $S_p$, in two phases. At the first phase $S_p$ is expanded using the native swap options of $S_p$. The solutions that are added to Open as a result of the application of the native swap options, will have cost greater than or equal to $C(S_p)$. In the second phase, i.e., during *lazy expansion*, $S_p$ is again expanded using a non native swap option. A solution $S_p$ may undergo the second phase $\kappa$ times where $0 \leq \kappa \leq (|\mathcal{L}(S_p)| - |\mathcal{N}(S_p, \sigma_k)|)$ and $\sigma_k$ is used to construct $S_p$. In every lazy expansion of $S_p$, a new solution is added to Open. Consider a solution $S_m$ which is constructed from $S_m'$ using $\sigma_j$ by Algorithm 5 where $S_m' \in Pred(S_m)$. Suppose swap option $\sigma_i \in \mathcal{L}(S_m)$, and $\sigma_i \notin \mathcal{N}(S_m, \sigma_j)$, i.e., $\sigma_i$ is not a native swap option of $S_m$. Clearly $\sigma_i \in \mathcal{L}(S_m')$. Suppose $S_c$ and $S_c'$ are the successors of $S_m$ and $S_m'$ respectively, constructed by the application of $\sigma_i$, i.e., $S_m' \xrightarrow{\sigma_i} S_c'$, and $S_m \xrightarrow{\sigma_i} S_c$. Also let $S_c'$ is added to Closed after $S_m$.

Consider the fact that Algorithm 5 does not apply swap option $\sigma_i$ to $S_m$, that is, $S_c$ is not added to Open until $S_c'$ is added to Closed. Since $C(S_m') \leq C(S_m)$, $C(S_c') \leq C(S_c)$. According to Algorithm 5, $\sigma_i$ is applied to $S_m$ (during the lazy expansion), and $S_c$ is added to Open right after $S_c'$ is added to Closed. Consider the time period between adding $S_m$ and adding $S_c'$ to Closed. During that period, every solution that is added to Closed has cost between $C(S_m)$ and $C(S_c')$, i.e., the cost is less or equal to $C(S_c)$. In general, the application of a swap option to add a solution to Open is delayed by such an amount of time, say $\Delta$, so that all the solutions, which are added to Closed during this $\Delta$ time interval, have cost less than or equal to the solution under consideration.

From the above facts it follow that Algorithm 5 adds the solutions to the **Closed** list in non-decreasing order of costs. □

**Lemma B.2** *Any two reconvergent paths in the SSDAG $\mathcal{G}^s$ of an alternating AND/OR tree $\hat{T}_{\alpha\beta}$ are of equal length.*

Proof: [**Lemma B.2**]  Consider the paths:

$$(i)\ p_1 = S_1 \xrightarrow{\sigma_1} S_p \xrightarrow{\sigma_2} \cdots \xrightarrow{\sigma_n} S_n, \text{ and } (ii)\ p_2 = S_1 \xrightarrow{\sigma'_1} S'_p \xrightarrow{\sigma'_2} \cdots \xrightarrow{\sigma'_m} S_n.$$

The edges in the paths represent the application of a swap option to a solution. Now $p_1$ and $p_2$ start from the same solution and also end at the same solution. Therefore the sets of swap options that are used in these paths are also same. Hence the lengths of those paths are equal, that is, in the context of $p_1$ and $p_2$, $n = m$.

**Lemma B.3** *For any set of reconvergent paths of any length $n$, Algorithm 5 generates at most one path.*

Proof: [**Lemma B.3**]  The following cases are possible.

[**Case 1** ($n = 2$) **:**]  Consider the following two paths:

$$(i)\ p_1 = S_1 \xrightarrow{\sigma_1} S_2 \xrightarrow{\sigma_2} S_3, \text{ and } (ii)\ p_2 = S_1 \xrightarrow{\sigma'_1} S'_2 \xrightarrow{\sigma'_2} S_3.$$

It is obvious that $\sigma_1 = \sigma'_2$ and $\sigma_2 = \sigma'_1$. Suppose $S_2 \prec_t S'_2$. Here Algorithm 5 does not apply the swap option $\sigma_1$ to $S'_2$. Therefore $p_2$ is not generated by Algorithm 5.

[**Case 2 (Any other values of $n$) :**]  In this case, any path belonging to the set of reconvergent paths, consists of $n$ different swap options, suppose $\sigma_1, \cdots, \sigma_n$. Also the start node and the end node of the paths under consideration are $S_p$ and $S_m$. Consider the nodes in the paths having length 1 from $S_p$. Clearly there can be $n$ such nodes.

Among those nodes, suppose Algorithm 5 adds $S_{p1}$ to **Closed** first, and $S_{p1}$ is constructed from $S_p$ by applying swap option $\sigma_1$. According to Algorithm 5, $\sigma_1$ will not be applied to any other node that is constructed from $S_p$ and is added to **Closed** after $S_{p1}$. Therefore, all those paths starting from $S_p$, whose second node is not $S_{p1}$, will not be generated by Algorithm 5. We can use the similar argument on the paths from $S_{p1}$ to $S_m$ of length $n-1$ to determine the paths which will not be generated by Algorithm 5. At each stage, a set of paths will not be grown further, and at most one path towards $S_m$ will continue to grow. After applying the previous argument $n$ times, at most one path from $S_p$ to $S_m$ will be constructed. Therefore Algorithm 5 will generate at most one path from $S_p$ to $S_m$. □

Definition B.s [**Connection Relation $\mathcal{R}_c$ and $\hat{\mathcal{R}}_c$**]  We define *connection* relation, $\mathcal{R}_c$, a symmetric order relation for a pair of OR nodes, $v_q$ and $v_r$, belonging to an alternating AND/OR tree $\hat{T}_{\alpha\beta}$ as:

$$(v_q, v_r) \in \mathcal{R}_c \mid \text{ if in } \hat{T}_{\alpha\beta} \text{ there exists an AND node } v_p, \text{ from which}$$
$$\text{there exist two paths, } (i)\ p_1 = v_p \rightarrow \ldots \rightarrow v_q, \text{ and}$$
$$(ii)\ p_2 = v_p \rightarrow \ldots \rightarrow v_r$$

Similarly the *connection* relation, $\hat{\mathcal{R}}_c$, is defined between two swap options as follows. Consider two swap options $\sigma_{iq}$ and $\sigma_{jr}$, where $\sigma_{iq} = \langle e_i, e_q, \delta_{iq} \rangle$ and $\sigma_{jr} = \langle e_j, e_r, \delta_{jr} \rangle$. Suppose OR edges $e_i$ and $e_q$ emanate from $v_p$, and OR edges $e_j$ and $e_r$ emanate from $v_t$. Now $(\sigma_{iq}, \sigma_{jr}) \in \hat{\mathcal{R}}_c$ if $(v_p, v_t) \in \mathcal{R}_c$.

**Definition B.t [Mutually Connected Set]**  For a solution $S_m$, a set $V_m$ of OR nodes is *mutually connected*, if

$$\forall v_1, v_2 \in V_m, \ \big((v_1 \neq v_2) \Rightarrow \{(v_1, v_2) \in \mathcal{R}_c\}\big)$$

Consider the set of OR nodes, $V_m = \{v_1, \cdots, v_k\}$, where swap option $\sigma_j$ belongs to $v_j$ and $1 \leq j \leq k$. Here the set of swap options $\hat{V}_m = \{\sigma_1, \cdots, \sigma_k\}$ is *mutually connected*.

**Lemma B.4**  *Suppose $S_m$ is a solution of an alternating AND/OR tree $\hat{T}_{\alpha\beta}$, $Pred(S_m) = \{S_1, \cdots, S_k\}$, and swap option $\sigma_j$ is used to construct $S_m$ from $S_j$ where $1 \leq j \leq k$. The swap options $\sigma_1, \cdots, \sigma_k$ are mutually connected.*

**Proof: [Lemma B.4]**  Since $S_m$ is constructed from $S_1, \cdots, S_k$ by applying $\sigma_1, \cdots, \sigma_k$ respectively, $\sigma_1, \cdots, \sigma_k$ are present in the signature of $S_m$. Suppose set $s_\sigma = \{\sigma_1, \cdots, \sigma_k\}$. We have to show that

$$\forall \sigma_a, \sigma_b \in s_\sigma, \ \big((\sigma_a, \sigma_b) \in \hat{\mathcal{R}}_c\big)$$

For the purpose of proof by contradiction, let us assume $(\sigma_{i_1}, \sigma_{i_2}) \notin \hat{\mathcal{R}}_c$. Also $S_m$ is constructed by applying $\sigma_{i_1}$ and $\sigma_{i_2}$ to $S_{i_1}$ and $S_{i_2}$ respectively. Consider the path $p_1$ in SSDAG of $\hat{T}_{\alpha\beta}$ which starts from $S_{opt}$ and ends at $S_m$, and along $p_1$, $S_{i_1}$ is the parent of $S_m$. Now along this path, $\sigma_{i_2}$ is applied before the application of the swap option $\sigma_{i_1}$. Similarly consider the path $p_2$ in SSDAG of $\hat{T}_{\alpha\beta}$ which starts from $S_{opt}$ and ends at $S_m$, and along $p_2$, $S_{i_2}$ is the parent of $S_m$. Along this path, $\sigma_{i_1}$ is applied before the application of the swap option $\sigma_{i_2}$.

Suppose $\sigma_{i_1}$ and $\sigma_{i_2}$ belongs to OR node $v_1$ and $v_2$ respectively. Since along path $p_1$, $\sigma_{i_1}$ is the swap option which is applied last, $S_m$ contains node $v_1$. Similarly along path $p_2$, $\sigma_{i_2}$ is the swap option which is applied last. Hence $S_m$ contains node $v_2$. Therefore, there must be an AND node $v_r$ in $\hat{T}_{\alpha\beta}$, from which there exist paths to node $v_1$ and $v_2$ – implies that $(\sigma_{i_1}, \sigma_{i_2}) \in \hat{\mathcal{R}}_c$. We arrive at a contradiction that proves $\sigma_1, \cdots, \sigma_k$ are *mutually connected*. □

**Definition B.u [Subgraph of SSDAG]**  Consider a solution $S_p$ of an alternating AND/OR tree $\hat{T}_{\alpha\beta}$ and mutually connected set $V_m$ of OR nodes in $S_p$, where $\forall v_q \in V_m, \ \big(C(S_p, v_q) = C_{opt}(v_q)\big)$. The subgraph $\mathcal{G}^s_{sub}(S_p, V_m) = \langle \mathcal{V}_{sub}, \mathcal{E}_{sub} \rangle$ of the SSDAG with respect to $S_p$ and $V_m$ is defined as follows. $\mathcal{V}_{sub}$ consists of only those solutions which can be constructed from $S_p$ by applying a sequence of swap options belonging to $V_m$, and $\mathcal{E}_{sub}$ is the set of edges corresponding to the swap options that belong to $V_m$.

**Lemma B.5**  *The number of total possible distinct solutions at each level $d$ in $\mathcal{G}^s_{sub}(S_p, V_m)$ is $\binom{n+d-2}{n-1}$, where $|V_m| = n$.*

**Proof: [Lemma B.5]** Consider the swap options that belong to the nodes in $V_m$. With respect to these swap options, every solution $S_r$ in $\mathcal{G}^s_{sub}(S_p, V_m)$ is represented by a sequence of numbers of length $n$, $Seq(S_r)$, where every number corresponds to a distinct node in $V_m$. The numerical value of a number represent the rank of the swap option that is chosen for a node $v_q \in V_m$. According to the representation, at each level:

i. the sum of numbers in $Seq(S_r)$ of a solution, $S_r$, is equal to the sum of numbers in $Seq(S'_r)$ of any other solution, $S'_r$, in that same level;

ii. the sum of numbers in $Seq(S_r)$ of a solution, $S_r$, is increased by 1 from the sum of numbers in $Seq(S''_p)$ of any solution, $S''_p$, of the previous level.

Hence, at the $d^{th}$ level, there are $n$ slots and $d - 1$ increments that need to be made to $Seq(S_r)$. This is an instance of the well known combinatorial problem of packing $n + d - 1$ objects in $n$ slots with the restriction of keeping at least one object per slot. This can be done in $\binom{n+d-2}{n-1}$ ways. □

**Theorem B.1** *The solution space tree constructed by Algorithm 5 is* complete.

**Proof: [Theorem B.1]** For the purpose of contradiction, suppose $S_m$ is the first solution which is not generated by Algorithm 5. Also $Pred(S_m) = \{S_{p_i}\}$ and $S_m$ can be constructed from $S_{p_i}$ by applying $\sigma_{q_i}$, where $1 \leq i \leq k$. From Lemma B.4 it follows that the set of swap options $\{\sigma_{q_i} \mid 1 \leq i \leq k\}$ is mutually connected. Therefore the set of OR nodes $V_m$ to which the swap options belong is also mutually connected. Suppose $|V_m| = n$.

Consider the solution $S_q$, where $V_m$ is mutually connected, and for $1 \leq i \leq k$, every $\sigma_{q_i}$ belongs to the set of native swap options of $S_q$ with respect the swap option that is used to construct $S_q$. Clearly

$$\forall v_t \in V_m, \; \big( C(S_q, v_t) = C_{opt}(v_t) \big)$$

We argue that $S_q$ is generated by Algorithm 5 because $S_m$ is the first solution which is not generated by Algorithm 5. Consider the subtree $\mathcal{T}^s_{sub}$ of $\mathcal{T}^s$ rooted at $S_q$, where only the edges corresponding to swap options that belong to $V_m$ are considered. Now we prove that the number of solutions generated by Algorithm 5 at every level of $\mathcal{T}^s_{sub}$ is equal to the number of solutions at the same level in $\mathcal{G}^s_{sub}(S_q, V_m)$.

Consider the solution $S_q$ and the set $Succ(S_q)$. Suppose $Succ(S_q, V_m)$ is the set of successor solutions that are constructed from $S_q$ by applying the swap options belonging to the nodes in $V_m$, and $S'_{min}$ is the minimum cost solution in $Succ(S_q, V_m)$. According to Algorithm 5 initially $Succ(S'_{min})$ is partially explored by using the set of native swap options of $S'_{min}$. Any other non native swap option, $\sigma_b$, that belongs to the nodes in $V_m$, is used to explore $Succ(S'_{min})$, right after the sibling solution of $S'_{min}$, constructed by applying $\sigma_b$ to $S_q$, is added to **Closed**. Consider the fact that for solution $S_q$, $\forall v_t \in V_m, \; \big( C(S_q, v_t) = C_{opt}(v_t) \big)$ holds. Therefore all the swap options belonging to $V_m$ will also be eventually used to explore the successors of $S'_{min}$. Similarly the second best successor of $S_q$ will be able use all but one swap option, $\sigma_c$, which is used to construct $S'_{min}$.

The immediate children of $S'_{min}$ in $\mathcal{T}^s_{sub}$ will consist of all solutions, that can be obtained by the application of one swap option in $V_m$ to $S'_{min}$. The native swap list of $S'_{min}$ contains the swap option ranking next to $\sigma_c$. The swap options, that are used to construct the other $n-1$ sibling solutions of $S'_{min}$, will be used again during lazy expansion, which accounts for another $n-1$ children of $S'_{min}$. Hence there would be $n$ children of $S'_{min}$.

Similarly, the second best successor of $S_q$ in $\mathcal{T}^s_{sub}$ will have $n-1$ immediate children. The third best successor of $S_q$ in $\mathcal{T}^s_{sub}$ will have $n-2$ children and so on. Now the children of these solutions will again have children solutions of their own, increasing the number of solutions at each level of the tree. This way, with each increasing level, the number of solutions present in the level keeps increasing. We prove the following proposition as a part of proving Theorem B.1.

**Proposition B.1** *At any level* $d$, *the number of solutions* $N(d, n, \mathcal{T}^s_{sub})$ *is given by*

$$N(d, n, \mathcal{T}^s_{sub}) = \sum_{k=1}^{n} N(d-1, k, \mathcal{T}^s_{sub}) = \binom{n+d-2}{n-1}$$

Proof: [**Proposition B.1**] At second level, there are $n$ solutions. These give rise to $\sum_{k=1}^{n} k$ solutions at third level. Similarly at fourth level we have

$$\sum_{k=1}^{n} k + \sum_{k=1}^{n-1} k + \sum_{k=1}^{n-2} k \dots + 1 = N(3, n, \mathcal{T}^s_{sub}) + N(3, n-1, \mathcal{T}^s_{sub}) + \dots + 1$$

We can extend this to any level $d$ and the result is as follows.

$$\begin{aligned}
N(1, n, \mathcal{T}^s_{sub}) &= 1 \\
N(2, n, \mathcal{T}^s_{sub}) &= n \\
N(3, n, \mathcal{T}^s_{sub}) &= \sum_{k=1}^{n} k = \binom{n+1}{2} \\
N(4, n, \mathcal{T}^s_{sub}) &= \sum_{k=1}^{n} N(3, k, \mathcal{T}^s_{sub}) = \binom{n+2}{3}
\end{aligned}$$

We determine the number of solutions at any level of $\mathcal{T}^s_{sub}$ by induction on the depth $d$.

[**Basis (d = 1) :**]   Clearly, $N(1, n, \mathcal{T}^s_{sub}) = n$.

[**Inductive Step :**]   Suppose, at $d^{th}$ level the number of solutions is $\binom{n+d-2}{n-1} = \binom{n+d-2}{d-1}$. Therefore at $d + 1^{th}$ level,

$$N(d+1, n, \mathcal{T}^s_{sub}) = \sum_{k=1}^{n} N(d, k, \mathcal{T}^s_{sub}) = \binom{n+d-2}{d-1} + \binom{n+d-3}{d-1} + \dots + 1 = \binom{n+d-1}{n-1}$$

Since Algorithm 5 does not generate duplicate node, and from Proposition B.1 the number of solutions in $\mathcal{G}^s_{sub}(S_q, V_m)$ at any level is equal to the number of solutions in that level of $\mathcal{T}^s_{sub}$, at any level the set of solutions in $\mathcal{G}^s_{sub}(S_q, V_m)$ is also generated by Algorithm 5 through $\mathcal{T}^s_{sub}$. Therefore, the level, at which $S_m$ belongs in $\mathcal{G}^s_{sub}(S_q, V_m)$, will also be generated by Algorithm 5. Therefore $S_m$ will also be generated by Algorithm 5 – a contradiction which establishes the truth of the statement of Theorem B.1.   □

## Appendix C. Conversion between AND/OR Tree and Alternating AND/OR Tree

An AND/OR tree is a generalization of alternating AND/OR tree where the restriction of strict alternation between AND and OR nodes are relaxed. In other words an intermediate OR node can be a child of another intermediate OR node and the similar parent child relation is also allowed for AND node. We present an algorithm to convert an AND/OR to an equivalent alternating AND/OR tree.

We use two operations namely, *folding* and *unfolding* for the conversions. Corresponding to every edge, a stack, *update-list*, is used for the conversions. In an AND/OR tree, consider two nodes, $v_q$ and $v_r$, of similar type (AND/OR) and they are connected by an edge $e_r$. Edges, $e_1, \cdots, e_k$ emanate from $e_r$.

[**Folding OR Node :**]   Suppose $v_q$ and $v_r$ are OR nodes. The *folding* of $v_r$ is performed as follows.

- The source of the edges $e_1, \cdots, e_k$ are changed from $v_r$ to $v_q$ and the costs are updated as $c_e(e_i) \leftarrow c_e(e_i) + c_e(e_r) + c_v(v_r)$ where $1 \leq i \leq k$, that is the new cost is the sum of the old cost and the cost of the edge that points to the source of $e_i$. The triplet $\langle v_r, c_v(v_r), c_e(e_r) \rangle$ is pushed into the *update-list* of $e_i$, $1 \leq i \leq k$.

- The edge $e_r$ along with node $v_r$ is removed from $v_q$.

[**Folding AND Node :**]   Suppose $v_q$ and $v_r$ are AND nodes. The *folding* of $v_r$ is performed as follows.

- The source of the edges $e_1, \cdots, e_k$ are changed from $v_r$ to $v_q$. One of the edges among $e_1, \cdots, e_k$, suppose $e_i$, is selected arbitrarily and the cost is updated as $c_e(e_i) \leftarrow c_e(e_i) + c_e(e_r) + c_v(v_r)$ where $1 \leq i \leq k$. The triplet $\langle v_r, c_v(v_r), c_e(e_r) \rangle$ is pushed into the *update-list* of $e_i$, whereas the triplet $\langle v_r, 0, 0 \rangle$ is pushed into the *update-list* of $e_j$, where $1 \leq j \leq k$ and $j \neq i$.

- The edge $e_r$ along with node $v_r$ is removed from $v_q$.

The *unfolding* operation is the reverse of the folding operation and it is same for both OR and AND nodes. It works on a node $v_q$ as follows.

---

**Procedure** `Unfold`(*node $v_q$*)

| | |
|---|---|
| **1** | **forall** *edge $e_i$ that emanate from $v_q$* **do** |
| **2** |     **if** *the update list of $e_i$ is not empty* **then** |
| **3** |         $\langle v_t, c_1, c_2 \rangle \leftarrow$ pop(update list of $e_i$); |
| **4** |         **if** *there exists no edge $e_t$ from $v_q$ that points to the node $v_t$* **then** |
| **5** |             Create a node $v_t$, and connect $v_t$ using edge $e_t$ from $v_q$; |
| **6** |             $c_v(v_t) \leftarrow c_1$; |
| **7** |             $c_e(e_t) \leftarrow c_2$; |
| **8** |         **else if** $c_2 \neq 0$ **then** |
| **9** |             $c_e(e_t) \leftarrow c_2$; |
| **10** |         **end** |
| **11** | **end** |

---

Function *Convert* takes the root node of AND/OR tree and transforms it to an equivalent alternating AND/OR tree recursively.

---

**Function `Convert(`$v_q$`)`**

---

**1**   **if** *every child of $v_q$ is a terminal node* **then**
**2**       **if** *$v_q$ and its parent $v_p$ are of same type* **then**
**3**           Apply *fold* operation to $v_q$;
**4**       **end**
**5**   **else**
**6**       **foreach** *child $v_r$ of $v_q$, where $v_r$ is an intermediate AND/OR node* **do**
            Convert($v_r$);
**7**   **end**

---

Function *Revert* takes the root node of an alternating AND/OR tree and converts it to the original AND/OR tree recursively.

---

**Function `Revert(`$v_q$`)`**

---

**1**   **if** *every child of $v_q$ is a terminal node* **then**
**2**       return;
**3**   Perform *unfold* operation to $v_q$;
**4**   **foreach** *child $v_r$ of $v_q$* **do**
**5**       Revert($v_r$);
**6**   **end**

---

The overall process of generating alternative solutions of an AND/OR tree is as follows. The AND/OR tree is converted to an alternating AND/OR tree using *Convert* function, and the solutions are generated using ASG algorithm. The solutions are transformed back using the *Revert* function. The proof of correctness is presented below.

### C.1 Proof of Correctness

Suppose in an AND/OR tree $T_{\alpha\beta}$ two nodes, $v_q$ and $v_r$, are of similar type (AND/OR) and they are connected by an edge $e_r$. Edges $e_1, \cdots, e_k$ emanate from $e_r$. Now *fold* operation is applied to $v_q$ and $v_r$. Let $T_{\alpha\beta}^1$ is the AND/OR tree which is generated by the application of the *fold* operation.

**Lemma C.1**  *In the context mentioned above, we present the claim of in the following two propositions.*

**Proposition C.1**  *The set of solutions of $T_{\alpha\beta}$ having node $v_q$ can be generated from the set of solutions of $T_{\alpha\beta}^1$ having node $v_q$ by applying the* unfold *operation to $v_q$ of the solutions of $T_{\alpha\beta}$.*

**Proposition C.2**  *For every solution $S_m^1$ of $T_{\alpha\beta}^1$ that contains node $v_q$, there exists a solution $S_m$ of $T_{\alpha\beta}$ that can be generated from $S_m^1$ by applying* unfold *to $v_q$.*

Proof: [**Proposition C.1**]  We present the proof for the following cases. Consider any solution of $S_m$ of $T_{\alpha\beta}$ that contains node $v_q$.

a. $v_q$ **and** $v_r$ **are OR nodes:** There are two cases possible.

    1. $v_r$ **is absent from** $S_m$**:** Since the *fold* operation modifies the edge $e_r$ only, all the other edges from $v_q$ in $T_{\alpha\beta}$ are also present in $T^1_{\alpha\beta}$. Therefore $S_m$ will also be present in the solution set of $T^1_{\alpha\beta}$ and it will remain unchanged after the application of *unfold* operation.

    2. $v_r$ **is present in** $S_m$**:** Since there are $k$ distinct OR edges emanating from $v_r$, let any one of those OR edges, say $e_i$, is present in $S_m$. We prove that there is a solution $S^1_m$ of $T^1_{\alpha\beta}$, such that the application of *unfold* operation to $S^1_m$ will generate $S_m$. The application of *fold* operation to the node $v_r$ modifies the source and the cost of edge $e_i$ from $v_r$ to $v_q$ and $c_e(e_i)$ to $c_e(e_i) + c_e(e_r) + c_v(v_r)$ in $T^1_{\alpha\beta}$. Suppose $S^1_m$ is a solution of $T^1_{\alpha\beta}$, where the edge $e_i$ is present in $S^1_m$. Also other than the subtree rooted at $v_q$, the remaining parts of $S^1_m$ and $S_m$ are identical with each other. Clearly $S^1_m$ exists as a solution of $T^1_{\alpha\beta}$ and the application of *unfold* operation to $v_q$ in $S^1_m$ generates $S_m$.

b. $v_q$ **and** $v_r$ **are AND nodes:** Since $v_q$ is an AND node $S_m$ will contain all of the AND edges that emanate from $v_q$. Therefore edge $e_r$ and $v_r$ both will be present in $S_m$. Consider the solution $S^1_m$ of $T^1_{\alpha\beta}$, such that the following holds.

    1. $v_q$ is present in $S^1_m$.

    2. The subtrees rooted at the children of $v_q$ other than $v_r$ in $S_m$ are identical with the subtrees rooted at those children of $v_q$ in $S^1_m$.

    3. Other than the subtree rooted at $v_q$, remaining parts of $S^1_m$ and $S_m$ are identical with each other.

Clearly $S^1_m$ exists as a solution of $T^1_{\alpha\beta}$ and the application of *unfold* operation to $v_q$ in $S^1_m$ generates $S_m$.

Any other solution $S'_m$ of $T_{\alpha\beta}$ that does not contain node $v_q$, is a valid solution for $T^1_{\alpha\beta}$ as well. □

Proof: [**Proposition C.2**] We present the proof for the following cases. Consider any solution $S^1_m$ of $T^1_{\alpha\beta}$ that contains node $v_q$.

a. $v_q$ **and** $v_r$ **are OR nodes:** Since $v_q$ is an OR node, exactly one OR edge $e_i$ of $v_q$ will belong to $S^1_m$. There are two cases possible.

    1. $e_i$ **was not modified while** *folding* $v_r$ **in** $T^1_{\alpha\beta}$**:** Since the *fold* operation modifies the edge $e_r$ and the OR edges of $v_r$ only, all the other edges from $v_q$ in $T_{\alpha\beta}$ are also present in $T^1_{\alpha\beta}$. Since $e_i$ was not modified during *folding*, the same solution $S^1_m$ is also a valid solution for $T_{\alpha\beta}$.

    2. $e_i$ **was modified while** *folding* $v_r$ **in** $T^1_{\alpha\beta}$**:** Suppose $e_i$ connects $v_q$ and $v_i$ in $S^1_m$. Apply the *unfold* operation to the node $v_q$ in $S^1_m$ and generate solution $S_m$. The edge $e_i$ will be replaced with edge $e_r$ which connects $v_q$ and $v_r$ and then $e_i$ will connect $v_r$ and $v_i$. We argue that $S_m$ is a valid solution of $T_{\alpha\beta}$ since the

subtree rooted at $v_i$ is not modified by the sequence of – $(a)$ the *folding* of $v_r$ to construct $T^1_{\alpha\beta}$ from $T_{\alpha\beta}$, and $(b)$ the *unfolding* of $v_q$ to construct $S_m$ from $S^1_m$.

b. $v_q$ **and** $v_r$ **are AND nodes:** Since $v_q$ is an AND node, $S^1_m$ will contain all of the AND edges that emanate from $v_q$. There are two types of AND edges emanating from $v_q$ in $T^1_{\alpha\beta}$ and they are $(a)$ *Type-1*: the edges from $v_q$ that are also present in $T_{\alpha\beta}$ from $v_q$, $(b)$ *Type-2*: the edges that are added to $v_q$ by *folding* and these edges are from $v_r$ in $T_{\alpha\beta}$. Apply the *unfold* operation to the node $v_q$ in $S^1_m$ and generate solution $S_m$. $S_m$ will contain *Type-1* edges, and another edge $e_r$ from $v_q$. In $S_m$, $v_q$ and $v_r$ are connected by $e_r$ and the *Type-2* edges are originated from $v_r$. We argue that $S_m$ is a valid solution of $T_{\alpha\beta}$ since the subtree rooted at nodes pointed by *Type-2* edges are not modified by the sequence of – $(a)$ the *folding* of $v_r$ to construct $T^1_{\alpha\beta}$ from $T_{\alpha\beta}$, and $(b)$ the *unfolding* of $v_q$ to construct $S_m$ from $S^1_m$.

Clearly any solution $S^{1'}_m$ of $T^1_{\alpha\beta}$ that does not contain node $v_q$ is valid solution for $T_{\alpha\beta}$ as well.

$\square$

**Lemma C.2** *If function Convert is applied to the root node of any AND/OR tree $T_{\alpha\beta}$, an alternating AND/OR tree $\hat{T}_{\alpha\beta}$ is generated.*

Proof: [**Lemma C.2**]  Function *Convert* traverses every intermediate node in a depth first manner. Consider any sequence of nodes, $v_{q_1}, v_{q_2}, \cdots, v_{q_n}$ of same type, where $v_{q_i}$ is the parent of $v_{q_{i+1}}$ in $T_{\alpha\beta}$ and $1 \leq i < n$. Obviously, the *fold* operation is applied to $v_{q_{i+1}}$ before $v_{q_i}$, where $1 \leq i < n$. In other words, the fold operation applied to the sequence of nodes in the reverse order and after *folding* $v_{q_{i+1}}$, all the edges of $v_{q_{i+1}}$ are modified and moved to $v_{q_i}$, where $1 \leq i < n$. When the function call $Convert(v_{q_2})$ returns, all the edges of $v_{q_2}, \cdots, v_{q_n}$ are already moved to $v_{q_1}$ and the sequence of nodes, $v_{q_1}, v_{q_2}, \cdots, v_{q_n}$ are flattened. Therefore, every sequence of nodes of same type are flattened, when the function call $Convert(v_R)$ returns, where $v_R$ is the root of $T_{\alpha\beta}$ and an alternating AND/OR tree $\hat{T}_{\alpha\beta}$ is generated.

**Lemma C.3** *If function Revert is applied to an alternating AND/OR tree $\hat{T}_{\alpha\beta}$, the update-list of every edge in $\hat{T}_{\alpha\beta}$ becomes empty.*

Proof: [**Lemma C.3**]  Follows from the description of *Revert*.

**Theorem C.1** *For any AND/OR tree $T_{\alpha\beta}$, it is possible to construct an alternating AND/OR tree $\hat{T}_{\alpha\beta}$ using function Convert, where the set of all possible solutions of $T_{\alpha\beta}$ is generated in the order of their increasing cost by applying Algorithm 4 to $\hat{T}_{\alpha\beta}$, and then converting individual solutions using function Revert.*

Proof: [**Theorem C.1**]  According to Lemma C.2, after the application of function *Convert* to $T_{\alpha\beta}$ an alternating AND/OR tree $\hat{T}_{\alpha\beta}$ is generated. Consider the intermediate AND/OR trees that are the generated after folding every node in $T_{\alpha\beta}$. Let $T^0_{\alpha\beta}, T^1_{\alpha\beta}, \cdots, T^n_{\alpha\beta}$ are the sequence of AND/OR trees and $T^0_{\alpha\beta} = T_{\alpha\beta}, \hat{T}_{\alpha\beta} = T^n_{\alpha\beta}$. Since $T^i_{\alpha\beta}$ is generated from $T^{i+1}_{\alpha\beta}$ after *folding* exactly one node in $T_{\alpha\beta}^i$, where $0 \le i < n$, according to Lemma C.1, the solutions of $T_{\alpha\beta}^i$ can be generated from $T_{\alpha\beta}^{i+1}$ by unfolding the same node. According to Lemma C.3, for any solution of $\hat{T}_{\alpha\beta}$, *Revert* unfolds every node $v_q$ in that solution, where $v_q$ was folded by *Convert* while transforming $T_{\alpha\beta}$ to $\hat{T}_{\alpha\beta}$. Therefore the solutions of $T_{\alpha\beta}$ can be generated from the solutions of $\hat{T}_{\alpha\beta}$.